\pdfoutput=1
\documentclass[10pt,twocolumn,letterpaper]{article}
\usepackage{preprint}
\usepackage[T1]{fontenc}
\usepackage{xspace}
\usepackage[dvipsnames]{xcolor}
\usepackage{graphicx}
\usepackage{amsmath,amssymb}
\usepackage{booktabs}
\usepackage[numbers,sort&compress]{natbib}
\usepackage[format=plain,labelformat=simple,labelsep=period,font=small]{caption}
\usepackage[font=footnotesize,skip=3pt,subrefformat=parens]{subcaption}
\usepackage[hyphens]{url}
\usepackage[shortlabels,inline]{enumitem}
\usepackage{placeins}
\usepackage{float}
\IfFileExists{stfloats.sty}{\usepackage{stfloats}}{}

\usepackage{multirow}
\usepackage{array}
\usepackage{colortbl}
\usepackage{amsthm}
\usepackage{bm}
\usepackage{tikz}
\usetikzlibrary{arrows.meta,calc,positioning,shapes.geometric}
\usepackage[skins,breakable]{tcolorbox}

\newcommand{\PENDING}[1]{\textcolor{red}{[\textbf{PENDING:} #1]}}
\makeatletter
\def\pend@compact#1/#2\@nil{\textcolor{red}{\textbf{[P#1]}}}
\newcommand{\compactpending}{\renewcommand{\PENDING}[1]{\pend@compact##1\@nil}}
\makeatother

\newcommand{\methodname}{LMSA\xspace}

\newcommand{\pithree}{$\pi^3$\xspace}

\newcommand{\Gbar}{\bar\Gamma}

\DeclareMathOperator{\Var}{Var}

\definecolor{hubearly}{HTML}{FAB04A}
\definecolor{hubmain}{HTML}{EE7A00}
\definecolor{bandc}{HTML}{1A6FAF}
\definecolor{routerc}{HTML}{A23B72}
\definecolor{hubdark}{HTML}{B86A00}
\definecolor{exactdark}{HTML}{1F6488}
\newcommand{\hubq}[1]{{\color{hubdark}#1}}
\newcommand{\exq}[1]{{\color{exactdark}#1}}

\definecolor{gammac}{HTML}{A23B72}
\definecolor{varc}{HTML}{2E7D32}
\definecolor{whyc}{HTML}{6E6E6E}
\newcommand{\gq}[1]{{\color{gammac}#1}}
\newcommand{\sq}[1]{{\color{varc}#1}}
\newcommand{\why}[1]{\qquad\text{\color{whyc}\footnotesize #1}}
\newcommand{\dstep}[1]{\par\smallskip\noindent{\color{exactdark}\bfseries #1}\enspace}

\definecolor{tabhead}{HTML}{E4ECF2}
\definecolor{tabgroup}{HTML}{FFFFFF}
\definecolor{tabours}{HTML}{EFEFEF}
\definecolor{failc}{HTML}{B03A2E}
\definecolor{passc}{HTML}{2E7D32}
\newcommand{\headrow}{}
\newcommand{\oursrow}{}
\newcommand{\fail}[1]{\textcolor{failc}{#1}}
\newcommand{\pass}[1]{\textcolor{passc}{#1}}

\definecolor{boxorange}{HTML}{F18F01}\definecolor{boxorangebg}{HTML}{FEF4E8}
\definecolor{boxgrey}{HTML}{8E8E8E}\definecolor{boxgreybg}{HTML}{F3F3F3}
\definecolor{boxblue}{HTML}{2E86AB}\definecolor{boxbluebg}{HTML}{EAF3F8}
\definecolor{boxpropbg}{HTML}{EAF5EA}
\tcbset{revarbox/.style={enhanced, breakable, frame hidden, boxrule=0pt, sharp corners, boxsep=0pt,
  left=5pt, right=4pt, top=2.5pt, bottom=2.5pt, before skip=4pt plus 1pt, after skip=4pt plus 1pt}}
\theoremstyle{plain}
\newtheorem{proposition}{Proposition}
\theoremstyle{definition}
\newtheorem{definition}{Definition}
\newtheorem{assumption}{Assumption}
\newtheorem{finding}{Finding}
\tcolorboxenvironment{proposition}{revarbox, colback=boxpropbg, borderline west={2pt}{0pt}{passc}}
\tcolorboxenvironment{finding}{revarbox, colback=boxgreybg, borderline west={2pt}{0pt}{boxgrey}}
\tcolorboxenvironment{definition}{revarbox, colback=boxbluebg, borderline west={2pt}{0pt}{boxblue}}
\tcolorboxenvironment{assumption}{revarbox, colback=boxorangebg, borderline west={2pt}{0pt}{boxorange}}

\AtEndPreamble{%
  \crefname{proposition}{Prop.}{Props.}%
  \Crefname{proposition}{Proposition}{Propositions}%
  \crefname{definition}{Def.}{Defs.}%
  \Crefname{definition}{Definition}{Definitions}%
  \crefname{assumption}{Assumption}{Assumptions}%
  \Crefname{assumption}{Assumption}{Assumptions}%
  \crefname{finding}{Finding}{Findings}%
  \Crefname{finding}{Finding}{Findings}%
  \crefname{figure}{Fig.}{Figs.}%
  \Crefname{figure}{Figure}{Figures}%
  \crefname{equation}{Eq.}{Eqs.}%
  \Crefname{equation}{Equation}{Equations}%
}

\renewcommand{\arraystretch}{1.06}

\makeatletter
\renewcommand\paragraph{\@startsection{paragraph}{4}{\z@}{2.2ex \@plus .4ex \@minus .2ex}{-1em}{\normalfont\normalsize\bfseries}}
\g@addto@macro\normalsize{%
  \setlength\abovedisplayskip{6pt \@plus 1.5pt \@minus 3pt}%
  \setlength\belowdisplayskip{6pt \@plus 1.5pt \@minus 3pt}%
  \setlength\abovedisplayshortskip{2pt \@plus 1pt}%
  \setlength\belowdisplayshortskip{4pt \@plus 1pt \@minus 2pt}}
\makeatother
\makeatletter
\DeclareRobustCommand\onedot{\futurelet\@let@token\@onedot}
\def\@onedot{\ifx\@let@token.\else.\null\fi\xspace}
\@ifundefined{eg}{\def\eg{\emph{e.g}\onedot}}{}
\@ifundefined{Eg}{}{}
\@ifundefined{ie}{}{}
\@ifundefined{Ie}{}{}
\@ifundefined{cf}{}{}
\@ifundefined{etc}{}{}
\@ifundefined{vs}{}{}
\@ifundefined{etal}{}{}
\makeatother
\definecolor{linkblue}{rgb}{0.21,0.49,0.74}
\usepackage[breaklinks,colorlinks,allcolors=linkblue]{hyperref}
\usepackage[capitalize]{cleveref}

\title{Layer-Matched Sparse Attention for Low-Redundancy 3D Reconstruction}
\author{
  Sammam Mahdi$^{1,*}$, \ Fariha Binta Salim$^{1}$, \ K. M. Jubair Sami$^{1}$, \ Rakin Bin Rabbani$^{1}$ \\[2pt]
  \normalfont\normalsize $^{1}$BRAC University \qquad $^{*}$Corresponding author \\[2pt]
  \normalfont\small\texttt{\{sammam.mahdi, fariha.binta.salim, km.jubair.sami\}@g.bracu.ac.bd} \quad
  \normalfont\small\texttt{rakin.rabbani@bracu.ac.bd}
}
\date{}

\begin{document}
\twocolumn[
  \begin{@twocolumnfalse}
\maketitle
\begin{abstract}
VGGT reconstructs cameras and geometry from unposed images in one pass, but its global attention grows quadratically with the number of views.
Training-free accelerators thin this attention and work well on redundant video, but lose pose accuracy when views share little geometry.
This paper traces these losses to two kinds of cross-view information that a uniform sparse approximation can fail to preserve together: diffuse layers aggregate distributed context, whereas matching layers depend on query-specific interactions that carry little attention mass.
A per-layer statistic measures the log-mass missed by a mean token. Its row-free proxy is the logit variance, and it selects the estimator for each layer.
Diffuse layers use shared, importance-sampled hub keys, with log-weights that make the softmax numerator and denominator unbiased. Matching layers stay exact: every aggressive thinning tested cost accuracy. At equal cost, the selected layers outperform the four shifted windows tested.
On 13 unseen low-redundancy scenes with 32 views, \methodname stays within 1.4 AUC@5 of dense attention, compared with 10.3 for AVGGT and 4.5 for SAF3R; its difference from contemporaneous GPA-VGGT is not significant, and only its gains over AVGGT pass both scene-level tests. With 100 views, it stays within 1.9 AUC@5 while its aggregator runs 2.0$\times$ faster.
Used post hoc to choose the layers GPA-VGGT protects, the statistic improves it by 1.2 AUC@5 for 5\% more FLOPs. Applied untuned to \pithree, it beats AVGGT but remains further from dense accuracy.
\end{abstract}

\vspace{0.35cm}
  \end{@twocolumnfalse}
]

\section{Introduction}
\label{sec:intro}

\begin{figure*}[t]
\centering
\includegraphics[width=\textwidth]{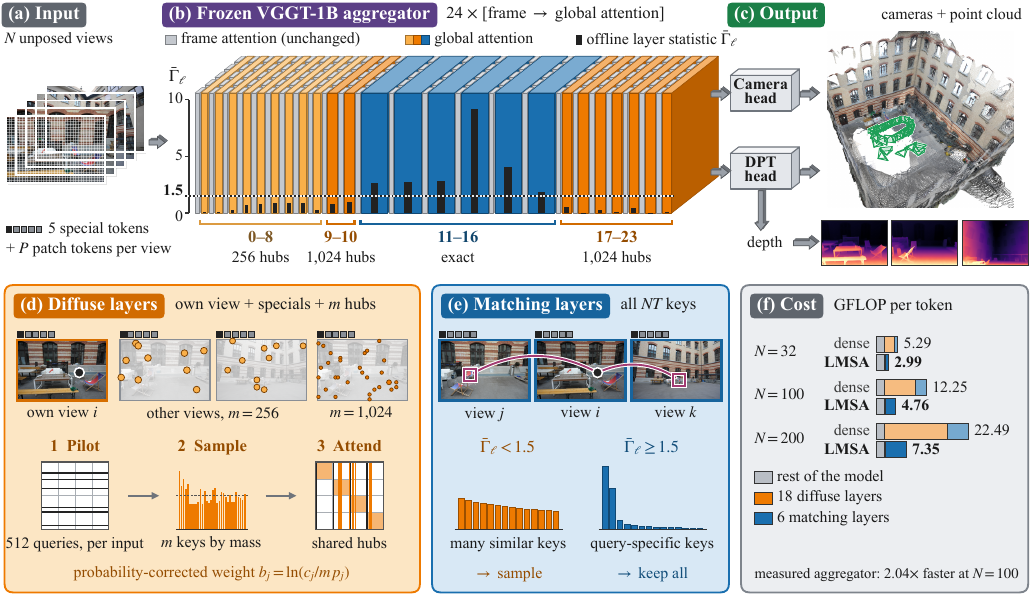}
\vspace{-1.4em}
\caption{\textbf{\methodname on a frozen VGGT-1B.}
\textbf{(a)}~$N$ unposed views (ETH3D courtyard~\cite{schops2017eth3d}), each $P$ patch and 5 special tokens.
\textbf{(b)}~Frame attention (grey) is unchanged. The offline statistic $\Gbar_\ell$ (black bars, once per model, \cref{sec:selector}) sets each global layer's estimator: above 1.5 (dotted), \emph{matching} layers (blue) keep exact attention; \emph{diffuse} layers (orange) use 256 or 1024 shared hubs. Block width: keys per query at $N{=}32$, log scale.
\textbf{(c)}~Cameras, point cloud and depth predicted by \methodname.
\textbf{(d)}~A diffuse-layer query (black dot) attends to its own view, all special tokens and $m$ hubs drawn per input (\cref{sec:hubs}); hub counts to scale, positions illustrative.
\textbf{(e)}~In a matching layer every token is a key (magenta: true correspondences).
\textbf{(f)}~Exact cost (\cref{sec:cost}); pale: the same layers at dense cost.}
\label{fig:overview}
\end{figure*}

Feed-forward reconstruction transformers~\cite{wang2024dust3r,leroy2024mast3r,wang2025vggt,wang2025pi3,yang2025fast3r,wang2025cut3r} predict cameras, depth and point maps for a set of unposed images in one forward pass.
VGGT~\cite{wang2025vggt} alternates attention within each frame with \emph{global} attention over the tokens of all $N$ views, and the global part grows as $N^2$.
A fast-growing family of training-free accelerators therefore thins global attention~\cite{sun2025avggt,shen2025fastvggt,wang2025blocksparse,you2026regimevggt,wang2025httm,sung2027gpavggt,deng2026saf3r}.
Many rely on the premise that cross-view information is redundant, so that most cross-view keys can be dropped, merged or summarised with little effect on geometry.
The premise often holds for overlapping video but breaks when views are few and far apart, a setting this paper calls \emph{low-redundancy} and measures by $\rho$, the mean number of views that see a point.
The resulting failure is severe: on ETH3D courtyard ($\rho\approx7.7$), AVGGT~\cite{sun2025avggt}, keeping one key per $3{\times}3$ patch cell ($\sigma{=}9$), lowers the pose AUC@5 of a frozen VGGT-1B from 58.3 to 8.1 and its AUC@30 from 90.9 to 34.8, averaged over five frame subsets.

AVGGT~\cite{sun2025avggt} observed that VGGT's early global layers form no cross-view correspondences, its middle layers align the views and its late layers only refine.
Thinning one group at a time (\cref{tab:factorial}) shows that these roles require different cross-view information.
The middle layers give the clearest example. Across three low-redundancy development scenes, removing the model's small correspondence-key sets from layers 10--16 (on TUM fr1\_room, 30 of 33{,}312 keys per query) costs 17.7--49.2 AUC@5.
Removing as many of the heaviest other keys costs 6.0--23.0, even though they carry 2.3--3.4$\times$ more attention mass; removing random keys costs nothing.
The keys with the highest query-specific lift cost the most, although only 13\% lie in correspondence windows (\cref{fig:stats}c).
A key can therefore carry little attention mass yet be critical to geometric reconstruction.
The other layers fail in different ways. Making early layers 0--8 frame-local, as AVGGT does, costs 33.1 AUC@5 on courtyard because these layers carry distributed cross-view context.
Layers 9 and 17--23 are diffuse again, but thinning them accounts for 14.8 of the 22.1 points AVGGT loses on the redundant TUM fr3\_long\_office scene: its mean token under-counts the dropped mass (\cref{sec:hubs}).

Attention approximation is therefore an estimation problem with two regimes.
A diffuse layer needs an accurate estimate of a sum over many comparable terms, whereas a matching layer needs query-specific terms that a small shared sample does not contain.
One per-layer statistic separates them: the log-mass $\Gamma$ that a mean token misses from a query's softmax denominator is about half the logit variance $s^2$ when the logits spread little (\cref{prop:gamma}), and $s^2$ needs only the key covariance, not attention rows.
A cut-off of 1.5 nats ($s^2\approx3$), chosen on the development scenes and fixed before any test run, labels layers 11--16 as matching; that these layers need exact attention, not just a better sample, is an empirical finding that \cref{sec:composed} tests.

\methodname (layer-matched sparse attention, \cref{fig:overview}) gives each regime its own estimator.
\emph{Diffuse} layers attend to their own frame, the special tokens and a few hundred shared \emph{hub} keys, importance-sampled per input from the attention of 512 pilot queries, with log-weights that make the softmax numerator and denominator unbiased.
Their per-query cost hardly grows with $N$, and they remove both the early failure and the late one, which is consistent with the mean token's under-weighting of the dropped keys.
\emph{Matching} layers keep exact attention: every aggressive way of thinning them that was tested, from truncated lattices with routed correspondence blocks to merging most keys, cost accuracy once the other layers were sparse (\cref{sec:composed}); the routed design is kept as a faster variant, \methodname-fast.

On 13 unseen low-redundancy scenes at $N{=}32$, named before any test run, \methodname stays within 1.4 AUC@5 of dense attention (95\% CI [0.7, 2.0]), the lowest mean loss of the training-free methods tested, against 10.3 to 18.9 for AVGGT ($\sigma{=}4$, 9), which is 7--18\% faster, and 4.5 for SAF3R; with 100 views it stays within 1.9 AUC@5 while its aggregator runs 2.0$\times$ faster.
At equal cost, the statistic's six exact layers beat the four shifted windows of six tested (\cref{sec:composed}).
The contemporaneous GPA-VGGT, as fast, loses 2.4, so the margin of \methodname (1.0, CI [$-0.3$, 2.5]) is not significant, and on 8 unseen redundant scenes GPA-VGGT loses 0.1 against 0.7 (\cref{sec:main}); protecting the statistic's layers 11--16 instead of 13--16 improves GPA-VGGT by 1.2 AUC@5 (CI [0.4, 2.1], post hoc).

\noindent The paper makes three contributions:
\begin{enumerate}
  \item \textbf{An interpretable criterion for choosing the attention estimator per layer:} controlled interventions show that reconstruction needs both distributed context and query-specific interactions that carry little attention mass, and the mean-token log-mass deficit, with a row-free logit-variance proxy, separates the layers that need each; simpler concentration statistics agree (\cref{sec:failures,sec:selector}).
  \item \textbf{\methodname:} probability-corrected shared hubs in diffuse layers and exact attention in matching layers, with no training and no known poses (\cref{sec:hubs,sec:band}).
  \item \textbf{Validation:} on 21 pre-named test scenes \methodname has the lowest mean loss of the training-free methods tested; at equal cost its exact layers beat the four shifted windows tested, the statistic also improves GPA-VGGT on the same backbone (post hoc), and a registered test on NRGBD, Sintel and long video confirms a variant with more hubs (\cref{sec:experiments}).
\end{enumerate}

\section{Related Work}
\label{sec:related}

\paragraph{Feed-forward 3D reconstruction.}
DUSt3R~\cite{wang2024dust3r} and MASt3R~\cite{leroy2024mast3r} regress point maps from image pairs; VGGT~\cite{wang2025vggt}, \pithree~\cite{wang2025pi3} and Fast3R~\cite{yang2025fast3r} process many views jointly with global attention, and CUT3R~\cite{wang2025cut3r} with a persistent state.
\citet{bratulic2025geometric} show that correspondence and epipolar structure emerge in the intermediate-layer attention of VGGT and related models, the behaviour that the matching layers of \cref{sec:band} rely on.
\methodname keeps these models frozen and changes only which keys their global attention reads.

\paragraph{Accelerating global attention in 3D transformers.}
FastVGGT~\cite{shen2025fastvggt}, HTTM~\cite{wang2025httm}, Co-Me~\cite{chen2025come} and, with fine-tuning, LiteVGGT~\cite{shu2026litevggt} merge tokens; QuantVGGT~\cite{feng2026quantvggt} quantizes the model, which is complementary.
AVGGT~\cite{sun2025avggt} analyses VGGT's global attention layer by layer and finds that early global layers form no cross-view correspondences, middle layers align the views and late layers only refine; it makes the global layers below a hand-set boundary frame-local and, in later layers, keeps one key per $\sigma$-cell, the reference frame and a count-biased mean token.
\methodname does not claim the role split, which AVGGT and RegimeVGGT establish; it asks which \emph{estimator} each layer's information needs, choosing between a shared sample and exact attention by a per-layer statistic, and replaces the frame-local early layers, which fail on low-redundancy input (\cref{tab:factorial}), and the biased mean token by probability-corrected shared hubs.
Block-sparse attention~\cite{wang2025blocksparse} keeps key blocks chosen by pooled scores, and HeSS~\cite{kim2026hess} redistributes its budget among the heads of each layer by a gradient-based sensitivity score; RegimeVGGT~\cite{you2026regimevggt} analyses VGGT's layer regimes and merges tokens and subsamples keys and values on a layer-wise schedule, choosing how much to compress rather than which estimator to use, and Good Token Hunting~\cite{zheng2026gotohunt} selects frames and prunes tokens per layer.
Other methods sparsify whole frames~\cite{li2026svggt}, as does the contemporaneous BlazeVGGT~\cite{yao2026blazevggt}, partition the views into chunks~\cite{park2026diversity}, or train and fine-tune~\cite{wang2025flashvggt,huang2026turbovggt,ren2026speed3r}.
The contemporaneous GPA-VGGT~\cite{sung2027gpavggt} adds to the early frame attention a shared cross-frame set of per-head cluster summaries with a $+\log(\text{size})$ bias, as in the proportional attention of ToMe~\cite{bolya2023tome}, and merges keys and values later; its merging operator is used in the merged variants of \cref{sec:composed}.
The hubs of \methodname are an importance sample with probability-correcting log-weights rather than summaries, and the same estimator serves the late layers, where it removes an estimation bias (\cref{sec:hubs}).
SAF3R~\cite{deng2026saf3r} assigns each head one of several sparse patterns from offline calibration and, for correspondence heads, keeps the top-$K$ keys per frame by feature similarity.

\paragraph{Sparse, sampled and summarised attention.}
HyperAttention~\cite{han2024hyperattention}, Scatterbrain~\cite{chen2021scatterbrain} and Support Basis~\cite{aliakbarpour2025support} (beyond the bounded-entry regime of~\cite{alman2023bounded}) keep large entries exactly and estimate the rest by uniform samples, low-rank factors or polynomials, MagicPIG~\cite{chen2024magicpig} estimates attention by hashed sampling, and vAttention~\cite{desai2026vattention} adds a uniform sample of the remaining keys to exact top-$k$ keys; Skeinformer~\cite{chen2022skeinformer} samples key columns from a pilot, the mechanism closest to the hubs of \methodname.
FMMformer~\cite{nguyen2021fmmformer}, Fast Multipole Attention~\cite{kang2023fma}, Multipole Semantic Attention~\cite{mitchell2025muse} and Multipole Attention~\cite{hooper2025multipole} treat distant or dissimilar keys coarsely (low-rank, hierarchical or by cluster centroids), the last also picking keys for exact attention by centroid.
In LLMs and video models the estimator is chosen per head, by retrieval heads~\cite{xiao2025duoattention,tang2025razorattention}, offline search or a divergence statistic~\cite{jiang2024minference,lai2025flexprefill}, sampled queries~\cite{zhu2025sampleattention} or online profiling~\cite{xi2025sparsevideogen}, and Kascade~\cite{deshmukh2025kascade} computes exact top-$k$ keys only in anchor layers chosen on a development set.
\methodname makes such a choice per layer from a closed-form statistic calibrated offline on development scenes, without gradients or a search over layer windows, and pairs it with a probability-corrected estimator.

\paragraph{Statistical physics and hubness.}
The random-energy model~\cite{derrida1981rem} has a freezing transition at which a partition function becomes dominated by its extreme terms; it has been mapped onto attention at initialisation~\cite{giorlandino2025two} and onto retrieval in dense associative memories~\cite{lucibello2024exponential}, and \methodname uses it at inference to motivate a choice of estimator.
Hubs that attract many queries are a known failure of similarity retrieval, corrected in text--image matching by the inverted softmax~\cite{liu2019strong}; the popularity--specificity split of \cref{sec:band} follows the same factorisation.

\section{Method}
\label{sec:method}

\begin{figure*}[t]
\centering
\includegraphics[width=\textwidth]{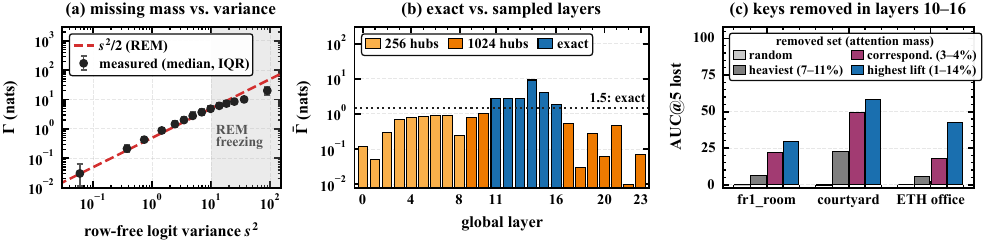}
\vspace{-1.6em}
\caption{\textbf{One statistic assigns the layer roles; attention mass is not importance.}
\textbf{(a)}~Log-mass deficit $\Gamma$ of a count-biased mean token against the row-free logit variance $s^2$ (\cref{eq:gamma,eq:rem}), for 1.15M (query, layer, head) triples from layers 9--23 of VGGT-1B on 8 scenes (median, IQR); it follows $s^2/2$ up to $s^2\approx10$ and falls below it at larger variance (grey), as the Gaussian REM, or light-tailed or correlated logits, would.
\textbf{(b)}~Offline layer statistic $\Gbar_\ell$ (median head, mean of the 5 development scenes), coloured by the estimator \methodname assigns; dotted: the empirical cut-off (1.5 nats).
\textbf{(c)}~AUC@5 lost when one key set is removed from layers 10--16 only (three low-redundancy development scenes, 5 seeds; Tab.~\ref{tab:knockout}); in brackets, the attention mass removed.}
\label{fig:stats}
\end{figure*}

\subsection{Setting}
\label{sec:setting}

VGGT-1B~\cite{wang2025vggt} cuts each of $N$ images into $P$ patch tokens and adds five special tokens (one camera, four register), so a frame has $T{=}P{+}5$ tokens ($T{=}1041$ for $392{\times}518$ images).
It alternates 24 frame blocks with 24 global blocks, numbered 0--23, of 16 heads of width $d_h{=}64$; \methodname changes only the global blocks.
Fix one query token $t$ with query vector $q_t$, one head and one layer.
A key $u$ with key vector $k_u$ and value $v_u$ has logit $l_u=\langle q_t,k_u\rangle/\sqrt{d_h}$ and weight $w_u=e^{l_u}$, and dense global attention over the set $\mathcal U$ of all $NT$ tokens returns
\begin{equation}
  o_t=\frac{\sum_{u\in\mathcal U}w_u v_u}{\underbrace{\textstyle\sum_{u\in\mathcal U}w_u}_{Z\text{: partition function}}}.
  \label{eq:dense}
\end{equation}
A sparse layer that misestimates $Z$ rescales every kept key, even correctly chosen ones.
Every sparse layer in this paper, of \methodname and of the baselines alike, keeps an \exq{\emph{exact}} key set \exq{$\mathcal K(t)$} and replaces the dropped set $D(t)=\mathcal U\setminus\mathcal K(t)$ by a few \hubq{\emph{stand-in}} keys \hubq{$\mathcal W(t)$}, each with a log-weight \hubq{$b_u$}:
\begin{equation}
  \hat o_t=\frac{\overbrace{\exq{\textstyle\sum_{u\in\mathcal K}w_u v_u}}^{\text{exact keys}}+\overbrace{\hubq{\textstyle\sum_{u\in\mathcal W}e^{b_u}w_u v_u}}^{\text{stand-ins for } D}}{\exq{\textstyle\sum_{u\in\mathcal K}w_u}+\hubq{\textstyle\sum_{u\in\mathcal W}e^{b_u}w_u}}.
  \label{eq:sparse}
\end{equation}
The factor $e^{b_u}$ states how many dropped keys the stand-in $u$ represents.
AVGGT's mean token, for example, is one stand-in with the mean key $\bar k_D$ and mean value $\bar v_D$ over $D$ and $b=\ln|D|$; truncated attention has $\mathcal W=\emptyset$; and the hubs of \methodname are sampled keys with importance weights (\cref{sec:hubs}).
Costs are counted in \emph{units}: one unit is one frame's $T$ keys for every query, so a dense global layer costs $N$ units.

\begin{table}[t]
\centering
\small
\setlength{\tabcolsep}{3pt}
\caption{\textbf{Three groups fail for different reasons.}
AUC@5 lost by frozen VGGT-1B when one group of global layers is thinned AVGGT-style and the rest stay dense: E, layers 0--8 made frame-local; B, layers 10--16 on a $\sigma{=}9$ lattice with a mean token; L, layers 9 and 17--23 likewise.
Mean of 5 frame subsets, $N{=}32$ ($N{=}22$ for ETH3D office); $\rho$ as in \cref{sec:protocol}; bold: largest single-group loss.}
\label{tab:factorial}
\begin{tabular}{@{}lcccc@{}}
\toprule
\headrow Scene ($\rho$) & E (0--8) & B (10--16) & L (9, 17--23) & all \\
\midrule
TUM fr1\_room (4.3) & 5.6 & \textbf{11.7} & $-0.4$ & 13.3 \\
ETH3D courtyard (7.7) & \textbf{33.1} & 28.5 & 5.4 & 50.2 \\
ETH3D office (5.5) & 1.5 & \textbf{6.0} & 3.5 & 12.3 \\
TUM fr3\_long\_office (15.4) & 0.5 & 4.4 & \textbf{14.8} & 22.1 \\
7-Scenes office (13.0) & $-1.1$ & 0.8 & \textbf{1.6} & 1.8 \\
\bottomrule
\end{tabular}
\end{table}

\subsection{Why one sparse pattern fails}
\label{sec:failures}

In the terms of \cref{eq:sparse}, the three failures of \cref{tab:factorial} are three different estimation errors.
In the early layers AVGGT keeps only the own frame, so the cross-view part of a diffuse sum is \emph{missing}; adding back the special tokens of all frames still loses 24.2 AUC@5 on courtyard.
In the middle layers a small set of correspondence keys matters for pose far beyond its attention mass, and the lattice misses them.
In layers 9 and 17--23 the sum is diffuse again, but the mean token \emph{under-counts} it; the size of this under-count is the statistic that the method builds on.

\subsection{One statistic: the mean-token deficit}
\label{sec:selector}

For a query $t$ and its dropped set $D$, the \textbf{mean-token deficit} $\bm{\Gamma_t}$ compares the true mass of the dropped keys with the mass that a count-biased mean token gives them:
\begin{equation}
  \gq{\Gamma_t}=\underbrace{\ln\textstyle\sum_{u\in D}w_u}_{\text{true log-mass of }D}\;-\;\underbrace{\big(\ln|D|+\bar l_D\big)}_{\text{log-weight of the mean token}}.
  \label{eq:gamma}
\end{equation}
The mean token's log-weight adds the count bias $\ln|D|$ to the logit of the mean key, $\bar l_D=\langle q_t,\bar k_D\rangle/\sqrt{d_h}$.
Because logits are linear in the key, this is also the mean logit $\frac{1}{|D|}\sum_{u\in D}l_u$.
$\Gamma_t{=}0$ means that the mean token reproduces the dropped mass; $\Gamma_t{>}0$ means that it gives only $e^{-\Gamma_t}$ of it.

\begin{proposition}[Mean-token deficit]
\label{prop:gamma}
Let $p_\tau(u)\propto e^{\tau l_u}$ on $D$ for $\tau\in[0,1]$.
\textbf{(i) Exact:} for some $\xi\in(0,1)$, $\gq{\Gamma_t}=\tfrac12\Var_{p_\xi}(l)\ge0$, with equality iff the logits are constant on $D$.
\textbf{(ii) Small spread:} with $\Sigma_K$ the $d_h{\times}d_h$ covariance of the keys in $D$,
\begin{equation}
  \begin{aligned}
  \gq{\Gamma_t}&\approx\tfrac12\,\sq{\bm{s_t^2}},\\
  \sq{s_t^2}&=\Var_{p_0}(l)=\underbrace{q_t^{\top}\Sigma_K\,q_t/d_h}_{\text{row-free}}.
  \end{aligned}
  \label{eq:s2}
\end{equation}
\end{proposition}
\noindent
Here $p_0$ is uniform on $D$; the proof (App.~\ref{sec:proof-gamma}) writes $\Gamma$ as a Taylor remainder.
The useful consequence is \cref{eq:s2}: the \textbf{logit variance} \sq{$s^2$} needs one key covariance per layer and head and $O(d_h^2)$ work per query and no attention row, hence \emph{row-free}.

\paragraph{Random-energy reading.}
A classical model shows where $\Gamma\approx s^2/2$ breaks.

\begin{assumption}[Random-energy model, REM~\cite{derrida1981rem}]
\label{ass:rem}
The $n=|D|$ dropped logits of a query behave like independent Gaussians with variance $s^2$.
\end{assumption}
\noindent
For large $n$, \cref{ass:rem} gives two phases separated by the \textbf{freezing variance} $2\ln n$:
\begin{equation}
  \gq{\Gamma}\approx\begin{cases}
    \hubq{\tfrac12 s^2}, & s^2<2\ln n\ \ \hubq{\text{\scriptsize(high temp.)}},\\[3pt]
    \exq{\sqrt{2s^2\ln n}-\ln n}, & s^2\ge2\ln n\ \ \exq{\text{\scriptsize(frozen)}}.
  \end{cases}
  \label{eq:rem}
\end{equation}
In the frozen phase the largest logits carry $Z$, and the correction $s^2/2$ overshoots by $(s-\sqrt{2\ln n})^2/2$.
On VGGT (\cref{fig:stats}a), measured $\Gamma$ stays within 0.3 nats of $s^2/2$ up to $s^2\approx10$ (rank correlation 0.986), consistent with \cref{eq:s2}, and then bends below it, earlier than the REM's $2\ln n\approx20.8$, as correlated or light-tailed logits would; the REM thus motivates the cut-off rather than setting it.
Below freezing, an importance sample estimates $Z$ and the numerator of \cref{eq:dense} without bias, while a mean token is short by about $s^2/2$ nats; above it, a shared sample of fixed size misses the extreme keys.
\Cref{prop:gamma} is exact, \cref{eq:s2} approximate, and the cut-off below empirical, validated in \cref{sec:composed}: $\Gamma$ measures how badly a mean token fails; that high-$\Gamma$ layers need exact attention rather than a better sample is empirical.

\paragraph{Selector.}
For each global layer $\ell$, the selector computes $\Gamma_t$ for 512 random patch queries per pilot scene and frame subset, with $D$ set to all patch keys of the other frames, and takes the median over heads and then the mean over pilot queries and pilot scenes to obtain the \textbf{pilot statistic} $\bm{\Gbar_\ell}$, a layer-level aggregate of the per-query $\Gamma_t$.
Layer $\ell$ is a \exq{\emph{matching}} layer if $\Gbar_\ell\ge1.5$ nats and keeps exact attention (\cref{sec:band}); otherwise it is a \hubq{\emph{diffuse}} layer and uses hubs (\cref{sec:hubs}).
The cut-off was chosen on the development scenes of VGGT after the factorial of \cref{tab:factorial} and fixed before any composed or test run.
$\Gbar_\ell$ is 1.83--9.21 in layers 11--16 and at most 1.00 elsewhere (\cref{fig:stats}b), so the matching set is 11--16, on each of the five development scenes.
The row-free form gives the same split: among layers 9--23, the median-head $s^2$ is at least 3 ($=2\times1.5$) in exactly layers 11--16 and at most 1.6 elsewhere.
The cut-off (1.5 nats, $s^2\approx3$) lies well below freezing: only layer 14 is frozen, and what the matching layers need are query-specific keys rather than heavy ones (\cref{sec:band}).
The selector runs once per model, offline, on the development scenes, and is applied to a new model without retuning (on \pithree it misses its target, \cref{sec:xarch}); at inference only the hub pilot runs.

\subsection{Diffuse layers: importance-sampled hubs}
\label{sec:hubs}

In a diffuse layer every query receives the same shared set of stand-in keys, drawn once per layer and input in three steps.
\emph{Hub pilot} (per input): 512 random patch queries $\mathcal R$ compute their full attention and estimate each key's \textbf{column mass} $\hat\pi_j=\frac{1}{|\mathcal R|}\sum_{t\in\mathcal R}\bar A(t,j)$, where $\bar A$ is the head-averaged attention.
\emph{Sample:} over the patch tokens $\mathcal C$, the proposal is $p_j\propto\hat\pi_j+\epsilon/|\mathcal C|$, where the floor $\epsilon{=}10^{-3}$ keeps every key reachable; $m$ i.i.d.\ draws from $p$ give the distinct \textbf{hubs} \hubq{$\mathcal H$} with multiplicities $c_j$.
\emph{Attend:} a query $t$ in frame $i$ uses \cref{eq:sparse} with the exact set $\mathcal K(t)=\mathcal F_i\cup\mathcal S$, the query's own frame $\mathcal F_i$ and the set $\mathcal S$ of all special tokens, and the stand-ins \hubq{$\mathcal W(t)=\mathcal H\setminus\mathcal F_i$} with the \textbf{hub log-weight}
\begin{equation}
  \hubq{b_j}=\underbrace{\ln c_j}_{\text{draws of }j}-\underbrace{\ln(m\,p_j)}_{\text{expected draws of }j}.
  \label{eq:hubs}
\end{equation}

\begin{proposition}[Unbiased hubs]
\label{prop:hubs}
For any function $f$ of a key, because $\mathbb E[c_j]=mp_j$,
\begin{equation*}
  \mathbb E\Big[\textstyle\sum_{j\in\mathcal W}\hubq{e^{b_j}}f(j)\Big]=\textstyle\sum_{j\in\mathcal C\setminus\mathcal F_i}f(j).
\end{equation*}
With $f(j)=w_jv_j$ and $f(j)=w_j$, \textbf{both the numerator and the denominator} of the dropped part are unbiased.
\end{proposition}
\noindent
A key drawn $c_j$ times enters once with weight $c_j/(mp_j)$, which corrects for both its draws and its selection probability.
The output is a ratio of the two estimates and is not itself unbiased; its bias is of order $1/m$ (App.~\ref{sec:proof-hubs}).
By contrast, the mean token under-counts the denominator by the factor $e^{-\Gamma}$ for every query, regardless of sample size (\cref{prop:gamma}).
Because $\mathcal H$ is shared, a hub layer needs no per-query gathers (App.~\ref{sec:recipe}) and costs about 1.6 units for $m{=}256$ and 2.4 for $m{=}1024$ at $N{=}32$, a cost that grows with $N$ only through the special tokens.
The statistic chooses the estimator, not $m$: \methodname uses $m{=}256$ in layers 0--8 and $m{=}1024$ in layers 9, 10 and 17--23, set on the development scenes, where 256 hubs in the later layers lose 11.4 AUC@5 on courtyard (2 subsets).

\paragraph{The late wall as estimation bias.}
The same deficit offers an explanation for why AVGGT's late layers fail on redundant scenes.
These layers keep the reference frame dense, a $\sigma{=}9$ lattice elsewhere and one mean token for the rest.
If $\varphi=\sum_{D}w_u/Z$ is the share of the true mass that falls on dropped keys, \cref{eq:sparse,eq:gamma} inflate every kept key by the factor
\begin{equation}
  \kappa=\frac{Z}{\hat Z}=\Big[\underbrace{(1-\varphi)}_{\text{kept share}}+\underbrace{\varphi\,e^{-\gq{\Gamma}}}_{\text{dropped share, under-counted}}\Big]^{-1}\ge 1,
  \label{eq:inflation}
\end{equation}
with $\hat Z$ the estimated denominator; $\kappa$ equals 1 only if $\varphi{=}0$ or $\Gamma{=}0$.
In late layers $\Gamma$ is small, but the inflation acts in one direction and falls mostly on the dense reference frame, so it predicts a systematic camera error rather than noise.

\begin{finding}[Late-layer failure]
\label{find:late}
With layers 9 and 17--23 thinned AVGGT-style (per-query mean token, other layers dense), adding the second-cumulant term of \cref{eq:s2} (capped at 3 nats) to the mean token's logit and the first-order term $\mathrm{Cov}_D(v,l)$ to its value recovers 9.7 of the 15.0 AUC@5 points lost on TUM fr3\_long\_office and cuts the rotation deviation from the dense prediction from 1.01$^\circ$ to 0.22$^\circ$ (2 subsets).
Replacing the dense reference frame, the lattice and the mean token by the own frame and 1024 hubs, which carry no such bias, leaves losses of at most 1.17 AUC@5 on the five development scenes (5 subsets) and a rotation deviation of 0.12$^\circ$ on TUM fr3\_long\_office.
\end{finding}
\noindent
The hub arm also changes the kept keys, so it supports the bias explanation without isolating it, and the cumulant correction serves only as a diagnosis: in the frozen layer 14 it overshoots by 3--9 nats, as \cref{eq:rem} predicts.

\subsection{Matching layers stay exact}
\label{sec:band}

A matching layer keeps dense attention, $\mathcal K(t)=\mathcal U$ and $\mathcal W(t)=\emptyset$ in \cref{eq:sparse}, at a price of $N$ units per layer.
\Cref{sec:composed} shows why: the truncated key sets tested there fail once the diffuse layers are also sparse.
\paragraph{Query-specific relevance, not popularity.}
Writing attention as $A(t,j)=\pi_j\,L(t,j)$ separates the column mass $\pi_j$ of key $j$, its \emph{popularity} and the quantity that hubs sample, from the \textbf{lift} $L(t,j)$, its \emph{specificity} to the query.
Removing in layers 10--16, per query, as many highest-lift keys as the query has correspondence keys, which hold 1--14\% of the attention mass per layer, drives AUC@5 to 0.05 / 0.13 / 0.59 on the three low-redundancy development scenes, although only 13\% of these keys lie in correspondence windows; removing a comparable 7--11\% of mass chosen by attention weight costs far less (\cref{sec:intro}).
A shared sample of popular keys is therefore unlikely to contain what these layers need.

\paragraph{Truncation amplifies whatever it keeps.}
A sparse matching layer is in effect truncated attention: the smallest layer statistic here is $\Gbar_\ell{\approx}1.8$, and a query with that deficit would give its mean token only $e^{-1.8}\approx17\%$ of the mass it stands for (an illustration, not a bound on individual queries), so \cref{eq:inflation} amplifies the kept keys by an estimated 2.0--3.8$\times$ for a $\sigma{=}9$ lattice.
With the right keys this sharpening helps: a $\sigma{=}4$ lattice with oracle correspondence keys matches dense accuracy as the only sparse group (loss $-1.38$, CI $[-3.81, 0.29]$).
But it applies equally to every error in the kept keys, including errors that sparse upstream layers pass in to the few query-specific, low-mass keys these layers depend on (App.~\ref{sec:evidence}).
The cheaper \methodname-fast nevertheless replaces exact matching layers by a $\sigma{=}4$ lattice plus routed correspondence blocks (App.~\ref{sec:fast}).

\begin{definition}[\methodname]
\label{def:lmsa}
For a frozen alternating-attention transformer, one offline dense pass on development scenes gives $\Gbar_\ell$ for every global layer $\ell$.
If $\Gbar_\ell\ge1.5$ nats, layer $\ell$ is \exq{matching} and keeps exact attention, $\mathcal K(t)=\mathcal U$.
Otherwise it is \hubq{diffuse}: a query in frame $i$ attends exactly to $\mathcal F_i\cup\mathcal S$ and to $m$ shared hubs drawn from the per-input hub pilot's column mass, weighted by \cref{eq:hubs}.
On VGGT-1B this gives matching layers 11--16, $m{=}256$ in layers 0--8 and $m{=}1024$ in layers 9--10 and 17--23.
\end{definition}

\subsection{Cost}
\label{sec:cost}

The cost of \methodname follows from the units of its layers.
Per token, a forward pass costs
\begin{equation}
  G(N)=\underbrace{2.02}_{\text{rest of net}}+0.004264\Big(\underbrace{\hubq{\textstyle\sum_{\ell\,\text{diff.}}u_\ell}}_{\approx\,\text{const in }N}+\underbrace{\exq{6N}}_{\text{exact layers}}\Big)
  \label{eq:cost}
\end{equation}
GFLOP, where 2.02 GFLOP is the cost outside global attention, $u_\ell$ the number of units of diffuse layer $\ell$, and 0.004264 GFLOP the cost of one unit per token.
Because the 18 diffuse layers have a nearly fixed cost, the attention cost approaches a quarter of dense as $N$ grows.
In this model, \methodname needs 2.99 / 3.82 / 4.76 / 7.35 GFLOP per token at $N{=}32$ / 64 / 100 / 200, against 5.29 / 8.57 / 12.25 / 22.49 for dense (\cref{tab:costmain}).

\section{Experiments}
\label{sec:experiments}

The experiments ask whether the design holds on unseen scenes (\cref{sec:main}), why the matching layers must stay exact (\cref{sec:composed}), what it costs and how far the selector transfers (\cref{sec:speed,sec:xarch}).

\subsection{Protocol}
\label{sec:protocol}

\paragraph{Model and inputs.}
All runs use the released VGGT-1B~\cite{wang2025vggt} with unchanged weights, the aggregator under bf16 autocast and the camera and depth heads in fp32.
Each scene contributes $N{=}32$ views at 518\,px, or $n{-}4$ of its $n$ views if it has fewer (\eg, 22 of 26 for ETH3D office), and a \emph{seed} is one frame subset (App.~\ref{sec:repro}); every number is a mean over 5 seeds unless marked. Ground-truth cameras only build and order the subsets, identically for every method, and never enter inference; in a random order the loss of \methodname changes by $+0.08$ AUC@5 (CI $[-1.05, 1.21]$, App.~\ref{sec:repro}).

\paragraph{Redundancy and scenes.}
The redundancy $\rho$ of a configuration is one plus the mean number of other views in which a patch's point is visible, with visibility taken from the dense model's own depth and cameras because Tanks and Temples has no ground-truth depth; a configuration is low-redundancy if $\rho<10.5$.
The 13 \emph{unseen low-$\rho$} test scenes come from ETH3D~\cite{schops2017eth3d} (8), Tanks and Temples~\cite{knapitsch2017tanks} (4) and TUM~\cite{sturm2012tum} (1), and the 8 \emph{unseen high-$\rho$} test scenes from TUM, 7-Scenes~\cite{shotton2013scene}, DTU~\cite{aanaes2016dtu} and Tanks and Temples (lists in App.~\ref{sec:scenes}).
The 5 \emph{development} scenes are TUM fr1\_room, ETH3D courtyard and office (low $\rho$), and TUM fr3\_long\_office and 7-Scenes office (high $\rho$).

\paragraph{Development protocol.}
Every design choice in \cref{sec:method} was made on the development scenes, with acceptance thresholds fixed before the corresponding runs (App.~\ref{sec:bars}).
\emph{Registered} tests had their plan and decision rules hashed before any run (App.~\ref{sec:bars}).
The 21 test scenes were named before any test run, and the frozen recipe was evaluated on them once, for \cref{tab:main}; later analyses on them are marked post hoc; development results, including the variants that failed, are reported separately (App.~\ref{sec:pilot}, \ref{sec:diffuse}).
Three high-$\rho$ test scenes (7-Scenes chess, DTU 24, 114) appear in the attention audit of \cref{fig:stats}a but in no end-to-end decision.

\paragraph{Metrics and statistics.}
AUC@$k$ averages, over thresholds $1..k$ degrees, the fraction of frame pairs whose larger rotation or translation-direction error is below the threshold.
AbsRel is the median over frames of the per-frame depth AbsRel after median scaling; Tanks and Temples has COLMAP poses and no depth.
A \emph{loss} is dense minus method, paired by seed; 95\% intervals come from a hierarchical bootstrap over scenes, then seeds (10{,}000 resamples).

\paragraph{Baselines and implementation.}
AVGGT~\cite{sun2025avggt} runs at $\sigma\in\{4,9\}$ in a re-implementation, since no official code is released, and need not reach its published accuracy (App.~\ref{sec:baselines}); the contemporaneous GPA-VGGT~\cite{sung2027gpavggt} runs at its paper defaults, SAF3R~\cite{deng2026saf3r} at its repository default (lazy top-4, shipped head assignment) and FastVGGT~\cite{shen2025fastvggt} at its default (90\% token merging, aimed at ${\sim}10^3$ views).
All methods share checkpoint, precision, frame subsets and software environment (App.~\ref{sec:baselines}), and each baseline but FastVGGT reproduces the dense predictions bit for bit with sparsity disabled (FastVGGT within 0.2 AUC@5).
SAF3R's head assignment was calibrated on 11 ETH3D training scenes that include courtyard, office and seven of the eight unseen low-$\rho$ ETH3D scenes, which favours it on ETH3D.
Development runs use a masked reference implementation of \methodname and the rows of \cref{tab:main} a fused one, whose pooled losses do not differ significantly (App.~\ref{sec:recipe}).

\begin{table*}[t]
\centering
\small
\compactpending
\setlength{\tabcolsep}{5pt}
\caption{\textbf{Main results} on the 21 pre-named test scenes, the frozen recipe's single evaluation ($N{=}32$, scene means of 5 seeds), and on the development scenes. AbsRel $\times100$, without Tanks and Temples (no depth). GFLOP/token at $N{=}32$: exact cost model (reference $T{=}1041$) for VGGT-based recipes, measured kept fractions otherwise. Bold: best sparse value per column (not a significant difference).}
\label{tab:main}
\begin{tabular}{@{}l ccc ccc c c@{}}
\toprule
 & \multicolumn{3}{c}{Unseen low-$\rho$ (13 scenes)} & \multicolumn{3}{c}{Unseen high-$\rho$ (8 scenes)} & Dev.\ (5) & \\
\cmidrule(lr){2-4}\cmidrule(lr){5-7}\cmidrule(lr){8-8}
Method & AUC@5$\uparrow$ & AUC@30$\uparrow$ & AbsRel$\downarrow$ & AUC@5$\uparrow$ & AUC@30$\uparrow$ & AbsRel$\downarrow$ & AUC@5$\uparrow$ & GFLOP/tok.$\downarrow$ \\
\midrule
Dense VGGT-1B & 57.4 & 85.4 & 4.34 & 72.9 & 92.6 & 3.95 & 49.2 & 5.29 \\
\midrule
AVGGT ($\sigma{=}4$)~\cite{sun2025avggt} & 47.1 & 81.1 & 4.08 & 67.4 & 91.4 & 3.87 & 35.2 & 2.64 \\
AVGGT ($\sigma{=}9$)~\cite{sun2025avggt} & 38.5 & 72.8 & 4.32 & 62.4 & 90.5 & \textbf{3.79} & 29.2 & \textbf{2.38} \\
GPA-VGGT~\cite{sung2027gpavggt} & 55.0 & 83.8 & 4.65 & \textbf{72.8} & \textbf{92.6} & 3.93 & 47.8 & 2.96 \\
SAF3R~\cite{deng2026saf3r} & 52.8 & 84.3 & 5.48 & 71.1 & 92.2 & 3.80 & 41.5 & 2.60 \\
FastVGGT~\cite{shen2025fastvggt} & 41.0 & 79.7 & 4.72 & 67.5 & 91.4 & 3.86 & 34.3 & -- \\
\midrule
\oursrow \textbf{\methodname} (ours) & \textbf{56.0} & \textbf{84.7} & 4.68 & 72.1 & 92.4 & 3.95 & 48.6 & 2.99 \\
\methodname-fast (ours) & 52.7 & 83.9 & \textbf{3.97} & 71.1 & 92.1 & 3.95 & 45.3 & 2.49 \\
\bottomrule
\end{tabular}
\end{table*}

\subsection{Main results}
\label{sec:main}

\Cref{tab:main} compares \methodname with dense attention and training-free baselines on scenes unused in the design.
On the 13 unseen low-redundancy scenes, \methodname loses 1.38 AUC@5 (95\% CI [0.71, 2.01]) and 0.70 AUC@30 ([0.17, 1.26]) against dense attention. It has the lowest mean loss of the training-free methods tested.
GPA-VGGT loses 2.42 ([1.28, 3.71]), SAF3R 4.53 ([1.73, 7.73]), AVGGT(4) and AVGGT(9) 10.27 and 18.91, and FastVGGT 16.42.
\methodname gains 8.89 AUC@5 over AVGGT(4) (CI [6.45, 11.97]) and 17.54 over AVGGT(9) (CI [13.16, 23.11]). Its gain over GPA-VGGT is 1.04 (CI [$-0.34$, 2.54]) at 1.01$\times$ its FLOPs.
This misses the pre-registered target of $+5$ over the best baseline (App.~\ref{sec:bars}); at equal cost the difference is not significant.
Only the gains over AVGGT pass both scene-level tests after Holm correction (App.~\ref{sec:perscene}).
Post hoc (App.~\ref{sec:robust}), the advantage over GPA-VGGT lies below $\rho=10.5$ (Spearman $-0.52$, CI $[-0.78, -0.14]$; $-0.48$, $[-0.72, -0.01]$, with $\rho$ from ground truth, under which no low-$\rho$ test scene changes side) and comes mainly from ETH3D, and the worst unseen low-$\rho$ scene costs 2.9 AUC@5, against 6.8 for GPA-VGGT.
On the 8 unseen redundant scenes, GPA-VGGT is more accurate (loss 0.08, CI [$-0.20$, 0.42], against 0.73, [0.25, 1.27]), and \methodname gains 9.77 over AVGGT(9) (CI [3.23, 15.78]).
The routed variant, \methodname-fast, loses 4.65 for 17\% fewer FLOPs.
Post hoc, more hubs (2\% more FLOPs) halve this low-redundancy loss to 0.66 (App.~\ref{sec:variance}; confirmed on fresh data in \cref{sec:speed}).
Depth suffers slightly (AbsRel$\times100$ 4.34 to 4.68 at low redundancy), and frame-local layers 0--8 or AVGGT's late layers raise the low-redundancy loss to 4.84 or 5.54 (App.~\ref{sec:ablation}).

\subsection{Why the matching layers stay exact}
\label{sec:composed}

In a pre-registered design study on the development scenes (Tab.~\ref{tab:composed}, \ref{sec:pilot}), sparsifying layers 11--16 with a $\sigma{=}4$ lattice and routed blocks (\methodname-fast) lost 11.6 AUC@5 on courtyard (2 seeds), above the threshold of 4.
Denser lattices and fewer exact layers each failed on at least one scene; exact layers 11--16 lose 1.10 AUC@5 at low redundancy (CI $[-0.54, 2.86]$).

\begin{finding}[Sparse matching layers]
\label{find:composition}
Every aggressive way of thinning layers 11--16 that was tested costs accuracy on low-redundancy views once the other layers are sparse: truncated lattices with routed blocks cost 3.27 AUC@5 on the test scenes (\methodname-fast against \methodname, CI $[1.34, 5.43]$), merging similar keys costs 1.6 and 5.7 when 50 and 20\% of them remain, per-query top-$k$ frame selection costs 4.2, and shared hubs cost 27.8; only a mild merge that keeps 80\% is not significantly worse (0.5), for little saving (post hoc, App.~\ref{sec:thinmatch}).
\end{finding}

An ablation (App.~\ref{sec:ablation}) shows each part matters: without log-weights the low-$\rho$ test loss is 36 AUC@5, and cut-offs of 2.0 or 3.0 nats cost 2.9 or 11.0 more.
At equal cost, the statistic's layers also beat the four shifted windows tested (registered before it ran; App.~\ref{sec:windows}): windows 9--14, 10--15, 12--17 and 13--18 lose 8.51, 4.19, 2.72 and 4.61 AUC@5 at low redundancy, against 1.38, each significantly more after Holm correction.
Entropy and effective key count pick the same layers, so the choice is not an artefact of the deficit formula (used for its link to normalisation error and its row-free proxy); the six layers with most cross-frame mass lose 27.5 more (App.~\ref{sec:selectors}).

\begin{table}[t]
\centering
\small
\compactpending
\setlength{\tabcolsep}{2.8pt}
\caption{\textbf{Cost, speed and accuracy at scale.} GFLOP per token from the exact cost model (\cref{eq:cost}) at the reference $T{=}1041$; speed-up of the aggregator over dense attention timed in the same session; AUC@5 lost against dense attention on 5 Tanks and Temples scenes at $N{=}100$. Absolute times: Tab.~\ref{tab:cost}. Bold: best sparse value per column.}
\label{tab:costmain}
\begin{tabular}{@{}lcccccc@{}}
\toprule
 & \multicolumn{2}{c}{GFLOP / token$\downarrow$} & \multicolumn{3}{c}{Speed-up vs.\ dense$\uparrow$} & T\&T loss$\downarrow$ \\
\cmidrule(lr){2-3}\cmidrule(lr){4-6}\cmidrule(lr){7-7}
Method & $N{=}32$ & 100 & 32 & 64 & 100 & $N{=}100$ \\
\midrule
Dense & 5.29 & 12.25 & 1.00 & 1.00 & 1.00 & 0 \\
\midrule
AVGGT ($\sigma{=}4$) & 2.64 & 3.77 & 1.54 & 1.99 & 2.41 & 7.10 \\
AVGGT ($\sigma{=}9$) & \textbf{2.38} & \textbf{2.94} & \textbf{1.63} & \textbf{2.22} & \textbf{2.78} & 12.61 \\
GPA-VGGT & 2.96 & 4.79 & 1.40 & 1.73 & 2.00 & \textbf{1.61} \\
SAF3R & 2.60 & 3.67 & 1.34 & 1.80 & 2.19 & 5.65 \\
FastVGGT & -- & -- & 1.22 & 1.58 & 1.89 & 13.96 \\
\midrule
\oursrow \textbf{\methodname} & 2.99 & 4.76 & 1.43 & 1.76 & 2.04 & 1.90 \\
\methodname-fast & 2.49 & 3.14 & 1.51 & 2.00 & 2.50 & 4.55 \\
\bottomrule
\end{tabular}
\end{table}

\subsection{Cost, time and scale}
\label{sec:speed}

\methodname needs 1.8$\times$ fewer FLOPs than dense at $N{=}32$ and 3.1$\times$ fewer at $N{=}200$ (\cref{tab:costmain}); because its six matching layers stay dense, the ratio tends to 4 as $N$ grows.
With every method timed in one session per $N$ (\cref{tab:costmain}), the aggregator of \methodname is about 2\% faster than GPA-VGGT's and takes 7--18\% longer than AVGGT(4)'s, and \methodname-fast is on par with AVGGT(4); end to end, with the heads, \methodname is 2--5\% slower than GPA-VGGT at $N{=}32$--100 (Tab.~\ref{tab:cost}).
Accuracy at $N{=}100$ was measured on five Tanks and Temples scenes (four at low $\rho$), with 5 strided seeds giving only 3--5 distinct subsets (App.~\ref{sec:tt100}).
There \methodname loses 1.90 AUC@5 (CI [1.11, 2.81]) and 0.34 AUC@30 against dense attention while running 2.0$\times$ faster on the aggregator; AVGGT(4) and AVGGT(9) lose 7.10 and 12.61, so \methodname gains 5.20 over AVGGT(4) (CI [2.88, 8.35]) at an indicative 1.14$\times$ its time (\cref{tab:costmain}).
Post hoc, GPA-VGGT loses 1.61 on the same runs, not significantly different from \methodname (CI of the difference [$-2.00$, 1.77]), and SAF3R 5.65.
On long, redundant 7-Scenes video (App.~\ref{sec:longseq}), \methodname loses 0.60 AUC@5 (CI [$-0.27$, 1.87]) with every third frame ($N{=}333$, 7 sequences) at 2.7$\times$ the dense speed, where GPA-VGGT exceeds dense by 0.82; with 800 frames it stays within 1.6 AUC@5 of dense on each of 4 sequences at 3.3$\times$, whereas AVGGT, at 5.1--8.1$\times$, departs from it by up to 7.9.
In a confirmatory test registered before it ran, on NRGBD, Sintel and these sequences (none used in design), the variant with more hubs stays on average within 0.41 AUC@5 of dense attention and closer to it than \methodname, and one merging 20\% of the keys in layers 11--16 stays within 1 AUC@5 of \methodname at 1.16$\times$ its speed for $N{=}333$ (App.~\ref{sec:confirm}).
On 9 outdoor RealEstate10K clips~\cite{zhou2018stereo} (post hoc; all redundant) \methodname loses 0.22 AUC@5, not significantly different from GPA-VGGT, against 2.11, 5.14 and 2.49 for AVGGT(4), AVGGT(9) and SAF3R (App.~\ref{sec:re10k}).

\subsection{Transfer to a second architecture}
\label{sec:xarch}

A dense pilot pass of the original \pithree~\cite{wang2025pi3} on the development scenes labelled its 18 global layers with the VGGT cut-offs before any sparse run: layers 11--15 and, only just ($\Gbar=1.54$), layer 7 are matching.
With these layers exact and hubs elsewhere, the rule loses 4.87 AUC@5 against dense \pithree on the 13 unseen low-$\rho$ scenes (CI [2.45, 7.51], 2 seeds each); AVGGT(4) loses 13.81 (gain 8.93, CI [4.07, 13.95]).
The transfer is thus preliminary: the rule beats AVGGT but misses the pre-registered target of a loss within 3 AUC@5 on the unseen scenes, and a pre-registered group-level test failed on the borderline layer 7 (App.~\ref{sec:pi3}); GPA-VGGT and SAF3R were not run on \pithree. Layer 7, which the row-free proxy does not select, loses nothing with hubs, so a statistic just above the cut-off is not sufficient.

\section{Limitations and Conclusion}
\label{sec:limitations}

Efficient attention for 3D reconstruction must preserve each layer's information, not merely attention mass. Probability-corrected hubs preserve diffuse context; no tested aggressive thinning preserved the few high-deficit matching layers' query-specific interactions.

\methodname is not significantly more accurate than contemporaneous GPA-VGGT. At equal FLOPs, its low-redundancy margin (1.0 AUC@5) misses the pre-registered $+5$; GPA-VGGT is better on redundant views.
\methodname is about as fast as GPA-VGGT and takes 13--18\% longer than AVGGT(4) at $N{=}64$--100; its six exact layers bound its FLOP advantage over dense attention by 4$\times$ (App.~\ref{sec:broad}).
More hubs reduce run-to-run hub-sampling noise; \pithree transfer is preliminary.

\section*{LLM Usage Statement}
Large language models were used in preparing this paper, under the authors’ supervision, for two purposes, both declared in the submission form. The tool used was ChatGPT (OpenAI). First, drafting: some passages were generated from outlines, derivations and result tables supplied by the authors, and then edited by the authors. Second, editing: existing prose was revised for concision, consistent terminology and clarity. The authors read, corrected and approved every generated or edited passage before including it. The authors developed the research problem, conceptual framework, hypotheses, theoretical results and benchmark design before any generative AI was used. AI was not used to originate ideas or hypotheses, to find or select related work, to derive or check proofs, or to produce any reported result. Every citation was checked against its source, and every reported number comes from the reported experiments. The authors take full responsibility for the manuscript and all its claims, results and artifacts.

{\small
\bibliographystyle{ieeenat_fullname}
\bibliography{refs}}

\clearpage
\appendix
\setcounter{table}{0}\setcounter{figure}{0}\setcounter{equation}{0}
\renewcommand{\thetable}{S\arabic{table}}
\renewcommand{\thefigure}{S\arabic{figure}}
\renewcommand{\theequation}{S\arabic{equation}}
\section*{Appendices}

\renewcommand{\thetable}{S\arabic{table}}
\renewcommand{\thefigure}{S\arabic{figure}}
\renewcommand{\theequation}{S\arabic{equation}}

This supplement has six parts: method details, the evaluation protocol, detailed results, the evidence behind the design, the validation of the layer selector, and robustness.
It starts with the derivations behind \cref{sec:method} and then gives the implementation and evaluation details needed to reproduce the results.

\section{Method details}
\subsection{Derivations}
\label{sec:derivations}

These derivations prove the identity for the mean-token deficit $\Gamma$, show which properties the hub estimator has, and state where both results stop applying.
Each derivation is set out in numbered steps, one relation per line, with the reason for a step in grey on its right.

\dstep{Colour key.}
\exq{Blue}: exact (kept) keys; \hubq{orange}: stand-ins and hubs; \gq{magenta}: the mean-token deficit $\Gamma$; \sq{green}: the row-free logit variance $s^2$.

\dstep{Notation.}
Fix one query $t$, one head and one layer, and drop $t$ from the notation.
For a key $u$ with value $v_u$ and a key set $X$,
\begin{align*}
  w_u &= e^{l_u} \why{($l_u$: logit of $u$)}\\
  Z_X &= \textstyle\sum_{u\in X} w_u \why{(mass of $X$)}\\
  A_X &= \textstyle\sum_{u\in X} w_u v_u \why{(numerator of $X$)}\\
  m_X &= A_X/Z_X \why{(weighted mean value)}
\end{align*}

\subsubsection{Proof of Proposition~\ref{prop:gamma}}
\label{sec:proof-gamma}

\dstep{Step 1: a cumulant generating function.}
For $\tau\in[0,1]$, define the log-mean of the tilted weights and the tilted distribution on $D$:
\begin{align*}
  \Phi(\tau) &= \ln\tfrac{1}{|D|}\textstyle\sum_{u\in D} e^{\tau l_u},\\
  p_\tau(u) &= e^{\tau l_u}\big/\textstyle\sum_{u'\in D} e^{\tau l_{u'}}.
\end{align*}
At the two ends of the interval,
\begin{align*}
  \Phi(0) &= 0,\\
  \Phi(1) &= \ln Z_D-\ln|D|.
\end{align*}

\dstep{Step 2: its derivatives.}
Differentiating,
\begin{align*}
  \Phi'(\tau)  &= \mathbb E_{p_\tau}[l],\\
  \Phi''(\tau) &= \Var_{p_\tau}(l)\ \ge 0.
\end{align*}
At $\tau=0$ the distribution $p_0$ is uniform on $D$, so
\begin{align*}
  \Phi'(0) &= \bar l_D \why{(uniform mean logit)}\\
           &= \langle q,\bar k_D\rangle/\sqrt{d_h}. \why{(logit is linear in the key)}
\end{align*}

\dstep{Step 3: $\Gamma$ is a Taylor remainder.}
By the definition in Eq.~\eqref{eq:gamma}, for some $\xi\in(0,1)$,
\begin{align*}
  \gq{\Gamma} &= \Phi(1)-\Phi(0)-\Phi'(0) \why{(Eq.~\eqref{eq:gamma})}\\
              &= \tfrac12\,\Phi''(\xi) \why{(Lagrange remainder)}\\
              &= \tfrac12\,\Var_{p_\xi}(l).
\end{align*}

\dstep{Step 4: sign and equality.}
Hence $\gq{\Gamma}\ge0$, and $\gq{\Gamma}=0$ iff $\Var_{p_\xi}(l)=0$.
Since $p_\xi$ has full support on $D$, this happens iff $l$ is constant on $D$.

\dstep{Step 5: the row-free approximation.}
With $\Sigma_K$ the uniform covariance of the keys in $D$,
\begin{align*}
  \Phi''(0) &= \Var_{p_0}(l)\\
            &= \tfrac{1}{|D|}\textstyle\sum_{u\in D}(l_u-\bar l_D)^2\\
            &= q^\top\Sigma_K q/d_h \;=\; \sq{s^2}.
\end{align*}
If the logit spread is small, $\Phi''$ varies little on $[0,1]$, so
\begin{equation*}
  \gq{\Gamma}\;\approx\;\tfrac12\,\Phi''(0)\;=\;\sq{s^2}/2 .
\end{equation*}

\dstep{Step 6: a range bound.}
Hoeffding's lemma gives, with $\Delta=\max_Dl-\min_Dl$ the logit range,
\begin{equation*}
  0\;\le\;\gq{\Gamma}\;\le\;\Delta^2/8 . \tag*{$\blacksquare$}
\end{equation*}

\subsubsection{The random-energy model}
\label{sec:rem}

\dstep{Setup.}
Let the $n$ logits be $l_u=\bar l+s\,g_u$, with $g_u$ i.i.d.\ standard normal.
Then the density of keys, the largest key and the mass are
\begin{align*}
  \#\{u: g_u\in[x,x+dx]\} &\approx n\,e^{-x^2/2}\,dx/\sqrt{2\pi},\\
  x_{\max} &\approx \sqrt{2\ln n}, \why{(no keys beyond)}\\
  Z &\approx e^{\bar l}\!\int^{x_{\max}}\! n\,e^{-x^2/2+sx}\,\tfrac{dx}{\sqrt{2\pi}} .
\end{align*}
The exponent peaks at $x=s$.

\dstep{Annealed (high-temperature) phase, $s<\sqrt{2\ln n}$.}
The peak lies inside the range, and $Z$ is dominated by the $\approx n\,e^{-s^2/2}$ keys near $x=s$, which are many:
\begin{align*}
  \ln Z &\approx \ln n+\bar l+\sq{s^2}/2,\\
  \gq{\Gamma} &= \sq{s^2}/2 .
\end{align*}

\dstep{Frozen phase, $s\ge\sqrt{2\ln n}$.}
The peak lies beyond the largest key, and $Z$ is dominated by the keys at the edge:
\begin{align*}
  \ln Z &\approx \bar l+s\sqrt{2\ln n},\\
  \gq{\Gamma} &= s\sqrt{2\ln n}-\ln n .
\end{align*}
The second-cumulant formula then overshoots by
\begin{align*}
  &\ln n+\sq{s^2}/2-s\sqrt{2\ln n}\\
  &\quad= (s-\sqrt{2\ln n})^2/2 .
\end{align*}

\dstep{Sample size.}
For a uniform sample of $m$ keys in the annealed phase, the relative variance of the estimate of $Z$ is $(e^{s^2}-1)/m$.
This suggests $m\propto e^{s^2}$, but the saved 256-sample estimators show no such dependence (\cref{sec:limitations}).
The REM therefore serves only to classify layers, not to size samples.

\subsubsection{Proof of Proposition~\ref{prop:hubs} (unbiased hubs)}
\label{sec:proof-hubs}
\label{sec:hub-proof}

\dstep{Step 1: sampling.}
Let $p$ be the proposal on the patch tokens $\mathcal C$, and let $c_j$ count how often $j$ appears among $m$ i.i.d.\ draws, so
\begin{equation*}
  \mathbb E[c_j]=m\,p_j .
\end{equation*}

\dstep{Step 2: the hub weights.}
For a query in frame $i$, the hub part of Eq.~\eqref{eq:sparse} uses
\begin{equation*}
  \hubq{e^{b_j}}=\frac{c_j}{m\,p_j},\qquad j\in\hubq{\mathcal H}\setminus\mathcal F_i .
\end{equation*}
Draws that land in the query's own frame are discarded, because that frame is already exact.

\dstep{Step 3: unbiased sums.}
For any $f$ (the Hansen--Hurwitz estimator),
\begin{align*}
  \mathbb E\Big[\sum_{j\in\mathcal C\setminus\mathcal F_i}\frac{c_j}{mp_j}f(j)\Big]
   &= \sum_{j\in\mathcal C\setminus\mathcal F_i}\frac{\mathbb E[c_j]}{mp_j}f(j)\\
   &= \sum_{j\in\mathcal C\setminus\mathcal F_i}f(j).
\end{align*}
With $f(j)=w_j$ and $f(j)=w_jv_j$, both estimates are unbiased:
\begin{align*}
  \mathbb E[\hubq{\hat Z_D}] &= Z_D, \why{(dropped mass)}\\
  \mathbb E[\hubq{\hat A_D}] &= A_D. \why{(dropped numerator)}
\end{align*}

\dstep{Step 4: the output ratio.}
The output is a ratio of two unbiased estimates, so its bias is $O(1/m)$ and its variance is $O(1/m)$, with a constant that falls as $p$ approaches the attention row.
The column-mass proposal approaches this ideal for all queries at once: it is the attention row averaged over pilot queries.

\dstep{Contrast: the mean token.}
The mean token instead has, deterministically,
\begin{equation*}
  \hat Z_D = Z_D\,e^{-\gq{\Gamma}},
\end{equation*}
and this bias does not shrink with any sample size.

\dstep{Real-data check (post hoc, registered before it ran).}
The proposition was tested directly on VGGT-1B activations: 5 development scenes $\times$ 2 seeds, 2048 queries, all 16 heads and all 18 hub layers, with 131{,}072 hub draws per layer and $m$ from 64 to 2048.
\begin{itemize}
  \item \textbf{Unbiased sums.} In every layer group the mean of $\hubq{\hat Z_D}/Z_D-1$ is within $1.6\times10^{-5}$ of 0, and every 95\% CI covers 0; the CIs of the numerator also cover 0.
  \item \textbf{Variance.} The output variance falls as $1/m$: log--log slopes $-0.99$, $-0.97$ and $-1.00$ for layers 0--8, 9--10 and 17--23.
  \item \textbf{Ratio bias.} At the frozen $m$ it is at most 0.11\% of the output's mean squared error, so the hub error is almost entirely variance.
  \item \textbf{A prediction not supported as registered.} The squared bias was predicted to fall as $1/m^2$. In layers 9--10 its slope is $-1.42$, and in the other groups the bias is too small to resolve beyond $m=128$ under the registered aggregation.
\end{itemize}
In a post hoc check, the leading-order constant of the $O(1/m)$ bias matches its analytic value within 1\% in layers 0--8 and 17--23; in layers 9--10, which approach that regime more slowly, it reaches 0.85 of the analytic value.
Because the hub error is almost entirely variance, and that variance falls as the number of draws grows, drawing more hubs, not changing the proposal, is the effective way to reduce it (\cref{sec:variance}).

\subsubsection{Error identity and the inflation factor}
\label{sec:identity}

\dstep{Setting.}
Let $\mathcal U=\exq{\mathcal K}\,\dot\cup\,D$, and let an estimator replace $(Z_D,A_D)$ by $(\hubq{\hat Z_D},\hubq{\hat A_D})$ with $\hat Z_D>0$.
Write
\begin{align*}
  \hat m_D &= \hat A_D/\hat Z_D, \why{(stand-ins' mean value)}\\
  \hat\varphi &= \hat Z_D/(Z_{\mathcal K}+\hat Z_D), \why{(their mass share)}\\
  \gamma &= \ln(\hat Z_D/Z_D). \why{(their log-mass error)}
\end{align*}

\dstep{Claim.}
The output error splits into a mean-value error and a mass error:
\begin{equation}
  \hat o-o=\hat\varphi\Big[(\hat m_D-m_D)+(1-e^{-\gamma})(m_D-o)\Big].
  \label{eq:identity}
\end{equation}

\dstep{Proof.}
Let $\hat Z=Z_{\mathcal K}+\hat Z_D$. Then
\begin{align*}
  \hat Z\,\hat o &= A_{\mathcal K}+\hat Z_D\,\hat m_D,\\
  \hat Z\,o &= A_{\mathcal K}+Z_D\,m_D+(\hat Z_D-Z_D)\,o.
\end{align*}
Subtracting the two lines,
\begin{align*}
  \hat Z(\hat o-o) &= \hat Z_D(\hat m_D-m_D)\\
                   &\quad+(\hat Z_D-Z_D)(m_D-o),
\end{align*}
and dividing by $\hat Z$ gives \cref{eq:identity}.\hfill$\blacksquare$

\dstep{Reading.}
The first bracket term is the error of the stand-ins' mean value; the second is the error of their mass.
For the mean token $\gamma=-\gq{\Gamma}$, so the mass error is
\begin{equation*}
  (1-e^{\gq{\Gamma}})(m_D-o),
\end{equation*}
which vanishes only if $\gq{\Gamma}=0$.
Every kept key's weight is multiplied by
\begin{equation*}
  \kappa=\frac{Z}{\hat Z}=\frac{1}{(1-\varphi)+\varphi\,e^{-\gq{\Gamma}}},
\end{equation*}
which is Eq.~\eqref{eq:inflation}.
Truncation ($\hat Z_D=0$) is the limit $\gq{\Gamma}\to\infty$, with $\kappa=1/(1-\varphi)$ and $\hat o=m_{\mathcal K}$.

\subsection{Full recipe}
\label{sec:recipe}

This section specifies the frozen recipe, that is, the exact configuration fixed before the main evaluation, and how it is implemented.
Development runs use a masked reference implementation of \methodname and its rows in \cref{tab:main} a fused one, whose AUC@5 loss differs from the reference by $-0.38$ (CI $[-1.45, 0.84]$) pooled over the development scenes but by 1.78 per run.

\begin{table*}[t]
\centering
\small
\caption{\textbf{\methodname on VGGT-1B} (frozen definition) and \methodname-fast. $T{=}1041$ tokens per frame at 518\,px, 16 heads, $d_h{=}64$. Units at $N{=}32$ from the exact cost model (one unit = one frame's keys for every query). Bold: the layers kept exact.}
\label{tab:recipe}
\begin{tabular}{@{}p{1.9cm}p{11.4cm}c@{}}
\toprule
\headrow Layers & Key set per query (query in frame $i$) & Units \\
\midrule
0--8 & Own frame (all tokens) + special tokens of every frame + 256 hubs importance-sampled from a 512-query pilot (head-mean full attention, column mass, floor $10^{-3}/|\mathcal C|$), log-weight $\ln(c_j/(mp_j))$; hubs in the own frame get no weight & 1.6 \\
9, 10 & Own frame + special tokens + 1024 hubs (same estimator) & 2.4 \\
11--16 & \textbf{Exact dense attention} (the layers with $\Gbar_\ell\ge1.5$) & $N$ \\
17--23 & As layers 9--10 & 2.4 \\
\midrule
\methodname-fast, 11--16 & \cref{eq:band}: frame 0 + special tokens + $\sigma{=}4$ lattice + diagonal + mean token + switched $6{\times}10$ blocks & 9--10 \\
\methodname-fast, router & Once, on layer 9's attention input; top-4 cosine per (query, frame) at full width; $4{\times}8$ tiles; $3{\times}3$-smoothed displacement mode; switch: share of the tile's top-1 votes in the smoothed peak $\ge0.3$ & $\approx0.5N$ \\
\bottomrule
\end{tabular}
\end{table*}

\Cref{tab:recipe} gives both recipes.
The random generator of the pilot for layer $\ell$ uses the seed $100003(\ell+1)+m$, so every run can be reproduced exactly.
The frozen recipe has no router, no correspondence blocks and no lattice-density controller; \methodname-fast, in contrast, uses a router and routed correspondence blocks.
\paragraph{Implementation}.
The hubs $\mathcal H$ are the same for every query, so a hub layer needs no per-query gathering of keys.
Instead, it runs one dense attention call per frame over the query's own frame $\mathcal F_i$ and one call, shared by all queries, over the special tokens and hubs $[\mathcal S;\mathcal H]$; the two partial results are then merged by log-sum-exp.
A hub layer costs $1+5(N{-}1)/T+(m+256)/T$ units, where the four terms are the own frame, the special tokens, the hubs and the pilot.
The pilot term counts one pass of query--key products for the 512 pilot queries, $512/(2T)=256/T$ units.
The fused implementation computes these products twice (once for the softmax normaliser and once for the column mass), which adds another $256/T$ units per hub layer, about 0.02 GFLOP per token over the 18 hub layers, under 1\% of the total.
The GFLOP figures in the paper use the single-pass count; the measured times include the actual cost.
Only the special-token term grows with $N$, and it is small, so the cost of a hub layer stays nearly constant as more frames are added.
The test rows and all timings use a fused implementation.
Per frame, it issues one FlashAttention~\cite{dao2022flashattention} call over the own frame and one shared Triton kernel over the special tokens and hubs.
It matches the masked reference implementation to within $1.5\times10^{-5}$ per layer in fp32, so the two implementations give the same model.
For models without a reference frame, a frame-0 rule was also defined: keep a frame dense in the matching layers if its pilot column-mass share exceeds $1.5\times$ the uniform share.
This rule is only logged, because it has no effect when the matching layers are already exact.

\subsection{Per-layer statistics of VGGT-1B}
\label{sec:layers}

This section reports, for every layer of the model, the statistics that the selector uses or that help to interpret its choice.

\begin{table}[t]
\centering
\footnotesize
\setlength{\tabcolsep}{4.2pt}
\renewcommand{\arraystretch}{1.12}
\caption{\textbf{Per-layer statistics} behind \cref{fig:stats}. $\Gbar$: pilot statistic (median head, mean of 5 development scenes). $s^2$: median-head row-free logit variance. ``$<3$'': heads (of 16) with $s^2<3$. ``$\ge7$'': share of (query, head) pairs with $s^2\ge7$. Tube: attention mass on the $3{\times}3$ correspondence window (low-$\rho$). Layers 0--8 were not in the attention audit. Bold: the layers kept exact (those above the cut-off) and their $\Gbar$.}
\label{tab:layers}
\begin{tabular}{@{}rccccc l@{}}
\toprule
\headrow Layer & $\Gbar$ & $s^2$ & $<3$ & $\ge7$ & Tube & Role \\
\midrule
0 & 0.12 & -- & -- & -- & -- & hubs, 256 \\
1 & 0.05 & -- & -- & -- & -- & hubs, 256 \\
2 & 0.30 & -- & -- & -- & -- & hubs, 256 \\
3 & 0.71 & -- & -- & -- & -- & hubs, 256 \\
4 & 0.78 & -- & -- & -- & -- & hubs, 256 \\
5 & 0.86 & -- & -- & -- & -- & hubs, 256 \\
6 & 0.92 & -- & -- & -- & -- & hubs, 256 \\
7 & 0.91 & -- & -- & -- & -- & hubs, 256 \\
8 & 0.25 & -- & -- & -- & -- & hubs, 256 \\
9 & 0.78 & 1.28 & 16 & 0.00 & -- & hubs, 1024 \\
10 & 1.00 & 1.57 & 15 & 0.00 & 0.006 & hubs, 1024 \\
11 & \textbf{2.66} & 4.92 & 4 & 0.29 & 0.029 & \textbf{exact} \\
12 & \textbf{2.76} & 6.82 & 4 & 0.39 & 0.030 & \textbf{exact} \\
13 & \textbf{2.82} & 6.59 & 4 & 0.47 & 0.050 & \textbf{exact} \\
14 & \textbf{9.21} & 26.64 & 3 & 0.73 & 0.098 & \textbf{exact} \\
15 & \textbf{4.08} & 8.32 & 3 & 0.49 & 0.038 & \textbf{exact} \\
16 & \textbf{1.83} & 3.15 & 6 & 0.04 & 0.008 & \textbf{exact} \\
17 & 0.53 & 0.88 & 16 & 0.00 & -- & hubs, 1024 \\
18 & 0.03 & 0.05 & 16 & 0.00 & -- & hubs, 1024 \\
19 & 0.28 & 0.44 & 16 & 0.00 & -- & hubs, 1024 \\
20 & 0.06 & 0.11 & 16 & 0.00 & -- & hubs, 1024 \\
21 & 0.48 & 1.04 & 11 & 0.12 & -- & hubs, 1024 \\
22 & 0.01 & 0.03 & 16 & 0.00 & -- & hubs, 1024 \\
23 & 0.07 & 0.12 & 16 & 0.00 & -- & hubs, 1024 \\
\bottomrule
\end{tabular}
\end{table}

\Cref{tab:layers} lists the statistics per layer.
Three observations in it matter for the method.
First, the two forms of the selector agree on layers 9--23: the pilot rule $\Gbar\ge1.5$ and the row-free rule (median-head $s^2\ge3$) both pick layers 11--16.
This also explains why \cref{tab:factorial} and \cref{fig:stats}c treat layers 10--16 as the middle group: those experiments were run before the statistic existed, and the statistic then classified layer 10 as diffuse ($\Gbar=1.00$), so the method gives it 1024 hubs rather than exact attention.
Second, layer 21 is diffuse at the median head, but 5 of its 16 heads have $s^2\ge3$, so a few of its heads are concentrated.
This fits two earlier observations: this layer needs 1024 rather than 256 hubs, and the corrected mean token alone still left a 5.4-point loss on TUM fr3.
Third, the correspondence window carries at most 10\% of the attention mass, even at layer 14, and yet removing it is catastrophic (\cref{tab:knockout}).
Attention mass is therefore the wrong criterion for choosing which keys to keep exact: keys that receive little mass can still carry the information the layer uses.

\section{Protocol}
\subsection{Reproducibility details}
\label{sec:repro}

This section documents the registered protocol: how frame subsets are drawn and ordered, how images and ground truth are prepared, and how the redundancy $\rho$ is computed.

\emph{Frame subsets.} Each seed $s$ selects a different subset of frames from a scene.
For TUM, 7-Scenes and Tanks and Temples, seed $s$ takes $N{=}32$ frames at a fixed stride, starting at frame $\mathrm{round}(s\cdot\text{stride}/5)$.
The stride is 20 for TUM (or the largest stride that fits; only frames with depth and pose within 20\,ms are used), 15 for 7-Scenes sequence 1, and $\lfloor n/32\rfloor$ over the $n$ registered Tanks and Temples images in filename order.
For ETH3D and DTU, seed $s$ draws views without replacement with NumPy \texttt{default\_rng}([20261111, $s$, $\sum$ord(scene key)]).
The drawn view nearest the subset's camera centroid becomes frame 0, and the other views follow in order of camera distance.
The exceptions are seed 0 of courtyard and the DTU scans, which take the 32 views nearest the view closest to the centroid of all cameras.
The ground-truth cameras serve only to build and order these benchmark subsets.
They are never an input to any model, and every method receives the same images in the same order, so dense attention and all sparse methods share the same frame 0.
For TUM, 7-Scenes and Tanks and Temples, the order is the recording or file order, which uses no pose information.

\emph{Random input order (post hoc control, registered before it ran).}
The ETH3D subsets are ordered using camera geometry, so the reference frame is a central view.
This control tests whether that geometry-informed order favours any method.
The 8 unseen low-$\rho$ ETH3D test scenes were rerun with the same views in a random order, so that frame 0 is an arbitrary view (5 seeds, 40 stored permutations), for dense attention, \methodname, GPA-VGGT and AVGGT(4).
Each loss is measured against dense attention run on the same order (\cref{tab:order}).
The loss of \methodname changes by only $+0.08$ AUC@5 (CI [$-1.05$, 1.21]), which lies inside the registered band of $\pm1.5$, and the loss of GPA-VGGT changes by $-0.19$.
Only AVGGT(4) degrades, by 3.48 AUC@30 (CI [0.24, 7.34]); this is expected, because this method keeps all of frame 0 in its later layers and so depends on which view comes first.
As a result, under the random order the lead of \methodname over AVGGT(4) grows (from 8.87 to 10.66 AUC@5), while its lead over GPA-VGGT changes little (from 1.44 to 1.17).
One deviation from the plan occurred.
One run of GPA-VGGT on the original order differed from its main-table run by up to 0.53$^\circ$, slightly beyond the planned parity bound of 0.5$^\circ$.
A registered amendment showed that this difference matches GPA-VGGT's own run-to-run spread, which arises because its merging uses atomic operations; its rows are therefore kept and marked as a deviation.
Scoring the random-order predictions in the original frame order gives the same picture (\methodname $+0.15$, GPA-VGGT $+0.11$, AVGGT(4) $+2.44$ AUC@5).

\begin{table}[t]
\centering
\small
\setlength{\tabcolsep}{1.40pt}
\caption{\textbf{Random input order} on the 8 unseen low-$\rho$ ETH3D scenes (5 seeds): AUC@5 lost against dense attention on the same order, mean and 95\% CI; change = random minus original. Bold: lowest loss per column and the change for \methodname.}
\label{tab:order}
\begin{tabular}{@{}lccc@{}}
\toprule
Method & Original order & Random order & Change \\
\midrule
\methodname & \textbf{1.57} & \textbf{1.65} & $\mathbf{+0.08}$ [$-1.05$, 1.21] \\
GPA-VGGT$^\dagger$ & 3.01 & 2.83 & $-0.19$ [$-2.08$, 1.35] \\
AVGGT(4) & 10.44 & 12.31 & $+1.87$ [$-1.02$, 4.95] \\
\bottomrule
\end{tabular}
\\[2pt]{\scriptsize $^\dagger$Rows kept under a registered parity amendment (see text).}
\end{table}
In every subset, VGGT's reference frame is frame 0.
\emph{Preprocessing.} Images are resized bicubically to a width of 518\,px and a height rounded to a multiple of 14, and the camera intrinsics are rescaled per axis to match.
The resulting sizes are 392$\times$518 for TUM, 7-Scenes and DTU (so $T{=}1041$ tokens per frame), 350$\times$518 for ETH3D and 294$\times$518 for Tanks and Temples.
Ground-truth depth is brought to the model's resolution without hole filling.
TUM and 7-Scenes depth is resized by nearest neighbour; DTU depth comes from the 160$\times$128 MVSNet depth maps; ETH3D depth is a per-pixel minimum-depth splat of the aligned laser scans.
Tanks and Temples has COLMAP poses but no depth.
\emph{Redundancy.} The redundancy $\rho$ measures how many views see each surface patch.
To compute it, every patch of every view is back-projected from its centre, at the median predicted depth of its pixels, using the dense model's cameras.
The patch counts as visible in another view if it lands in front of that camera, inside the image, and within 15\% of the median predicted depth of the surrounding 5$\times$5 pixels.
No confidence threshold is applied.
\emph{Selector for a new model.} The selector computes $\Gamma$ on 512 pilot rows, which are uniformly random patch queries drawn per scene and subset.
The same rows are used in every layer, and they are separate from the hub pilot.
$\Gamma$ is computed in fp32 from the post-RoPE queries and keys.
Its median over heads is averaged over rows and then over the pilot scene-subsets, which are the five development scenes (on VGGT, subsets 0--4 for the low-$\rho$ scenes and 0--1 for the others; on \pithree, subsets 0--2, after a parity check of the masked pass).
\emph{Finding~\ref{find:late}.} The corrected arm of this finding keeps, in layers 9 and 17--23, the keys of the uncorrected arm (frame 0, special tokens, $\sigma{=}9$ lattice and self term) and its per-query mean token.
It then corrects the mean token in two ways: it adds $\min(\tfrac12 s^2,3)$ to the token's logit, and it adds the first-order term $\mathrm{Cov}_D(v,l)$ to the token's value.
The covariances are computed per layer and head over the patches that lie off the lattice and outside frame 0.
Code will be released.

\subsection{Scenes}
\label{sec:scenes}

This section lists the development and test scenes used in the registered evaluation.

All lists were fixed before the composed method was first run, based on a redundancy scan of 70 configurations.
\emph{Unseen low-$\rho$ (13):} ETH3D~\cite{schops2017eth3d} delivery\_area, electro, kicker, playground, relief, relief\_2, terrace and terrains; Tanks and Temples~\cite{knapitsch2017tanks} Barn, Caterpillar, Courthouse and Meetingroom (full span); TUM~\cite{sturm2012tum} fr1\_plant.
\emph{Unseen high-$\rho$ (8):} TUM fr2\_desk and fr3\_structure\_texture\_far; 7-Scenes~\cite{shotton2013scene} chess, fire and pumpkin; DTU~\cite{aanaes2016dtu} scans 24 and 114; Tanks and Temples Ignatius.
\emph{Development (5):} TUM fr1\_room, ETH3D courtyard and office; TUM fr3\_long\_office and 7-Scenes office.
\emph{Held out (10; used only to fit a threshold of an earlier, routed recipe):} ETH3D facade, pipes and meadow; TUM fr1\_teddy, fr1\_desk and fr3\_cabinet; 7-Scenes heads, redkitchen and stairs; Tanks and Temples Truck.
The Tanks and Temples low-$\rho$ scenes have low redundancy by this count, but they are easy for the model: dense VGGT already reaches 90--91 AUC@5 on Barn, Caterpillar and Meetingroom (5-seed means).
Differences between methods are therefore expected to be small on these scenes.
ETH3D scenes with fewer than 36 images use $n{-}4$ of their $n$ images (office 22 of 26; kicker, relief and relief\_2 27 of 31; terrace 19 of 23).

\subsection{Baselines}
\label{sec:baselines}

This section documents how each baseline was implemented, where the setup deviates from the authors' own, and which methods were not run.

\paragraph{AVGGT.}
No official code is released for AVGGT~\cite{sun2025avggt}, so a re-implementation is used.
It makes layers 0--8 frame-local, meaning that each query attends only to its own frame.
In layers 9--23, each query keeps the special tokens, frame 0 and one patch per $\sigma$-cell ($\sigma{=}4$: $2{\times}2$; $\sigma{=}9$: $3{\times}3$), plus a self column for dropped queries; all dropped tokens are replaced by one count-biased mean key.
The constants follow the AVGGT paper: frame-local layers 0--8; the first patch of each $2{\times}2$ ($\sigma{=}4$) or $3{\times}3$ ($\sigma{=}9$) window; frame 0 uncompressed; all queries and special tokens kept; and the self term plus one mean key and value for the dropped patches, all under a shared softmax.
The only constant that the AVGGT paper does not state is the $+\ln n_{\text{dropped}}$ bias on the mean key.
This re-implementation does not show AVGGT at its best.
At $N{=}32$ it loses 5.5 AUC@5 on the unseen redundant scenes (on TUM fr2\_desk, from 88.9 to 74.5).
In contrast, the AVGGT paper reports accuracy close to dense VGGT at 333 and 800 frames on its own benchmarks (\eg, 7-Scenes AUC@30 77.4 against 74.2 at 800 frames)~\cite{sun2025avggt}.
The likely causes are the smaller $N$ used here, the absence of official code and the unstated constant above.
\paragraph{GPA-VGGT.}
GPA-VGGT~\cite{sung2027gpavggt} is contemporaneous work (arXiv, 27 October 2026); it is run with its paper defaults.
In global blocks 0--8, each query keeps its own frame, all special tokens and 20 cluster summaries per frame and head, with a $+\log(\text{size})$ bias that accounts for how many tokens each summary stands for.
In blocks 9--23, it keeps frame 0, the special tokens and one random token per $2{\times}2$ window, and merges every other token into its most similar kept token.
Blocks 13--16 are treated more gently: they merge only 20\% of their tokens.
\paragraph{SAF3R.}
SAF3R~\cite{deng2026saf3r} is run with its repository default for VGGT.
Each of the 384 global heads is assigned one fixed pattern: frame 0 only, strided with anchors, query-probe top-$k$, own frame only, or per-frame top-4.
The assignment comes from an offline calibration on 11 ETH3D training scenes, as listed in its repository's calibration configuration (the SAF3R paper reports 10 sequences).
\paragraph{FastVGGT.}
FastVGGT~\cite{shen2025fastvggt} is run with its repository defaults (official code, commit 6526e27) and no tuning.
It merges tokens in all 24 global blocks with merge ratio 0.9, which means that 90\% of the tokens of each block, queries included, are merged into kept tokens.
The other defaults are a $2{\times}2$ destination stride, protected tokens enabled and merge seed 33.
FastVGGT was added after the main results, under a plan registered before its first run.
It is designed for about $10^3$ views, and its authors evaluate it at that scale, so its rows here, at $N{=}32$ and 100, need not reflect its best case.
\paragraph{Not run.}
Other training-free accelerators were not run in the time available, among them RegimeVGGT~\cite{you2026regimevggt}, whose code is public.
Their published speed-ups and accuracies use other checkpoints, hardware, sequence lengths and timing conventions, so they cannot be compared with \cref{tab:cost} or \cref{tab:main}.
\paragraph{Deviations from the authors' setups.}
The GPA-VGGT and SAF3R repositories default to the tracker-fixed VGGT checkpoint, and the FastVGGT authors also evaluate on it; here, in contrast, VGGT-1B is used for all methods.
FastVGGT casts its whole model to bf16 and normalises its input in bf16.
Here it runs at the precision shared by all methods, but keeps its own normalisation and its per-block cache release.
As a result, unlike the other baselines, FastVGGT with merging disabled does not reproduce dense attention bit for bit; it stays within 0.2 AUC@5 and 0.3 AUC@30 of it on TUM fr1\_room (largest rotation difference 0.33$^\circ$).
GPA-VGGT also runs at the precision shared by all methods here (fp32 weights, bf16 autocast in the aggregator, fp32 heads) rather than in fp16.
On TUM fr1\_room this changes its AUC@5 by 0.37 (single seed).
All rows of \cref{tab:main} run in one software environment.
This matters because dense predictions differ by up to about 1 AUC@5 between the two environments used, so mixing environments would add a spurious difference between methods.
Dense attention and SAF3R give bit-identical outputs from run to run, but GPA-VGGT does not, because its key merging and pooling use atomic scatter operations.
Its largest output difference between runs is 4.6, on aggregator outputs of magnitude up to about 111 ($N{=}32$).
For timing, GPA-VGGT and SAF3R were run in the same sessions as all other rows of \cref{tab:cost}.
GPA-VGGT was timed with bf16 autocast; fp16 gives the same aggregator times within 0.4\%.
FastVGGT was added later and was therefore timed in its own sessions, together with dense attention and \methodname; its speed-ups in \cref{tab:costmain} are relative to the dense run of those sessions.

\begin{table}[t]
\centering
\small
\setlength{\tabcolsep}{3pt}
\caption{\textbf{Baseline bring-up check} (single seed, $N{=}32$, one environment). Not used for any claim; the main comparison is \cref{tab:main}. Bold: the better baseline per scene and metric.}
\label{tab:bringup}
\begin{tabular}{@{}llccc@{}}
\toprule
\headrow Scene & Method & AUC@5 & AUC@30 & AbsRel \\
\midrule
TUM fr1\_room & dense & 29.15 & 77.90 & 0.0418 \\
 & SAF3R & 23.02 & 73.50 & \textbf{0.0414} \\
 & GPA-VGGT & \textbf{26.94} & \textbf{76.13} & 0.0429 \\
ETH3D courtyard & dense & 54.40 & 90.22 & 0.0198 \\
 & SAF3R & 30.97 & 80.34 & 0.0245 \\
 & GPA-VGGT & \textbf{41.37} & \textbf{84.69} & \textbf{0.0222} \\
\bottomrule
\end{tabular}
\end{table}

\subsection{Pre-registered bars}
\label{sec:bars}

This section records the pre-registered bars, that is, the pass/fail thresholds written down before each experiment, and separates confirmatory results from post-hoc analyses.
The main table and the fresh-data test are confirmatory; the remaining analyses are post hoc unless they are identified above as registered.

\paragraph{What ``registered'' means.}
A test is called registered when its plan was written into a file before any of its runs: the scenes and frame subsets, the methods, the metrics, the predictions and the rule that decides the outcome.
The plan file was then hashed with SHA-256, so any later change to it would change the hash; amendments were written as new, separately hashed files that point to the original.
\Cref{tab:registry} lists the first 12 hexadecimal digits of each hash and the time the plan was fixed (UTC$+$6).
The bars of the main evaluation and of the design study are the exception: they were fixed in the design specification before the test runs, but that document was not hashed on its own.

\begin{table}[t]
\centering
\footnotesize
\setlength{\tabcolsep}{3pt}
\caption{\textbf{Registered plans.} SHA-256 prefix of each plan file and the time it was fixed, before any run of that test (2026, UTC$+$6). Bold: the confirmatory test.}
\label{tab:registry}
\begin{tabular}{@{}llr@{}}
\toprule
Test & Plan hash (prefix) & Fixed \\
\midrule
\pithree predictions, newer release & \texttt{20947731e395} & Nov 13, 15:02 \\
\pithree predictions, original checkpoint & \texttt{d338191c0597} & Nov 13, 15:44 \\
\pithree transfer evaluation & \texttt{5ed86d7cbc44} & Nov 13, 18:19 \\
Selector applied to GPA-VGGT & \texttt{1d5d99725695} & Nov 13, 21:37 \\
7-Scenes long sequences & \texttt{d360a911d5f0} & Nov 13, 21:52 \\
\textbf{Confirmatory test (first version)} & \texttt{2ee8727dd5a1} & Nov 14, 04:31 \\
800 frames & \texttt{21e5457f0f36} & Nov 14, 00:49 \\
Equal-cost windows & \texttt{ff4462e4c988} & Nov 14, 15:53 \\
Random input order & \texttt{5d4caa447be6} & Nov 14, 22:31 \\
Simpler selectors & \texttt{782dc763645b} & Nov 14, 22:57 \\
Hub estimator on real data & \texttt{3d57a51caf67} & Nov 15, 01:37 \\
RealEstate10K outdoor & \texttt{d7ac3d273aed} & Nov 15, 05:00 \\
\bottomrule
\end{tabular}
\end{table}

All bars were written before the corresponding runs.
They were set for four stages.
\begin{itemize}
  \item \emph{Composed pilot (seeds 0--1).} The loss must be $\le4$ AUC@5 on each low-$\rho$ development scene and $\le3$ on the redundant ones; the interaction between groups must be $\le1.5$; the in-pass router must stay within 0.03 vote recall and $+0.03$ orphan share of the dense-pass router; and the late layers may add $\le1.5$.
  A stop rule ends the pilot if the low-$\rho$ mean loss exceeds 5 or any scene exceeds 8; in that case, any swap-back (returning one group to dense attention) that recovers more than 2.5 names the group to fix.
  \item \emph{Repair (seeds 0--1 screen, seeds 0--4 confirmation).} The loss bars are the same. The cheapest arm that passes the screen is selected and confirmed on 5 seeds; if it fails, the next screen-passer is tried; exact layers 11--16 are the fallback.
  \item \emph{Main table.} On the unseen low-$\rho$ scenes, the loss must be within 3 AUC@5 and 2 AUC@30 of dense attention (CI upper bounds $\le5$ and $\le4$); the gain over the best baseline must be $\ge+5$ AUC@5 at $\le1.10\times$ its FLOPs (CI lower bound $>0$); and on the unseen high-$\rho$ scenes the loss must be within 3 of dense attention.
  \item \emph{Transfer.} The groups predicted as matching for \pithree must carry $\ge50\%$ of the loss.
\end{itemize}
\paragraph{Outcomes.}
\emph{Pilot.} The pilot fired its stop rule (courtyard 11.55).
Its swap-backs named layers 11--16 as the group to fix: returning them to dense attention recovered 3.66 pooled, the only recovery above 2.5.
The router bars passed.
\emph{Repair.} The repair screen passed the $\sigma{=}2$ lattice and exact layers 13--15, and failed exact layer 14 (courtyard 10.28).
In the confirmation on more seeds, the $\sigma{=}2$ lattice then failed (courtyard 4.92), and so did exact layers 13--15 (TUM fr1 4.10).
The fallback, exact layers 11--16, passed every bar and became the frozen recipe.
\emph{Main table.} The bars for closeness to dense accuracy pass: on the unseen low-$\rho$ scenes the loss is 1.38 AUC@5 (CI upper bound 2.01) and 0.70 AUC@30 (upper bound 1.26), and on the unseen high-$\rho$ scenes it is 0.73.
On the high-$\rho$ scenes, the gain over AVGGT(9) is 9.77 (CI [3.23, 15.78]).
\textbf{The bar over the best baseline fails.}
On the unseen low-$\rho$ scenes the best baseline is GPA-VGGT, and \methodname gains only 1.04 AUC@5 over it (CI [$-0.34$, 2.54]) at 1.01$\times$ its FLOPs.
This gain is short of the required $+5$, and its CI includes 0.
The larger gains over SAF3R (3.15, CI [0.20, 6.39], at 1.15$\times$ its FLOPs) and over AVGGT(4) (8.89, CI [6.45, 11.97], at 1.13$\times$) do not satisfy the bar either, for two reasons: the bar is set against the best baseline only, and both comparisons exceed 1.10$\times$ the baseline's FLOPs.
\emph{Transfer.} The group rule fails on the original \pithree: it predicted layer 7 and the late group as matching, but only the late group was observed to be (\cref{sec:pi3}).
With the rule applied untuned, the target of a loss within 3 AUC@5 fails on the unseen low-$\rho$ scenes (4.87, CI [2.45, 7.51]) and passes on the development scenes (2.53, CI [0.40, 4.96]).

\section{Results in detail}
\subsection{Per-scene results and intervals}
\label{sec:perscene}

This section breaks the main evaluation down by scene and by dataset and gives the uncertainty intervals and scene-level tests behind it.

\begin{table*}[t]
\centering
\footnotesize
\compactpending
\setlength{\tabcolsep}{3pt}
\caption{\textbf{Paired differences to dense} on the unseen scenes (pooled; hierarchical bootstrap 95\% CI), $N{=}32$, 5 seeds. Loss = dense $-$ method; $\Delta$AbsRel = method $-$ dense. Bold: lowest value per column.}
\label{tab:deltas}
\begin{tabular}{@{}lcccc@{}}
\toprule
\headrow Method & AUC@5 loss, low-$\rho$ & AUC@30 loss, low-$\rho$ & AUC@5 loss, high-$\rho$ & $\Delta$AbsRel$\times100$, low-$\rho$ \\
\midrule
AVGGT ($\sigma{=}4$) & 10.27 [7.98, 13.22] & 4.30 [2.25, 6.85] & 5.50 [1.40, 9.53] & -0.26 \\
AVGGT ($\sigma{=}9$) & 18.91 [14.64, 24.51] & 12.63 [7.73, 18.59] & 10.50 [4.01, 16.65] & -0.02 \\
GPA-VGGT & 2.42 [1.28, 3.71] & 1.67 [0.50, 3.09] & \textbf{0.08} [$-0.20$, 0.42] & +0.31 \\
SAF3R & 4.53 [1.73, 7.73] & 1.09 [$-0.52$, 2.73] & 1.73 [0.60, 3.20] & +1.14 \\
\methodname & \textbf{1.38} [0.71, 2.01] & \textbf{0.70} [0.17, 1.26] & 0.73 [0.25, 1.27] & +0.34 \\
\methodname-fast & 4.65 [2.74, 6.72] & 1.54 [0.71, 2.49] & 1.81 [0.24, 3.67] & \textbf{-0.37} \\
\bottomrule
\end{tabular}
\end{table*}

\begin{table*}[t]
\centering
\footnotesize
\compactpending
\setlength{\tabcolsep}{3pt}
\caption{\textbf{Per-scene AUC@5} (5-seed means, $N{=}32$ unless the scene has fewer views). Rows are generated from the main-table results file. Bold: highest sparse value per scene (ties all bold).}
\label{tab:perscene}
\begin{tabular}{@{}llccccccc@{}}
\toprule
\headrow Split & Scene & Dense & AVGGT(4) & AVGGT(9) & GPA-VGGT & SAF3R & \methodname & \methodname-fast \\
\midrule
\multirow{13}{*}{\rotatebox{90}{Unseen low-$\rho$}} & ETH3D delivery\_area & 25.1 & 14.8 & 8.9 & 24.1 & 24.4 & \textbf{24.6} & 19.0 \\
 & ETH3D electro & 61.2 & 48.1 & 35.5 & 58.0 & 44.5 & \textbf{58.3} & 54.6 \\
 & ETH3D kicker & 90.2 & 84.8 & 79.0 & \textbf{89.1} & 88.6 & 88.4 & 86.7 \\
 & ETH3D playground & 25.8 & 17.9 & 9.0 & 20.9 & 21.5 & 24.0 & \textbf{25.8} \\
 & ETH3D relief & 23.5 & 13.1 & 4.5 & 18.6 & 21.2 & \textbf{22.7} & 19.0 \\
 & ETH3D relief\_2 & 10.1 & 4.9 & 1.7 & 7.0 & 7.3 & \textbf{9.0} & 7.2 \\
 & ETH3D terrace & 81.5 & 57.4 & 36.7 & 74.7 & 78.9 & \textbf{80.4} & 68.8 \\
 & ETH3D terrains & 29.8 & 22.8 & 20.4 & 30.8 & \textbf{35.7} & 27.2 & 27.3 \\
 & T\&T Barn & 90.5 & 81.4 & 70.5 & \textbf{89.1} & 85.4 & 88.4 & 85.6 \\
 & T\&T Caterpillar & 90.2 & 82.3 & 73.5 & \textbf{89.9} & 84.4 & 89.5 & 84.6 \\
 & T\&T Courthouse & 58.8 & 45.9 & 35.0 & 55.8 & 46.8 & 59.1 & \textbf{59.6} \\
 & T\&T Meetingroom & 91.0 & 81.6 & 74.7 & \textbf{89.9} & 86.8 & 89.4 & 88.5 \\
 & TUM fr1\_plant & 68.0 & 57.5 & 50.5 & 66.6 & 61.4 & \textbf{67.0} & 58.8 \\
\midrule
\multirow{8}{*}{\rotatebox{90}{Unseen high-$\rho$}} & TUM fr2\_desk & 88.9 & 74.5 & 65.8 & 87.9 & 88.2 & 87.0 & \textbf{88.6} \\
 & TUM fr3\_str.\_tex.\_far & 89.3 & 78.9 & 72.4 & \textbf{89.7} & 83.4 & 89.2 & 83.8 \\
 & 7-Scenes chess & 46.9 & \textbf{47.6} & 46.3 & 47.0 & 45.6 & 46.2 & 47.0 \\
 & 7-Scenes fire & 52.3 & \textbf{55.7} & 55.6 & 52.5 & 48.7 & 51.1 & 51.2 \\
 & 7-Scenes pumpkin & 15.9 & 15.0 & 14.5 & \textbf{16.2} & 14.6 & 15.4 & 14.9 \\
 & DTU scan 24 & 100.0 & 89.7 & 82.7 & 99.9 & 99.8 & \textbf{100.0} & 99.8 \\
 & DTU scan 114 & 99.8 & 95.3 & 88.8 & \textbf{99.8} & 99.4 & 99.7 & \textbf{99.8} \\
 & T\&T Ignatius & 89.9 & 82.3 & 72.8 & 89.4 & \textbf{89.5} & 88.6 & 83.4 \\
\midrule
\multirow{5}{*}{\rotatebox{90}{Development}} & TUM fr1\_room & 29.8 & 23.1 & 16.5 & 28.3 & 23.4 & \textbf{29.8} & 25.6 \\
 & ETH3D courtyard & 58.5 & 21.0 & 8.1 & 53.1 & 36.4 & \textbf{57.1} & 52.5 \\
 & ETH3D office & 43.2 & 33.9 & 30.7 & \textbf{44.0} & 39.5 & 42.8 & 40.5 \\
 & TUM fr3\_long\_office & 87.5 & 70.8 & 65.5 & \textbf{86.7} & 82.8 & 86.5 & 81.7 \\
 & 7-Scenes office & 26.9 & \textbf{27.2} & 25.1 & 26.6 & 25.5 & 26.9 & 25.9 \\

\bottomrule
\end{tabular}
\end{table*}

\begin{table*}[t]
\centering
\small
\setlength{\tabcolsep}{4.5pt}
\caption{\textbf{Per-dataset comparison.} AUC@5 lost against dense attention, mean over each dataset's unseen test scenes of the 5-seed scene means ($N{=}32$; in brackets, how many of the scenes are low-$\rho$). Negative: above dense attention. Bold: lowest loss in the row. Descriptive: datasets with 2--3 scenes carry no interval. $^{\ast}$A separate post-hoc test (\cref{sec:re10k}): 9 outdoor test clips, 5 frame subsets each, all of high redundancy; \methodname-fast was not run.}
\label{tab:datasets}
\begin{tabular}{@{}lccrrrrrr@{}}
\toprule
\headrow Dataset & Scenes & Dense AUC@5 & AVGGT(4) & AVGGT(9) & GPA-VGGT & SAF3R & \methodname & \methodname-fast \\
\midrule
ETH3D & 8 (8) & 43.4 & $10.44$ & $18.93$ & $3.01$ & $3.14$ & $\mathbf{1.57}$ & $4.84$ \\
Tanks and Temples & 5 (4) & 84.1 & $9.40$ & $18.78$ & $1.25$ & $5.50$ & $\mathbf{1.10}$ & $3.78$ \\
TUM RGB-D & 3 (1) & 82.1 & $11.81$ & $19.20$ & $\mathbf{0.73}$ & $4.46$ & $1.04$ & $5.04$ \\
7-Scenes & 3 (0) & 38.4 & $\mathbf{-1.08}$ & $-0.46$ & $-0.21$ & $2.04$ & $0.80$ & $0.64$ \\
DTU & 2 (0) & 99.9 & $7.36$ & $14.13$ & $0.06$ & $0.30$ & $\mathbf{0.05}$ & $0.06$ \\
\midrule
RealEstate10K outdoor$^{\ast}$ & 9 (0) & 33.2 & $2.11$ & $5.14$ & $\mathbf{-0.76}$ & $2.49$ & $0.22$ & -- \\

\bottomrule
\end{tabular}
\end{table*}

\Cref{tab:deltas,tab:perscene} complete \cref{tab:main}: the first gives each method's paired difference to dense attention with intervals, and the second gives every scene.
\Cref{tab:datasets} groups the test scenes by dataset.
\methodname has the lowest loss on ETH3D, Tanks and Temples and DTU; GPA-VGGT has the lowest on TUM; and AVGGT(4) has the lowest on 7-Scenes.
On that dataset, however, every method except AVGGT(9) and SAF3R stays within 1 AUC@5 of dense attention, so the differences there are small.
On the unseen low-$\rho$ scenes, the paired AUC@5 gain of \methodname is:
\begin{itemize}
  \item 8.89 over AVGGT(4) (CI [6.45, 11.97]);
  \item 17.54 over AVGGT(9) (CI [13.16, 23.11]);
  \item 3.15 over SAF3R (CI [0.20, 6.39]);
  \item 1.04 over GPA-VGGT (CI [$-0.34$, 2.54]), an interval that includes zero.
\end{itemize}
The difference to GPA-VGGT is uneven across scenes.
On the unseen low-$\rho$ scenes, \methodname is ahead on ETH3D terrace (5.6), relief (4.1), playground (3.1) and relief\_2 (2.1) and on Tanks and Temples Courthouse (3.3); it is behind on ETH3D terrains ($-3.6$); and it is within 1 AUC@5 on the other seven.
On the unseen high-$\rho$ scenes, in contrast, \methodname is behind GPA-VGGT on seven of eight, by at most 1.4.
On ETH3D terrains, SAF3R exceeds dense attention by 6.0.
This scene is not unseen for this baseline: its head assignment was calibrated on ETH3D training scenes that include terrains (\cref{sec:baselines}).
\Cref{tab:scenetests} tests whether these gains hold at the scene level, treating the 13 scene means as the samples.
The Holm correction runs over the four baselines that were part of the main evaluation from the start (AVGGT(4), AVGGT(9), SAF3R and GPA-VGGT); FastVGGT was added later and is not part of these tests.
After a Holm correction over the four baselines, the results are as follows.
The gains over AVGGT(4) and AVGGT(9) are significant in both tests.
The gain over SAF3R is significant only in the signed-rank test, which is robust to the terrains outlier.
The gain over GPA-VGGT is significant in neither test.
On the 8 high-$\rho$ test scenes, no comparison survives the correction; GPA-VGGT is ahead of \methodname on 7 of them (raw $p=0.023$ in both tests).

\subsection{Point-map metrics}
\label{sec:pointmap}

This post-hoc analysis asks whether the pose results are also reflected in the quality of the predicted point maps.

\emph{This analysis of existing predictions is post hoc; its plan was registered before scoring.}
It uses the 9 unseen low-$\rho$ scenes that have ground-truth geometry (8 ETH3D scenes and TUM fr1\_plant, 5 seeds each).
Point maps were scored with the accuracy, completeness and normal-consistency metrics of AVGGT and \pithree, with 5\,cm voxels for ETH3D and 2\,cm for TUM.
The Chamfer distance is the mean of accuracy and completeness, in metres, so lower is better.
Before scoring, each prediction must be aligned to the ground truth, and two alignments are reported (\cref{tab:pointmap}).
Alignment \emph{A} fits a similarity transform between the predicted and ground-truth camera centres, so its Chamfer distance also carries pose error.
Alignment \emph{B} fits the transform on per-pixel point correspondences, which isolates the quality of the geometry itself.

\begin{table}[t]
\centering
\scriptsize
\setlength{\tabcolsep}{2pt}
\caption{\textbf{Point maps on the 9 unseen low-$\rho$ scenes with ground truth} (post hoc). Change in Chamfer distance (m) against dense attention, and paired contrasts, with hierarchical 95\% CIs; negative is better. A: alignment on camera centres (includes pose error); B: alignment on points. Bold: \methodname against dense attention, and its two contrasts that are significant (alignment A).}
\label{tab:pointmap}
\begin{tabular}{@{}lcc@{}}
\toprule
\headrow & A (cameras) & B (points) \\
\midrule
\oursrow \methodname $-$ dense & $\mathbf{+0.007}$ [$-0.014$, 0.029] & $\mathbf{+0.001}$ [$-0.014$, 0.015] \\
\methodname-fast $-$ dense & $-0.057$ [$-0.116$, $-0.010$] & $+0.034$ [$-0.010$, 0.095] \\
GPA-VGGT $-$ dense & \fail{$+0.066$} [0.018, 0.130] & $+0.011$ [$-0.006$, 0.035] \\
SAF3R $-$ dense & \fail{$+0.064$} [0.031, 0.098] & \fail{$+0.020$} [0.005, 0.038] \\
AVGGT(4) $-$ dense & \fail{$+0.120$} [0.033, 0.219] & -- \\
AVGGT(9) $-$ dense & \fail{$+0.283$} [0.108, 0.541] & -- \\
\midrule
\methodname $-$ GPA-VGGT & \pass{$\mathbf{-0.059}$} [$-0.121$, $-0.009$] & $-0.010$ [$-0.045$, 0.016] \\
\methodname $-$ SAF3R & \pass{$\mathbf{-0.057}$} [$-0.097$, $-0.021$] & $-0.019$ [$-0.044$, 0.003] \\
\bottomrule
\end{tabular}
\end{table}

Under both alignments, the point-map differences between \methodname and dense attention are not statistically significant.
The normals of \methodname are slightly worse, however (normal consistency $-0.006$, CI [$-0.010$, $-0.003$], under A).
Under A, \methodname is 0.059\,m better than GPA-VGGT and 0.057\,m better than SAF3R.
Under B, these gaps shrink to 0.010 and 0.019\,m and are no longer significant, which means that most of the gap under A comes from pose error rather than from worse geometry.
Under B, only SAF3R remains significantly worse than dense attention ($+0.020$\,m).
\methodname-fast has a lower Chamfer distance than dense attention under A, but this is an artefact of aligning on its less accurate cameras, not better geometry.
Two ETH3D scenes (relief\_2 and relief) carry 73\% of this improvement, and under B, which does not depend on the cameras, its change is $+0.034$\,m.
All arms were scored from float32 depth; storing depth as float16, as was done for the later runs, changes the Chamfer distance by at most $10^{-3}$\,m.
Further metrics for every arm and split are released with the code: rotation and translation accuracy at 5, 15 and 30 degrees, pose AUC at the same thresholds, ATE and RPE.

\subsection{Cost and timing}
\noindent This section reports the measured and modelled cost of the evaluated methods.

\begin{table}[t]
\centering
\small
\compactpending
\setlength{\tabcolsep}{2.37pt}
\caption{\textbf{Cost and time}. GFLOP per token from the exact cost model (\cref{eq:cost}). Times: one session per $N$ in which every method, dense included, was timed interleaved on the same input (1 warm-up and 3 timed rounds, median, CUDA events) on one 16\,GB GPU. ``Agg.'': aggregator only, including the hub pilot; ``full'': aggregator (which includes the DINOv2 patch embedding) plus camera and depth heads, i.e.\ the whole network; image loading and resizing are excluded (at $N{=}100$, in a separate session: dense 33.26\,s, \methodname 17.19\,s, GPA-VGGT 16.84\,s). $^\dagger$Own sessions, in which dense took 5.14, 14.88 and 31.30\,s (aggregator). Bold: lowest sparse value per column.}
\label{tab:cost}
\begin{tabular}{@{}lccccccc@{}}
\toprule
 & \multicolumn{2}{c}{GFLOP / token} & \multicolumn{3}{c}{Agg.\ time (s)} & \multicolumn{2}{c}{Full time (s)} \\
\cmidrule(lr){2-3}\cmidrule(lr){4-6}\cmidrule(lr){7-8}
Method & $N{=}32$ & 100 & 32 & 64 & 100 & 32 & 64 \\
\midrule
Dense & 5.29 & 12.25 & 5.27 & 15.02 & 31.32 & 5.71 & 16.05 \\
\midrule
AVGGT ($\sigma{=}4$) & 2.64 & 3.77 & 3.43 & 7.54 & 12.98 & 3.90 & 8.63 \\
AVGGT ($\sigma{=}9$) & \textbf{2.38} & \textbf{2.94} & \textbf{3.22} & \textbf{6.76} & \textbf{11.26} & \textbf{3.72} & \textbf{7.88} \\
GPA-VGGT & 2.96 & 4.79 & 3.76 & 8.68 & 15.69 & 4.03 & 9.41 \\
SAF3R & 2.60 & 3.67 & 3.94 & 8.34 & 14.28 & 4.41 & 9.34 \\
FastVGGT$^\dagger$ & -- & -- & 4.20 & 9.41 & 16.55 & 4.60 & 10.11 \\
\midrule
\oursrow \textbf{\methodname} & 2.99 & 4.76 & 3.68 & 8.54 & 15.35 & 4.22 & 9.69 \\
\methodname-fast & 2.49 & 3.14 & 3.48 & 7.51 & 12.55 & 3.94 & 8.50 \\
\bottomrule
\end{tabular}
\end{table}

\subsection{Tanks and Temples at \texorpdfstring{$N{=}100$}{N=100}}
\label{sec:tt100}
This registered evaluation asks how the method performs on longer inputs, with more frames per subset, on the Tanks and Temples scenes.

Each of the five scenes of \cref{tab:costmain} was run with 5 seeds under the strided rule of \cref{sec:repro}.
At $N{=}100$, however, the strided rule gives only 4, 3, 5, 3 and 3 distinct frame subsets for Barn, Caterpillar, Courthouse, Meetingroom and Ignatius.
Some subsets are therefore repeated, and repeated subsets enter the means more than once.
On the four low-$\rho$ scenes \methodname loses 2.17 AUC@5 (CI [1.33, 3.09]); on Ignatius it loses 0.80; and on the held-out Truck, reported separately, it loses 0.17.
Over all five scenes it gains 10.71 AUC@5 over AVGGT(9) at an indicative 1.32$\times$ its time.
The times in this run are medians of single timed forwards, so they serve only as indications.
GPA-VGGT and SAF3R were run afterwards (post hoc) on the same scenes, seeds and dense reference.
Over the five scenes, GPA-VGGT loses 1.61 AUC@5 (CI [0.16, 4.19]) and SAF3R loses 5.65 (CI [2.89, 8.93]).
The paired difference between \methodname and GPA-VGGT is $-0.29$ (CI [$-1.77$, 2.00]), which is not significant, whereas \methodname is 3.76 ahead of SAF3R (CI [1.38, 6.55]).
Per scene group, the losses of \methodname, GPA-VGGT and SAF3R are:
\begin{itemize}
  \item 2.17, 1.91 and 6.75 on the four low-$\rho$ scenes;
  \item 0.80, 0.40 and 1.28 on Ignatius;
  \item 0.17, $-0.61$ and 1.39 on the held-out Truck, so on Truck GPA-VGGT is ahead of \methodname by 0.77 (CI [0.13, 1.43]); with each distinct frame subset counted once, this lead is 0.99 (CI [$-0.08$, 2.09]).
\end{itemize}
Repeated subsets give bit-identical runs, so they can also be counted only once.
Doing so moves every five-scene mean by at most 0.2 AUC@5: \methodname loses 1.95 (CI [1.16, 2.82]), GPA-VGGT 1.60 (CI [0.10, 4.20]) and SAF3R 5.52.
The difference between \methodname and GPA-VGGT stays non-significant ($-0.35$, CI [$-1.83$, 1.96]).

\subsection{Long sequences: 333 and 800 frames}
\label{sec:longseq}
This evaluation measures accuracy and runtime on long video sequences, where the cost of dense attention matters most.

Following the AVGGT protocol~\cite{sun2025avggt}, 7-Scenes test sequences were run in two settings, both with the frozen recipe and no retuning.
The first uses every third frame ($N{=}333$ on six sequences, 167 on stairs).
The second uses 800 frames on four sequences: chess-03, office-02, fire-03 and redkitchen-03, the last of which was held out from all design choices.
Dense attention at $N{=}800$ needs more than 16\,GB of GPU memory.
Dense attention, and any other method that also exceeds this memory, was therefore run by a chunked engine that writes activations to disk between segments and is bit-identical to the reference forward; the reported times include this disk I/O.
At these lengths, the official implementations of GPA-VGGT and SAF3R also exceed 16\,GB.
For $N{=}333$, GPA-VGGT was therefore run through a port that applies its head-merging step two heads at a time, which is arithmetically identical to the original.
At $N{=}16$, 32 and 100, the port matches the official outputs within twice GPA-VGGT's own run-to-run variation over three official runs.
This criterion was adopted after a single-pair version of it failed at $N{=}16$ on a low noise sample.
SAF3R, and GPA-VGGT at $N{=}800$, were not run.
At $N{=}333$, \methodname loses 0.60 AUC@5 against dense attention (paired 95\% CI [$-0.27$, 1.87]) and 0.16 AUC@30, and its aggregator runs 2.7$\times$ faster.
For comparison, AVGGT(4) loses 0.06 and 0.45 at 3.9$\times$, and AVGGT(9) loses 1.69 and 1.08 at 5.4$\times$.
GPA-VGGT exceeds dense attention by 0.82 AUC@5 (CI [0.20, 1.77]) and exceeds \methodname by 1.42 (CI [0.20, 3.55]), at 2.9$\times$ (2.6$\times$ if the first call in each process, which includes kernel autotuning, is kept).
This is in line with the advantage of GPA-VGGT on redundant views at $N{=}32$.
At $N{=}800$ (\cref{tab:longseq}), \methodname stays within 1.6 AUC@5 of dense attention on every sequence and runs 3.3$\times$ faster.
AVGGT is faster still, at 5.1--8.1$\times$, but it is less faithful to dense attention: its AUC@5 departs from dense attention by up to 7.9, upward on fire-03 and redkitchen-03, and its AUC@30 falls below dense attention on three of the four sequences.
These speed-ups include the disk I/O of the chunked runs, which affects all dense runs and one run each of AVGGT(4) and \methodname; without that I/O, the speed-ups are 3.3, 5.1 and 7.8$\times$.
The AVGGT paper reports a similar effect, with AVGGT(9) above dense VGGT at 800 frames (7-Scenes AUC@30 77.4 against 74.2).
This suggests that frame-local pooling can act as a regulariser on long redundant video; \methodname, which aims to reproduce dense attention, does not show this effect.
LMSA-auto is a variant that selects the exact layers per sequence from the in-pass $\bar\Gamma$, rather than using the frozen exact layers.
At this length it changes only layer 16, whose $\bar\Gamma$ lies at 1.41--1.74 and falls below the cut-off on three of the four sequences; in contrast, $\bar\Gamma$ is 1.9--8.2 for layers 11--15 and at most 0.97 for the hub layers.
The confirmatory test of \cref{sec:confirm} also covers these $N{=}333$ sequences.

\emph{Every tenth frame (post hoc).} This run uses $N{=}100$ frames taken at stride 10 from each of the 18 test sequences (7 scenes).
\methodname loses 0.53 AUC@5 against dense attention (paired 95\% CI [0.14, 0.94]), whereas GPA-VGGT is 0.41 above dense attention (CI [$-0.03$, 0.91]).
AVGGT(4), AVGGT(9) and SAF3R lose 1.15, 3.57 and 1.58, with CIs that include 0.
Mean end-to-end times are 16.0\,s for \methodname, 16.6\,s for GPA-VGGT and 30.8\,s for dense attention.
So, as at stride 3, GPA-VGGT is slightly more accurate than \methodname on this redundant video, at a similar time.
Back at stride 3, LMSA-auto chose a different layer set from the frozen recipe on 2 of the 7 sequences, and its mean AUC@5 is 34.12, against 34.24 for the frozen recipe, so choosing layers per sequence made almost no difference.

\begin{table}[t]
\centering
\small
\setlength{\tabcolsep}{1.47pt}
\caption{\textbf{800 frames} on four 7-Scenes sequences, one frame subset each. Per-sequence AUC@5 minus that of dense attention (positive = above dense), mean $|\Delta|$ of AUC@5, mean $\Delta$ of AUC@30, and mean end-to-end speed-up over dense attention. Dense AUC@5: 43.2, 15.4, 53.1 and 6.7. Bold: the two lowest mean absolute differences (both \methodname variants), the largest departure from dense attention and the largest speed-up.}
\label{tab:longseq}
\begin{tabular}{@{}lrrrrccc@{}}
\toprule
\headrow Method & chess & office & fire & kitchen & $|\Delta_5|$ & $\Delta_{30}$ & Speed-up \\
\midrule
AVGGT(4) & $-0.85$ & $+0.96$ & $\mathbf{+7.85}$ & $+4.28$ & 3.48 & $-0.11$ & 5.1$\times$ \\
AVGGT(9) & $-1.92$ & $+0.32$ & $+5.93$ & $+5.04$ & 3.30 & $-0.90$ & \textbf{8.1}$\boldsymbol{\times}$ \\
\methodname & $-0.33$ & $+0.94$ & $-1.57$ & $+1.35$ & \textbf{1.05} & $+0.59$ & 3.3$\times$ \\
LMSA-auto & $-0.13$ & $+0.94$ & $-0.97$ & $+1.35$ & \textbf{0.85} & $+0.70$ & 3.4$\times$ \\
\bottomrule
\end{tabular}
\end{table}

\subsection{Outdoor scenes: RealEstate10K}
\label{sec:re10k}

This test asks whether \methodname keeps its accuracy on outdoor scenes, a domain that none of the main-table datasets covers well.
\emph{It is post hoc relative to the main table, but its scene list, frame rule, methods and predictions were registered before it ran, and it is reported in full.}

\paragraph{Data and scene selection.}
RealEstate10K~\cite{zhou2018stereo} consists of real-estate video clips with camera poses estimated from the videos.
The test clips come from the preprocessed release of the pixelSplat authors~\cite{charatan2024pixelsplat}; their full test split could not be downloaded, so the candidate pool is the 41 test clips of their published subset.
Every clip was labelled indoor or outdoor from thumbnails before any model was run, and the test set is every outdoor clip with at least 32 frames: 9 clips of yards, facades, pools, porches and decks.
Each of 5 frame subsets takes $N{=}32$ frames spread over the whole clip.
The poses are known only up to scale, so the scores use the relative rotation and translation direction, as in the main table; the dataset has no depth.

\paragraph{Result.}
\methodname loses 0.22 AUC@5 against dense attention (95\% CI [$-1.55$, 2.33]) and 0.40 AUC@30 (CI [$-0.51$, 1.33]), so on these clips it is practically as accurate as dense attention.
GPA-VGGT is 0.76 above dense attention, and the difference between the two methods is not significant ($+0.98$, CI [$-1.73$, 4.05]).
AVGGT(4), AVGGT(9) and SAF3R lose 2.11, 5.14 and 2.49; only the gap to AVGGT(9) is significant ($-4.92$, CI [$-9.89$, $-0.20$]).
Per clip, \methodname loses more than 3 AUC@5 on one clip (at most 7.0, a covered porch), as does GPA-VGGT, whereas AVGGT(9) does so on four clips.

\paragraph{What this test does not show.}
Every clip is of high redundancy ($\rho$ between 16 and 30), because a real-estate video revisits the same space many times.
The test therefore shows that \methodname carries over to outdoor scenes, but it does not test the low-redundancy regime that the main table targets.
With 9 clips the intervals are wide, and GPA-VGGT's negative loss comes mostly from one clip with moving people, on which it beats dense attention by 14.4.
Without that clip and one clip that fades to white, \methodname loses 0.82 (CI [$-0.85$, 3.19]) and GPA-VGGT 1.05 (CI [$-0.22$, 2.71]).
The frames come from YouTube videos and are used for research only; the poses are released under CC BY 4.0.

\subsection{Confirmatory test on fresh datasets}
\label{sec:confirm}
This registered test asks whether two variants of \methodname generalise to fresh datasets and long video.

The two variants were chosen after the main results, so they were tested on data not used for any design choice, with their predictions registered before the runs.
\emph{LMSA-Accurate} uses more hubs (1024 in layers 0--8, 2048 elsewhere; \cref{sec:variance}) to reduce sampling noise.
\emph{LMSA-Efficient} saves cost by merging 20\% of the keys in layers 11--16 with the operator of GPA-VGGT (\cref{sec:thinmatch}).
The test uses three datasets: NRGBD (9 scenes, 3 keyframe subsets each, $N{\approx}120$), MPI Sintel (14 sequences, 5 random 32-frame subsets each) and the $N{=}333$ 7-Scenes sequences above.
Both variants and \methodname were run with three hub draws per subset, and the intervals come from a nested bootstrap over scenes, subsets and draws (10{,}000 resamples).
The registered predictions were:
\begin{itemize}
  \item LMSA-Accurate is within 0.5 AUC@5 and 0.3 AUC@30 of dense attention on every dataset, at least as close to it as \methodname, and has a smaller draw-to-draw SD;
  \item LMSA-Efficient is within 1 AUC@5 of \methodname and 1.1--1.3$\times$ faster at $N{=}333$;
  \item the point-map accuracy and completeness of LMSA-Accurate are within 3\% of dense attention.
\end{itemize}
All of these predictions held (\cref{tab:confirm}).
The predictions concern mean differences, however, and the 95\% intervals of LMSA-Accurate reach $-0.78$, $-0.65$ and $-0.55$ AUC@5.
The test therefore does not establish equivalence to dense attention within 0.5; it shows only that the mean differences fall inside that bar.
LMSA-Accurate is slightly below dense attention on NRGBD ($-0.41$, CI [$-0.78$, $-0.10$]), which is within the bar.
It is closer to dense attention than \methodname on every dataset (\methodname: $-1.10$, $-0.62$ and $-0.89$).
Its point maps lie within 1.3\% of dense attention, and on Sintel its depth error equals that of dense attention (AbsRel 0.290).
LMSA-Efficient is 1.16$\times$ faster than \methodname at $N{=}333$ (aggregator, steady state), 1.11$\times$ at $N{\approx}120$ and 1.02$\times$ at $N{\le}32$, so its saving grows with the number of frames.
LMSA-Accurate costs no more than \methodname (0.95--1.00$\times$ its time).
One registered baseline prediction failed in part.
AVGGT(9) has the lowest AUC@5 everywhere, as predicted, but SAF3R is above AVGGT(4) on NRGBD and Sintel instead of being among the lowest.
On Sintel, GPA-VGGT is slightly above dense attention (28.51 against 28.27 AUC@5), as it was on the 7-Scenes video.

\begin{table*}[t]
\centering
\small
\setlength{\tabcolsep}{5pt}
\caption{\textbf{Confirmatory test} (registered before the runs; mean and nested 95\% CI). Left: LMSA-Accurate minus dense attention. Middle: LMSA-Efficient minus \methodname. Right: draw-to-draw SD of AUC@5 within a subset. Bold: the pose-accuracy difference of LMSA-Accurate to dense attention (first column), and the lowest draw SD per dataset.}
\label{tab:confirm}
\begin{tabular}{@{}lccccccc@{}}
\toprule
 & \multicolumn{2}{c}{Accurate $-$ dense} & \multicolumn{2}{c}{Efficient $-$ \methodname} & \multicolumn{3}{c}{Draw SD of AUC@5} \\
\cmidrule(lr){2-3}\cmidrule(lr){4-5}\cmidrule(lr){6-8}
Dataset & AUC@5 & AUC@30 & AUC@5 & AUC@30 & Acc. & Eff. & \methodname \\
\midrule
NRGBD & $\mathbf{-0.41}$ {\scriptsize[$-0.78$, $-0.10$]} & $-0.08$ & $+0.32$ {\scriptsize[$-0.25$, 0.94]} & $+0.07$ & \textbf{0.68} & 1.44 & 1.40 \\
Sintel & $\mathbf{-0.30}$ {\scriptsize[$-0.65$, 0.02]} & $-0.08$ & $+0.27$ {\scriptsize[$-0.38$, 0.93]} & $+0.04$ & \textbf{1.30} & 2.01 & 2.04 \\
7-Scenes & $\mathbf{+0.00}$ {\scriptsize[$-0.55$, 0.48]} & $-0.06$ & $+0.55$ {\scriptsize[$-0.77$, 2.44]} & $-0.00$ & \textbf{0.79} & 0.85 & 1.11 \\
\bottomrule
\end{tabular}
\end{table*}

\begin{table}[t]
\centering
\footnotesize
\setlength{\tabcolsep}{2.6pt}
\caption{\textbf{Scene-level tests} on the 13 unseen low-$\rho$ scenes: $d$ = AUC@5 of \methodname minus that of the baseline, per scene (5-seed means). Two-sided Wilcoxon signed-rank and exact sign-flip tests on $d$; Holm correction over the four baselines within each test. Bold: significant after Holm correction.}
\label{tab:scenetests}
\begin{tabular}{@{}lccccc@{}}
\toprule
\headrow Baseline & Mean $d$ & Ahead & Wilcoxon (Holm) & Sign-flip (Holm) \\
\midrule
AVGGT ($\sigma{=}4$) & +8.89 & 13/13 & \textbf{$<$0.001 (0.001)} & \textbf{$<$0.001 (0.001)} \\
AVGGT ($\sigma{=}9$) & +17.54 & 13/13 & \textbf{$<$0.001 (0.001)} & \textbf{$<$0.001 (0.001)} \\
GPA-VGGT & +1.04 & 8/13 & 0.340 (0.340) & 0.165 (0.165) \\
SAF3R & +3.15 & 11/13 & \textbf{0.017 (0.034)} & 0.064 (0.128) \\
\bottomrule
\end{tabular}
\end{table}

\begin{figure*}[t]
\centering
\includegraphics[width=\textwidth]{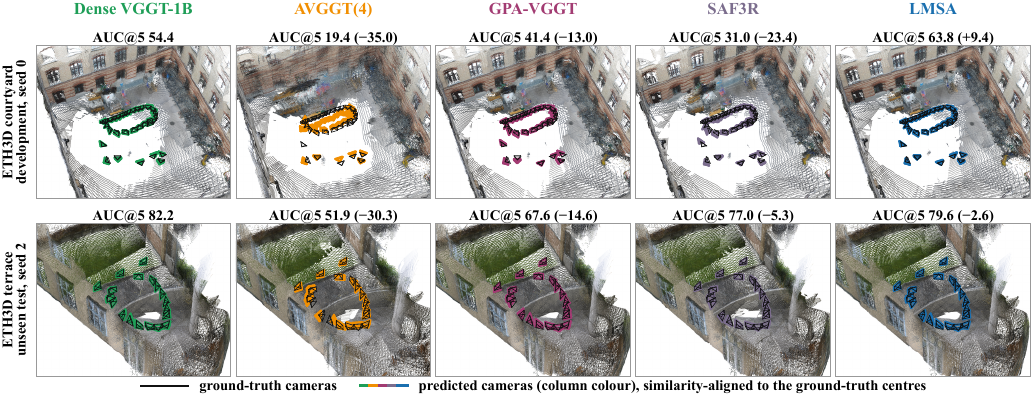}
\caption{\textbf{Reconstructions on two ETH3D scenes~\cite{schops2017eth3d}.} Each panel shows the predicted depth, unprojected with the predicted cameras (86\% most confident pixels), and the predicted cameras (frusta in the column colour), both mapped to the ground-truth frame by the least-squares similarity transform between predicted and ground-truth camera centres; black frusta are the ground-truth cameras, and all panels of a row share one viewpoint. Top: courtyard, a development scene, subset 0. Bottom: terrace, an unseen low-redundancy test scene, subset 2. In each scene the subset is the one on which GPA-VGGT loses the most AUC@5 against dense attention (13.0 and 14.6 points), chosen by GPA-VGGT's loss, not by the result of \methodname. Titles give that subset's pose AUC@5 and, in brackets, its difference from dense VGGT-1B. On courtyard the cameras of AVGGT(4) are visibly displaced and its facades blurred; smaller pose errors are hard to see at this scale. Courtyard was used to design \methodname, so its $+9.4$ there is not a test result.}
\label{fig:qualitative}
\end{figure*}

\section{Why the design}
\subsection{Diffuse layers in isolation}
\label{sec:diffuse}

This section examines the diffuse layers on their own: each diffuse group is made sparse while every other layer stays dense, so that its loss can be measured without interference from the other groups.

\begin{table}[t]
\centering
\footnotesize
\setlength{\tabcolsep}{2.4pt}
\caption{\textbf{Diffuse layers in isolation}. AUC@5 lost against dense when only the named layers are sparse; development scenes, 5 seeds, $\dagger$: 2 seeds. ``mean'': AVGGT's count-biased mean token. Bold: the key sets of \methodname.}
\label{tab:components}
\begin{tabular}{@{}lccccc@{}}
\toprule
\headrow Layers and key set & fr1 & court & ETH-o & fr3 & 7S \\
\midrule
\rowcolor{tabgroup}\multicolumn{6}{@{}l}{\emph{Early, 0--8} (own frame exact)} \\
\quad frame-local (AVGGT) & 5.6 & 33.1 & 1.5 & 0.5 & $-1.1$ \\
\quad + specials & 5.9 & 24.2 & -- & 0.1 & $-1.0$ \\
\quad + specials, $\sigma{=}16$ lattice, mean & $-0.6$ & 1.6 & -- & 1.0 & $-0.4$ \\
\oursrow \quad + specials, 256 hubs (\methodname) & \textbf{0.2} & \textbf{0.9} & -- & \textbf{0.1} & \textbf{0.0} \\
\rowcolor{tabgroup}\multicolumn{6}{@{}l}{\emph{Late, 9 and 17--23}} \\
\quad $\sigma{=}9$ lattice, frame 0, mean$^\dagger$ & -- & 4.5 & 5.6 & 15.0 & -- \\
\quad + cumulant correction$^\dagger$ & -- & 3.6 & 2.5 & 5.4 & -- \\
\quad own, specials, 256 hubs$^\dagger$ & -- & 11.4 & 2.0 & 1.8 & -- \\
\oursrow \quad own, specials, 1024 hubs (\methodname) & \textbf{1.0} & \textbf{1.0} & \textbf{0.3} & \textbf{1.2} & \textbf{0.1} \\
\bottomrule
\end{tabular}
\end{table}

\Cref{tab:components} shows the two diffuse groups in this isolated setting.
In the \emph{early} layers, adding the special tokens of other frames to the own frame is not enough to recover accuracy.
A lattice with a mean token nearly closes the gap on the low-$\rho$ scenes but costs 1.0 on TUM fr3, whereas 256 hubs close the gap on every scene tested.
In the \emph{late} layers, the AVGGT key set loses 15.0 AUC@5 on fr3.
Correcting the mean token's deficit $\Gamma$ recovers most of this loss (\cref{find:late}), which supports the deficit as an explanation without isolating it.
1024 hubs remove the deficit without any correction term, whereas 256 hubs are too few and lose 11.4 on courtyard.
One caveat remains: the 1024-hub courtyard loss of 1.0 has a seed-level 95\% upper bound of 4.4, which is above the pre-registered bound of 4.

\subsection{Further evidence for the design}
\label{sec:evidence}

This post-hoc evidence examines which keys the matching layers rely on and why the other components of the design were chosen.

\begin{table}[t]
\centering
\small
\setlength{\tabcolsep}{3pt}
\caption{\textbf{Which keys matter in layers 10--16.} AUC@5 lost against dense when a key set is removed from layers 10--16 only (5 seeds, low-$\rho$ development scenes). ``Correspondence'': the model's own $3{\times}3$ correspondence windows. ``Heaviest'': as many of the highest-attention non-correspondence keys. ``Highest lift'': as many keys ranked by $A(t,j)/\pi_j$. ``Random'': as many random keys. Mass: head-averaged attention mass removed (for highest lift: range over layers). Bold: the two largest losses per scene.}
\label{tab:knockout}
\begin{tabular}{@{}lcccc@{}}
\toprule
\headrow Removed set & fr1 & court & ETH-o & Mass \\
\midrule
Correspondence & \textbf{22.1} & \textbf{49.2} & \textbf{17.7} & 0.03--0.04 \\
Heaviest & 6.5 & 23.0 & 6.0 & 0.07--0.11 \\
Highest lift & \textbf{29.7} & \textbf{58.2} & \textbf{42.5} & 0.01--0.14 \\
Random & 0.1 & $-0.3$ & 0.0 & -- \\
\bottomrule
\end{tabular}
\end{table}

\paragraph{Keys.}
\Cref{tab:knockout} collects the knockouts, in which one set of keys is removed from the matching layers and the loss is measured.
Removing the correspondence windows, which hold very little attention mass, costs far more than removing as many of the heaviest keys, and removing random keys costs nothing.
The highest-lift keys are those whose attention from a given query most exceeds their column mass, that is, the keys most specific to that query.
Removing them is the most destructive of all, even though only 13\% of them lie in correspondence windows.

\paragraph{Early layers.}
Making only layers 0--2, 3--5 or 6--8 frame-local costs 18.4, 12.7 and 17.2 AUC@5 on courtyard.
Each block of three early layers therefore carries cross-view information on wide-baseline input, so no early block can simply be restricted to its own frame.

\paragraph{Proposal of the importance sample}.
A layer-local audit measured the output error of each layer on its own, for the far-field keys of layers 10--16.
In it, 256 samples drawn in proportion to the pilot's column mass have about half the output error of 256 uniform samples (0.109 against 0.203), so the column mass is a useful proposal.

\paragraph{Shared versus per-query blocks.}
This test asks whether one block of keys shared by a tile of queries can replace a separate window for each query.
Blocks of $6{\times}10$ keys centred by the oracle lose 0.93 AUC@5 (CI $[-0.84, 2.80]$) relative to per-query $3{\times}3$ oracle windows, pooled over the low-$\rho$ scenes.
On courtyard alone, the loss is 2.62 (CI $[-0.31, 5.55]$).
The pooled result therefore supports shared blocks, but the support is weaker per scene.

\paragraph{Denser lattices instead of routing.}
Without correspondence keys, the lattice must be dense to stay accurate: at $\sigma{=}16$, the loss relative to $\sigma{=}9$ with oracle keys is 3.86 AUC@5 (CI [1.45, 6.64]) on the low-$\rho$ scenes.
Making every visible frame dense also recovers less than oracle $3{\times}3$ windows do (courtyard 49.3 vs.\ 54.7), so on courtyard adding whole frames helps less than adding the few correct keys.

\paragraph{Refining the router later does not help.}
The router finds correspondences from the features of one layer, so a natural question is whether deeper features would find them better.
The best correspondence readout from a sparse pass recalls 0.450 of the $3{\times}3$ targets at layer 9 and also 0.450 at layer 12, so routing from deeper features gains nothing.

\subsection{Composition study details}
\label{sec:pilot}

This pre-registered composition study asks how the layer groups behave when each is made sparse alone and when all are made sparse together.

\begin{table}[t]
\centering
\small
\setlength{\tabcolsep}{2.4pt}
\setlength{\tabcolsep}{1.6pt}
\caption{\textbf{Composition study: swap-backs and single groups}. AUC@5 lost against dense, development scenes, seeds 0--1. Top: \methodname-fast with one group returned to dense attention. Bottom: only the named group sparse. Bold: in the TUM long-office and pooled low-$\rho$ columns, the matching-layer rows have the lowest loss of the top block and the highest of the bottom block.}
\label{tab:swap}
\begin{tabular}{@{}lcccccc@{}}
\toprule
\headrow Arm & fr1 & court & ETH-o & fr3 & 7S & low-$\rho$ \\
\midrule
\rowcolor{tabgroup}\multicolumn{7}{@{}l}{\emph{\methodname-fast with one group dense}} \\
layers 0--8 dense & 0.36 & 2.10 & 1.30 & 5.91 & 1.61 & 1.25 \\
layers 9--10 dense & 0.91 & 5.30 & 2.51 & 7.66 & 1.55 & 2.91 \\
\oursrow layers 11--16 dense & 0.10 & $-0.93$ & 1.17 & \textbf{0.00} & 0.77 & \textbf{0.11} \\
layers 17--23 dense & $-0.28$ & 6.11 & 2.29 & 4.68 & 1.67 & 2.71 \\
\rowcolor{tabgroup}\multicolumn{7}{@{}l}{\emph{only the named layers sparse}} \\
0--8 (256 hubs) & $-0.20$ & 1.29 & 0.56 & $-0.16$ & 0.28 & 0.55 \\
9--10 (1024 hubs) & $-0.58$ & $-0.48$ & 0.09 & $-0.14$ & $-0.46$ & $-0.33$ \\
11--16 ($\sigma{=}4$, blocks) & $-0.32$ & 4.15 & 1.77 & \textbf{5.40} & 1.33 & \textbf{1.87} \\
17--23 (1024 hubs) & 0.75 & 0.36 & 0.82 & 1.37 & 0.99 & 0.64 \\
\bottomrule
\end{tabular}
\end{table}

\begin{table}[t]
\centering
\small
\setlength{\tabcolsep}{3pt}
\caption{\textbf{AUC@30 lost against dense} in the composition study (development scenes; $\dagger$: seeds 0--1). Bold: lowest loss per column.}
\label{tab:pilot30}
\begin{tabular}{@{}lcccccc@{}}
\toprule
\headrow Layers 11--16 & fr1 & court & ETH-o & fr3 & 7S & low-$\rho$ \\
\midrule
exact (\methodname) & \textbf{1.41} & \textbf{0.24} & \textbf{0.81} & \textbf{0.14} & \textbf{0.55} & \textbf{0.82} \\
$\sigma{=}4$ + blocks$^\dagger$ & 4.06 & 2.88 & 7.01 & 0.95 & 1.93 & 4.65 \\
$\sigma{=}2$ + blocks & 2.76 & 1.21 & 2.68 & 0.37 & 0.58 & 2.22 \\
13--15 exact & 7.10 & 0.80 & 3.62 & 0.41 & 1.27 & 3.84 \\
only 11--16 sparse$^\dagger$ & 5.65 & 1.25 & 6.83 & 0.91 & 1.41 & 4.57 \\
\bottomrule
\end{tabular}
\end{table}

\begin{table}[t]
\centering
\small
\setlength{\tabcolsep}{3pt}
\caption{\textbf{\methodname per seed} (AUC@5 lost against dense, seeds 0--4, masked implementation). Bold: the two largest seed-level losses.}
\label{tab:perseed}
\begin{tabular}{@{}lccccc@{}}
\toprule
\headrow Scene & s0 & s1 & s2 & s3 & s4 \\
\midrule
TUM fr1\_room & 0.52 & $-0.32$ & 0.81 & \textbf{7.26} & $-1.81$ \\
ETH3D courtyard & $-3.51$ & 1.65 & \textbf{8.19} & 2.46 & $-3.31$ \\
ETH3D office & $-0.09$ & 2.42 & $-1.13$ & $-0.17$ & 3.55 \\
TUM fr3\_long\_office & $-2.50$ & 2.50 & $-0.52$ & 1.90 & 2.18 \\
7-Scenes office & 1.01 & 0.52 & 0.36 & 0.20 & $-0.32$ \\
\bottomrule
\end{tabular}
\end{table}

\begin{table}[t]
\centering
\small
\setlength{\tabcolsep}{1.20pt}
\caption{\textbf{Design study} (development scenes): AUC@5 lost against dense when layers 11--16 are exact or sparse and all other layers use hubs. fr1: TUM fr1\_room; court: ETH3D courtyard; ETH-o: ETH3D office; fr3: TUM fr3\_long\_office; 7S: 7-Scenes office; low-$\rho$: pooled over the first three; G$_{100}$: GFLOP per token at $N{=}100$. 5 seeds, $\dagger$: seeds 0--1. ``Blocks'': routed correspondence blocks. Red: above the pre-registered acceptance threshold (4 on the low-$\rho$ scenes, 3 on the others). Further arms: \cref{tab:swap}. Bold: the frozen recipe and its pooled low-$\rho$ loss, the lowest of all arms.}
\label{tab:composed}
\begin{tabular}{@{}lccccccc@{}}
\toprule
\headrow Layers 11--16 & fr1 & court & ETH-o & fr3 & 7S & low-$\rho$ & G$_{100}$ \\
\midrule
\oursrow \textbf{exact (\methodname)} & 1.29 & 1.10 & 0.92 & 0.71 & 0.35 & \pass{\textbf{1.10}} & 4.76 \\
$\sigma{=}4$ + blocks (fast)$^\dagger$ & $-1.21$ & \fail{11.55} & 1.73 & \fail{5.56} & 1.79 & 4.02 & 3.14 \\
$\sigma{=}9$ + blocks$^\dagger$ & \fail{6.41} & \fail{16.77} & 3.38 & \fail{7.44} & 2.96 & 8.85 & 2.82 \\
$\sigma{=}2$ + blocks & 1.99 & \fail{4.92} & 1.32 & 2.06 & 0.56 & 2.74 & 3.77 \\
13--15 exact, rest $\sigma{=}4$ & \fail{4.10} & 3.69 & 1.09 & 2.15 & 0.86 & 2.96 & 4.05 \\
14 exact, rest $\sigma{=}4$$^\dagger$ & 0.36 & \fail{10.28} & \fail{4.33} & \fail{4.50} & 1.53 & 4.99 & 3.44 \\
\bottomrule
\end{tabular}
\end{table}

\Cref{tab:swap} gives the arms behind \cref{sec:composed}: arms in which a single group is sparse, and arms in which one group of \methodname-fast is returned to dense attention.
Three results stand out.
The diffuse groups cost little when made sparse alone.
The sparse matching layers alone already exceed the thresholds on courtyard and TUM fr3\_long\_office.
Only returning layers 11--16 to exact attention removes most of the composed loss, which points to the matching layers as the group to keep exact.
\emph{Sequence of the study.} The study ran in three pre-registered steps, each with acceptance thresholds and a stop rule (\cref{sec:bars}).
(1) In the composed pilot (2 seeds), \methodname-fast lost 11.55 on courtyard (both seeds above 11.5), which fired the stop rule.
Returning layers 11--16 to dense attention recovered 3.66 AUC@5 pooled; this was the only group whose return recovered more than 2.5, so it was named as the group to repair.
(2) Repairs were tried in a fixed order, cheapest passing arm first.
A $\sigma{=}2$ lattice passed the 2-seed screen but then lost 4.92 on courtyard over 5 seeds.
Exact layers 13--15 lost 4.10 on TUM fr1\_room over 5 seeds.
Exact layer 14 alone lost 10.28 on courtyard already in the screen.
(3) The fallback, exact layers 11--16, passed every threshold on 5 seeds.
The later fused runs give a smaller courtyard loss for \methodname-fast than the masked pilot did.
In the fused 5-seed runs of \cref{tab:main}, \methodname-fast loses 5.9 on courtyard, close to the sum of its single-group losses (5.3); seed 1 alone loses 1.81 there, against 11.53 in the masked pilot.
\Cref{tab:pilot30} gives AUC@30 for the main rows of \cref{tab:composed}.
It shows that the AUC@30 loss of the sparse recipes also comes almost entirely from layers 11--16.
Depth is barely affected: AbsRel changes by at most 0.009 in every arm.
\Cref{tab:perseed} shows how much the loss of \methodname varies from seed to seed: the 5-seed means pass the bars, but two individual seeds lose more than 7 AUC@5.
The composed loss could in principle come from the sparse early layers disturbing the router of \methodname-fast, but they do not: vote recall is 0.6806 in the composed pass against 0.6805 in a dense pass, and the gated orphan share is 0.0668 against 0.0670.
The pilot's key selections (hub counts, orphan shares, frame-0 shares) are also nearly identical between the composed and single-group passes.
The extra loss of the composed pass therefore arises in the activations, not in a different choice of keys; per-layer errors were not measured.

\subsection{\methodname-fast: the routed variant}
\label{sec:fast}

This section describes \methodname-fast, the faster variant that routes correspondence keys instead of keeping the matching layers exact, and the evidence behind its switch.

\methodname-fast keeps the diffuse layers of \methodname and replaces the exact matching layers 11--16 by the sparse key set
\begin{equation}
  \mathcal K(t)=\mathcal F_0\cup\mathcal S\cup\Lambda_4\cup\{t\}\cup\textstyle\bigcup_{j\ne i}\mathcal B_j(u),
  \label{eq:band}
\end{equation}
plus a count-biased mean token over the remaining keys.
Here $\mathcal F_0$ is the reference frame, $\mathcal S$ the special tokens, $\Lambda_4$ one patch per $2{\times}2$ cell in every frame, and $t$ the query itself.
$\mathcal B_j(u)$ is the routed block of the query's $4{\times}8$ tile $u$ in another frame $j$; it is included only when its switch is on.
The query's own frame contributes only its lattice keys, because keeping the own frame dense was found to cost 7.9 AUC@5 on courtyard.

\paragraph{Router.}
The router finds, for each tile of queries, where the corresponding region lies in every other frame.
It runs once per pass on the input of layer 9's attention, with patch tokens normalised to unit length.
For every patch query and every other frame $j$, it finds the four most similar tokens by cosine similarity.
The hits of all queries in a tile then vote for a displacement in frame $j$; the vote histogram is smoothed with a $3{\times}3$ box, and its mode $\Delta^\star_j(u)$ is taken as the tile's displacement.
The block $\mathcal B_j(u)$ is the tile's footprint shifted by $\Delta^\star_j(u)$ and dilated by one token ($6{\times}10$ keys), and it is shared by all 32 queries of the tile.
An orphan query is one with no exact correspondence key in any view.
Without gating, the blocks cut the share of orphan queries from 0.371 / 0.193 / 0.288 to 0.035 / 0.021 / 0.076 on TUM fr1 / courtyard / ETH3D office at $\sigma{=}4$, so most queries regain a correspondence key.
The router's search grows as $N^2$ (54 / 214 / 521\,ms at $N{=}32$ / 64 / 100); letting only 8 of the 32 queries per tile vote still keeps block recall at 0.637.
The router's input is unaffected by the sparse early layers: its vote recall is 0.681 both in the composed pass and in a dense pass.
Not every routed block is right, and a wrong block can harm accuracy.
A decision-theoretic switch (\cref{sec:switch}) therefore keeps a block only when its chance of being right exceeds the break-even precision $p^\star=H/(H+og)$.
On courtyard $p^\star=3.4\%$, so even fairly unreliable blocks pay off; on redundant scenes ($o\to0$), $p^\star\to1$, so blocks are almost never worth keeping.

\subsubsection{Break-even switch}
\label{sec:switch}

\begin{figure}[t]
\centering
\includegraphics[width=\columnwidth]{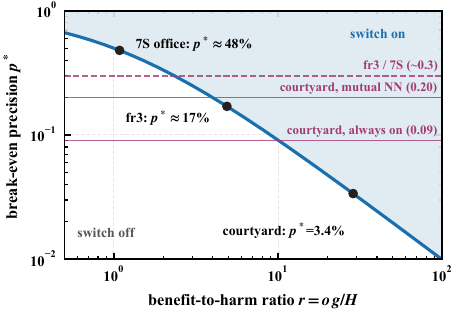}
\caption{\textbf{Break-even precision} $p^\star=1/(1+r)$ of a routed slot (\cref{eq:breakeven}) against the benefit-to-harm ratio $r=og/H$.
Black dots show the development scenes.
Horizontal lines show measured slot precisions; above the curve, switching a slot on pays.
On courtyard the per-slot rule predicts a net loss for always-on slots ($-1.05$ AUC@5, \cref{tab:signals}), so the switch helps by removing wrong slots.
On 7-Scenes office it should be off.}
\label{fig:breakeven}
\end{figure}

Consider one query and one routed slot with probability $p$ of holding the query's correspondence.
If the query is an orphan (probability $o$) and the slot is right, the slot removes it from the orphan share.
Pose accuracy rises by $g$ per unit of orphan share, so the expected gain is $p\,o\,g$.
If the slot is wrong (probability $1-p$), its keys are amplified like right ones and cost $H$.
The expected value is $p\,o\,g-(1-p)H$, and it is positive iff
\begin{equation}
  p>p^\star=\frac{H}{H+o\,g}.
  \label{eq:breakeven}
\end{equation}
The numerator is the cost of a wrong slot and $o\,g$ the expected gain of a right one.
The gain saturates per query, because one right block is enough to rescue an orphan, whereas the harm adds up over every wrong slot.
On redundant input there are almost no orphans ($o\to0$), so $p^\star\to1$ and the switch should stay off.
The three inputs $g$, $H$ and $o$ are estimated on courtyard at $\sigma{=}4$ from 5-seed runs.
Oracle correspondences remove an orphan share of 0.193 and gain 8.07 AUC@5, so $g=8.07/0.193=42$.
Similarity keys placed only in the 24.3 slots per query whose frame does not see the point cost 6.87, so $H=6.87/24.3=0.283$.
With $o=0.193$ this gives $p^\star=3.4\%$.
\Cref{fig:breakeven} shows the curve, and \cref{tab:signals} lists the switch signals measured on saved router outputs.
Of these signals, the vote-peak rule is the one used by \methodname-fast.
The break-even accounting is consistent with earlier runs: on courtyard it predicts net changes of $-1.05$ and $+0.3$ AUC@5 for ungated similarity windows and vote blocks, against $-0.69$ and $-0.02$ measured.
The model's predicted gain for the vote-peak switch ($\approx+3.8$) is only an estimate, because the switched blocks were measured only inside the composed recipe (\cref{tab:composed}), not on their own.

\begin{table*}[t]
\centering
\small
\setlength{\tabcolsep}{1.20pt}
\caption{\textbf{Switch signals} on saved layer-9 router outputs (courtyard unless stated). ``Right-slot precision'' counts a slot as right if its frame sees the point and its window holds the target. The last column is the net AUC@5 change on courtyard predicted by the break-even model, not a measurement. The vote-peak gate is a block-level rule; for it, and for the in-degree rule, only visibility precision is available. Bold: the rule used by \methodname-fast, the predicted net loss of always-on slots, and the highest predicted net among the signals with a right-slot precision.}
\label{tab:signals}
\begin{tabular}{@{}lccccc@{}}
\toprule
\headrow Signal & Slots kept & Visible kept & Right-slot precision, fr1 / court / ETH-o & Wrong / query & Predicted net \\
\midrule
Always on & 1.00 & 1.00 & 0.06 / 0.09 / 0.12 & 24.3 & $\mathbf{-1.05}$ \\
Mutual nearest neighbour at layer 9 & 0.21 & 0.38 & 0.20 / 0.20 / 0.36 & 3.9 & $\mathbf{+2.63}$ \\
Layer-12 attention top-1 agrees with hit & 0.25 & 0.36 & 0.20 / 0.20 / 0.33 & 5.5 & $+1.98$ \\
\quad same, inside the epipolar band & 0.29 & 0.43 & 0.23 / 0.20 / 0.36 & 6.0 & $+2.53$ \\
Mutual NN and layer-12 agreement & 0.07 & 0.17 & 0.41 / 0.35 / 0.59 & 1.2 & $+1.93$ \\
Hit lands on a non-hub (in-degree $\le2$) & 0.35 & 0.53 & visibility 0.32 & 7.4 & $+2.2$ \\
\textbf{Vote-peak share $\ge0.3$ (\methodname-fast)} & 0.42 & 0.68 & visibility 0.34 & 8.6 & $\approx+3.8$ (estimate) \\
\bottomrule
\end{tabular}
\end{table*}

\subsection{Thinner matching layers}
\label{sec:thinmatch}

This post-hoc study tests how far the matching layers can be thinned before accuracy suffers.

\emph{These tests are post hoc: their plans were registered after the results of \cref{tab:main} were in, before any of their own runs, with predictions and no pre-registered bar.}
They ask whether layers 11--16 can be thinned in other ways than the truncated lattices of \cref{sec:composed}.
In every arm, the other layers keep the key sets of \methodname (hub256 in layers 0--8, hub1024 in 9--10 and 17--23), so only the matching layers change.
The \emph{merge} arms apply the key merging of GPA-VGGT in layers 11--16 at merge ratios 0.2, 0.5 and 0.8, so that 80, 50 and 20\% of the keys remain.
As a check, the merge arm at ratio 0 reproduces \methodname bit for bit.
The \emph{top-$k$} arm uses, in layers 11--16 only, the per-query top-$k$ frame selection of the SAF3R heads, which keeps 21--27\% of the keys.
Its layers were checked to match SAF3R's own forward bit for bit.
Every arm ran on the 5 development and 21 unseen test scenes with 5 seeds and was scored as in \cref{tab:main} (\cref{tab:thinmatch}).

\begin{table*}[t]
\centering
\small
\setlength{\tabcolsep}{6pt}
\caption{\textbf{Thinner matching layers} (post hoc), $N{=}32$, 5 seeds. Keys: share of keys kept in layers 11--16. FLOPs: reduction against dense attention at $N{=}100$. Loss: AUC@5 lost against dense attention on the 13 unseen low-$\rho$ scenes. $\Delta$: paired AUC@5 difference to \methodname (negative = less accurate) with hierarchical 95\% CIs. Red: CI below zero. Bold: the lowest and highest loss, and the only thinner arm that is not significantly worse on the low-$\rho$ scenes.}
\label{tab:thinmatch}
\begin{tabular}{@{}lcccccc@{}}
\toprule
\headrow Layers 11--16 & Keys & FLOPs & Loss & $\Delta$ low-$\rho$ & $\Delta$ high-$\rho$ \\
\midrule
\oursrow Exact (\methodname) & 100\% & 2.58$\times$ & \textbf{1.38} & -- & -- \\
Merge 0.2 & 80\% & 2.86$\times$ & 1.86 & $\mathbf{-0.49}$ [$-1.24$, 0.12] & $+0.06$ [$-0.35$, 0.49] \\
Merge 0.5 & 50\% & 3.49$\times$ & 3.00 & \fail{$-1.62$} [$-2.53$, $-0.80$] & $+0.22$ [$-0.20$, 0.67] \\
Merge 0.8 & 20\% & 4.25$\times$ & 7.11 & \fail{$-5.74$} [$-7.57$, $-3.93$] & $-0.43$ [$-1.14$, 0.13] \\
Top-$k$ frames & 21--27\% & 4.34$\times$ & 5.60 & \fail{$-4.22$} [$-5.70$, $-2.91$] & $-0.75$ [$-1.77$, 0.33] \\
Shared hubs (1024) & own frame + 1024 & 5.40$\times$ & \textbf{29.17} & \fail{$-27.79$} [$-34.62$, $-21.22$] & \fail{$-14.88$} [$-23.04$, $-7.49$] \\
\bottomrule
\end{tabular}
\end{table*}

On the low-$\rho$ scenes, accuracy falls steadily as more keys are removed, relative to \methodname:
\begin{itemize}
  \item keeping 80\% of the keys is not significantly worse than exact attention;
  \item keeping 50\% costs 1.6 AUC@5, and keeping 20\% costs 5.7;
  \item top-$k$ frame selection costs 4.2, which is no better than SAF3R with its own early and late layers ($-1.07$, CI [$-3.58$, 1.48]).
\end{itemize}
The most extreme arm replaces layers 11--16 by the shared hub keys of the diffuse layers, so that no layer stays exact.
It costs 27.8 AUC@5 on the low-$\rho$ scenes and 14.9 even on the high-$\rho$ scenes, yet its aggregator runs only 1.1, 1.4 and 1.5$\times$ faster than that of \methodname at $N{=}32$, 64 and 100 (same session).
This means that the matching layers depend on query-specific keys, which a shared sample cannot provide, and that keeping these layers exact is cheap relative to their effect on accuracy.
On the high-$\rho$ scenes, all other arms stay within 0.8 AUC@5 of \methodname, so thinning matters mainly where views overlap little.
The three arms that use over 4$\times$ fewer FLOPs than dense attention at $N{=}100$ lose 4.2, 5.7 and 27.8 AUC@5 against \methodname on the low-$\rho$ scenes.
Every arm that keeps accuracy there retains at least 80\% of the keys in the matching layers, so its cost stays quadratic in $N$.
In other words, no tested way of thinning the matching layers both keeps accuracy and removes their quadratic cost.
A final arm replaces the early and late layers of SAF3R by the key sets of \methodname.
It recovers none of the baseline's low-$\rho$ loss, so that loss therefore sits in its matching layers.

\subsection{Why the matching layers resist thinning}
\label{sec:broad}

This post-hoc analysis asks why the matching layers resist thinning, by looking directly at their attention.

\emph{This attention audit is post hoc. Its plan, screening rule and amendments were registered before its GPU runs, and its results gate no claim.}
It asks whether any \emph{selection} of keys, that is, any rule that keeps a small subset of keys exactly, could replace exact attention in layers 11--16.
The candidate rules borrow ideas from other fields:
\begin{itemize}
  \item an epipolar band around each query's line in every other frame, computed from ground-truth poses (an oracle);
  \item reuse of each query's heaviest keys across layers;
  \item sharing heavy keys among the queries of a $4{\times}8$ tile;
  \item block-wise upper bounds on $q^\top k$;
  \item a per-(query, head) switch between exact attention and hubs driven by $s^2$.
\end{itemize}
In each case, the keys outside the selected set were covered by the unbiased hub estimator of \methodname, so a rule only has to find the keys that matter most.
To evaluate the rules, exact fp32 attention was computed on the 5 development scenes with 2 seeds, for 2048 random patch queries in each of the 16 heads of layers 11--16.
A rule was screened as viable if, even for the worst 10\% of (query, head) pairs, it kept at least 90\% of the attention mass with at most 25\% of the keys.

No rule passed in any of the six layers.
\begin{itemize}
  \item \textbf{Epipolar band.} Even with true poses, a band of $\pm5$ patches (15\% of the keys) kept only 21--52\% of the mass of the median peaked pair, where a peaked pair is one with $\Gamma_{\text{row}}\ge1.5$.
  \item \textbf{Each query's own 256 heaviest keys} (an oracle that no practical rule can beat) kept only about 60\%.
  \item \textbf{Reuse across layers.} Heavy keys do not persist from layer to layer: a query's heaviest cross-frame key matched that of layer 11 in under 2\% of cases, so keys chosen in one layer are a poor guess for another.
  \item \textbf{Block bounds and the $s^2$ switch.} Both reach exact accuracy, but only by keeping 51--99\% of the keys, which saves at most 12\% of the FLOPs of \methodname.
\end{itemize}
The switch uses the proxy $s^2\ge3$ in place of $\Gamma_{\text{row}}\ge1.5$; the two agree on 89--96\% of (query, head) decisions ($\kappa=0.71$--$0.91$).

\Cref{tab:broad} shows why these rules fail.
Even where attention is peaked, it is spread over many keys: half the mass of a peaked pair lies on 48--269 keys, and 90\% lies on 349--2540, out of about 33{,}000.
Most of this mass falls on other frames, and the geometric structure, although visible, is small.
On peaked pairs, the $3{\times}3$ window around the true correspondence is enriched up to 74-fold in layer 14, meaning that it receives far more attention than its size alone would give it, yet it holds only 1.6--12\% of the cross-frame mass; a $\pm9$-patch epipolar band (16\% of the keys) holds 29--56\%.
In contrast, layers 10 and 17, which \methodname serves with hubs, are mostly diffuse and show no such concentration.
The matching layers thus sit between the two cases that cheap approximations handle well: for the approximations tried, their attention is too broad for any small key set tested, and too focused for a shared sample.
The bound of about 4$\times$ is therefore an observed bottleneck under the training-free approximations tried, not a lower bound on sparse matching attention.
It suggests that learned compression, rather than selection, is the route past it.

\begin{table*}[t]
\centering
\small
\setlength{\tabcolsep}{5pt}
\caption{\textbf{Breadth of attention in the matching layers} (post hoc; 5 development scenes $\times$ 2 seeds, 2048 random queries, all heads). Peaked: (query, head) pairs with $\Gamma_{\text{row}}\ge1.5$. $n_{\text{eff}}$: exponential of the attention entropy, median. $k_{50}$, $k_{90}$: median number of keys holding 50\% and 90\% of the mass. Band and window: share of the cross-frame mass inside the ground-truth epipolar band ($\pm9$ patches, 16\% of cross-frame keys) and the $3{\times}3$ correspondence window (0.17\%). Bold: the most concentrated matching layer, which still needs hundreds of keys to hold most of its mass.}
\label{tab:broad}
\begin{tabular}{@{}lcccccccc@{}}
\toprule
\headrow Layer & Peaked share & $n_{\text{eff}}$ (peaked) & $k_{50}$ & $k_{90}$ & Cross-frame & Band $\pm9$ & $3{\times}3$ window & $n_{\text{eff}}$ (diffuse) \\
\midrule
10 (hubs) & 0.22 & 3480 & 491 & 5806 & 0.93 & 0.27 & 0.016 & 10231 \\
\oursrow 11 & 0.77 & 1121 & 158 & 2061 & 0.88 & 0.32 & 0.045 & 10537 \\
\oursrow 12 & 0.79 & 1510 & 223 & 2540 & 0.90 & 0.38 & 0.044 & 8972 \\
\oursrow 13 & 0.73 & 1138 & 172 & 1734 & 0.87 & 0.47 & 0.079 & 12590 \\
\oursrow 14 & \textbf{0.91} & \textbf{284} & \textbf{48} & \textbf{349} & 0.85 & \textbf{0.56} & \textbf{0.120} & 10437 \\
\oursrow 15 & 0.87 & 1007 & 171 & 1302 & 0.89 & 0.43 & 0.050 & 11332 \\
\oursrow 16 & 0.60 & 1760 & 269 & 2478 & 0.92 & 0.29 & 0.016 & 11921 \\
17 (hubs) & 0.08 & 2775 & 419 & 3944 & 0.93 & 0.21 & 0.003 & 18871 \\
\bottomrule
\end{tabular}
\end{table*}

\section{Selector validation}
\subsection{Equal-cost windows of exact layers}
\label{sec:windows}

This registered control, post hoc relative to the main table, compares the selected window with four shifted windows.

\emph{This control is post hoc: it was added after the main results, but its plan, arms and decision rules were registered before it ran, and no arm could replace the method.}
The ablation (\cref{sec:ablation}) shows that layers 11--16 beat fewer or more exact layers, but that comparison changes the cost as well as the placement.
This control therefore asks a sharper question: do the layers chosen by the statistic beat other placements of six exact layers at exactly the same cost?
Each arm keeps six consecutive global layers exact and treats every other layer as \methodname does (256 hubs in layers 0--8, 1024 elsewhere).
As a result, all arms cost 2.993 GFLOP per token at $N{=}32$ and 4.758 at $N{=}100$, and they differ only in where the exact layers sit.
The arm with layers 11--16 is \methodname itself; its 65 test runs reproduce the main-table predictions bit for bit.
The registered outcome rule labels layers 11--16 ``best'' if they have the lowest mean loss and the 95\% CI of every other arm's difference to them excludes 0.

\begin{table}[t]
\centering
\footnotesize
\setlength{\tabcolsep}{3pt}
\caption{\textbf{Equal-cost windows of six exact layers} (AUC@5 lost against dense attention, mean and 95\% hierarchical-bootstrap CI; $\Delta$: arm minus layers 11--16). Development: 5 scenes $\times$ 5 seeds; test: the 13 unseen low-$\rho$ scenes $\times$ 5 seeds. All arms: 2.99 GFLOP per token at $N{=}32$. Bold: the selected window, which has the lowest loss.}
\label{tab:windows}
\begin{tabular}{@{}lcccc@{}}
\toprule
 & \multicolumn{2}{c}{Development (5)} & \multicolumn{2}{c}{Unseen low-$\rho$ (13)} \\
\cmidrule(lr){2-3}\cmidrule(lr){4-5}
Exact layers & Loss & $\Delta$ & Loss & $\Delta$ \\
\midrule
9--14 & 7.60 & $+7.07$ [2.70, 13.10] & 8.51 & $+7.14$ [5.36, 9.18] \\
10--15 & 2.89 & $+2.36$ [0.80, 4.89] & 4.19 & $+2.81$ [1.68, 4.11] \\
\textbf{11--16} (\methodname) & \textbf{0.53} & -- & \textbf{1.38} & -- \\
12--17 & 3.03 & $+2.50$ [0.31, 4.73] & 2.72 & $+1.34$ [0.49, 2.31] \\
13--18 & 6.09 & $+5.56$ [0.72, 12.31] & 4.61 & $+3.24$ [1.91, 4.85] \\
\bottomrule
\end{tabular}
\end{table}

Layers 11--16 meet the rule on both the development and the unseen test scenes, so the registered verdict is that the selected window is best among the five tested.
All four differences remain significant after Holm correction over the four comparisons (largest adjusted $p$ 0.001 on the test scenes, 0.025 on the development scenes, hierarchical bootstrap).
The cost of a misplaced window grows with the shift: shifting the window by one layer costs 1.3 to 2.8 AUC@5 on the test scenes, and shifting it by two layers costs 3.2 to 7.1.
AUC@30 gives the same ordering, with every difference excluding 0.
For depth, the windows 12--17 and 13--18 are not significantly different from 11--16 (AbsRel$\times100$ $-0.10$, CI [$-0.33$, 0.07], and $-0.06$, CI [$-0.39$, 0.18]), whereas 9--14 and 10--15 are worse.
The summed $\Gbar_\ell$ inside each window was recorded before the runs (11--16: 23.4; 10--15: 22.5; 12--17: 21.2; 9--14: 19.2; 13--18: 18.5).
The accuracy ranking follows this sum, with two exceptions: 9--14 and 13--18 swap places, and on the test scenes 10--15 and 12--17 also swap.

\subsection{Simpler layer statistics as selectors}
\label{sec:selectors}

This control asks whether simpler per-layer statistics, given the same budget of six exact layers, choose layers as good as those chosen by $\Gbar_\ell$.

\emph{This control is post hoc: it was added after the main results, but its statistic definitions, selection rule, arms and decision rules were registered before it ran, and no arm could replace the method.}
For each query and head, the statistics are computed over the patch keys $X$ of the other frames:
\begin{itemize}
  \item the row-free logit variance $s^2$;
  \item the attention entropy $H_X$ of the softmax restricted to $X$;
  \item the effective number of keys $n_{\mathrm{eff}}$, the exponential of the entropy over all keys;
  \item the share of attention mass on $X$.
\end{itemize}
The first three measure how concentrated a layer's attention is; the last measures how much attention goes to other frames.
They are aggregated in the same way as $\Gbar_\ell$ (mean over 2048 random queries, median over heads, mean over 2 seeds and the 5 development scenes).
Each statistic keeps exact the six layers it ranks first: highest $s^2$, lowest $H_X$, lowest $n_{\mathrm{eff}}$, or highest cross-frame mass.
Every other layer is treated as in \methodname.
Computing these statistics needed one dense measurement pass, which computes no accuracy metric.
The selected sets were frozen after that pass and before any accuracy run, so no selection could be tuned on accuracy.

The three concentration statistics select exactly layers 11--16, both on the pooled development scenes and on each of the 5 scenes separately.
Their rank correlations with $\Gbar_\ell$ over the 24 layers are 0.995 ($s^2$), $-0.996$ ($H_X$) and $-0.994$ ($n_{\mathrm{eff}}$); the negative signs only reflect that low entropy and few effective keys mean high concentration.
This agreement is expected, because all three measure the same property: $\Gamma$ is the divergence $\mathrm{KL}(U_X\,\|\,p_X)$ of the attention from uniform attention, and $\ln|X|-H_X$ is the reverse divergence.
Cross-frame mass, in contrast, selects layers 0, 1, 4, 8, 22 and 23, which have no layer in common with 11--16.
It is even anti-correlated with $\Gbar_\ell$ (Spearman $-0.82$), because the matching layers put slightly more of their mass on their own frame.

\begin{table}[t]
\centering
\footnotesize
\setlength{\tabcolsep}{2pt}
\caption{\textbf{Mass-selected against $\Gbar$-selected exact layers} (AUC@5 lost against dense attention, mean and 95\% hierarchical-bootstrap CI; $\Delta$: arm minus layers 11--16). Development: 5 scenes $\times$ 5 seeds; test: the 13 unseen low-$\rho$ scenes $\times$ 5 seeds. GFLOP per token at $N{=}32$: 2.993 (11--16) and 3.005 (mass). Bold: the layers selected by $\Gbar$, which have the lower loss.}
\label{tab:selectors}
\begin{tabular}{@{}lcc@{}}
\toprule
Exact layers & Dev.\ (5) & Test (13) \\
\midrule
\textbf{11--16} ($\Gbar$, $s^2$, $H_X$, $n_{\mathrm{eff}}$) & \textbf{0.53} & \textbf{1.38} \\
0, 1, 4, 8, 22, 23 (mass) & 25.21 & 28.91 \\
$\Delta$ & $+24.68$ [13.76, 39.68] & $+27.54$ [21.03, 34.42] \\
\bottomrule
\end{tabular}
\end{table}

Since the three concentration statistics pick the same layers as $\Gbar_\ell$, only the mass-selected set needed an accuracy run.
The mass-selected layers cost 0.4\% more than the selected layers and lose far more accuracy (\cref{tab:selectors}).
They are worse on all 13 test scenes and all 5 development scenes, and the difference also holds for AUC@30 ($+20.65$, CI [12.73, 28.98], test) and AbsRel.
With only one comparison, Holm correction leaves $p$ at the bootstrap floor of 0.0002.
The registered verdict is therefore that layers 11--16 are best.
The run with layers 11--16 reproduces the main-table predictions bit for bit.
This control shows that the selection follows from how concentrated a layer's attention is, a property that several equivalent statistics measure.
Selecting by cross-frame attention mass, by contrast, keeps exact the layers that the tested recipe can thin, and leaves the layers that need exact attention to be thinned.
One limit remains: the control does not test a per-layer sensitivity selector, because no single-layer thinning measurement exists for all 24 layers.

\subsection{The selector applied to GPA-VGGT}
\label{sec:gpasel}

This post-hoc test asks whether the selector helps GPA-VGGT on the same backbone.

\emph{This test is post hoc: its plan was registered after the results of \cref{tab:main} were in, before any of its own runs, and it has no pre-registered bar.}
It asks whether the two parts of \methodname that can be moved to another method, the layer selector and the hubs, improve a published training-free baseline.
GPA-VGGT already protects layers 13--16 by merging only 20\% of their keys, and it summarises keys more coarsely in the other layers.
Three arms modify it:
\begin{itemize}
  \item \emph{GPA+sel} extends this protection to the matching set that the statistic selects, layers 11--16; this is a change of configuration only;
  \item \emph{GPA+hubs} replaces the early cluster summaries of GPA-VGGT in layers 0--8 by the hub256 key sets of \methodname;
  \item \emph{GPA+sel+hubs} applies both changes.
\end{itemize}
Each arm ran on the 5 development and the 13 unseen low-$\rho$ scenes with 5 seeds and was scored as in \cref{tab:main} (\cref{tab:gpasel}).

\begin{table*}[t]
\centering
\small
\setlength{\tabcolsep}{6pt}
\caption{\textbf{The selector applied to GPA-VGGT} (post hoc), 13 unseen low-$\rho$ scenes, $N{=}32$, 5 seeds. Loss: AUC@5 lost against dense attention; $\Delta$: paired AUC@5 difference (positive = the arm is more accurate), with hierarchical 95\% CIs. Bold: the lowest loss and AbsRel, and the gain of GPA+sel over GPA-VGGT, the main result of this test.}
\label{tab:gpasel}
\begin{tabular}{@{}lccccc@{}}
\toprule
\headrow Arm & Loss & $\Delta$ vs GPA-VGGT & $\Delta$ vs \methodname & AbsRel & GFLOP/tok \\
\midrule
GPA-VGGT & 2.42 & -- & $-1.04$ & .0465 & 2.96 \\
\oursrow \methodname & 1.38 & $+1.04$ [$-0.34$, 2.54] & -- & .0468 & 2.99 \\
GPA+sel & 1.21 & \textbf{+1.20} [0.41, 2.12] & $+0.16$ [$-0.73$, 1.00] & \textbf{.0451} & 3.10 \\
GPA+hubs & 2.26 & $+0.15$ [$-0.30$, 0.60] & $-0.89$ [$-2.33$, 0.31] & .0494 & 2.96 \\
GPA+sel+hubs & \textbf{1.13} & $+1.28$ [0.03, 2.55] & $+0.24$ [$-0.41$, 0.90] & .0480 & 3.10 \\
\bottomrule
\end{tabular}
\end{table*}

Protecting the selected set improves GPA-VGGT by 1.20 AUC@5 and brings it level with \methodname, at the price of 5\% more FLOPs and 1--6\% more aggregator time at $N{=}32$--100.
The development scenes agree ($+1.34$, CI [0.14, 2.90]).
The hubs alone, in contrast, leave GPA-VGGT unchanged ($+0.15$), so in this baseline the gain comes from the choice of layers, not from the early estimator.
The test therefore supports the statistic as a layer selector that transfers to another method on the same backbone.
It does not show that hubs are better than the early cluster summaries of GPA-VGGT.

\subsection{Component ablation}
\label{sec:ablation}

This post-hoc ablation measures how much each component of the frozen recipe contributes, by changing it and measuring the loss.

\emph{The development-scene arms were planned before the test runs. The test-scene arms are post hoc, added after the main results, and have no pass/fail rule.}
Each arm changes one component of the frozen recipe and keeps everything else unchanged (\cref{tab:ablation}), so any change in loss can be attributed to that component.

\begin{table*}[t]
\centering
\small
\setlength{\tabcolsep}{5pt}
\caption{\textbf{Component ablation}, $N{=}32$, 5 seeds. AUC@5 lost against dense attention, with hierarchical 95\% CIs. Dev: 5 development scenes. Test: the 13 unseen low-$\rho$ scenes (post hoc). Each row changes one component of the frozen recipe (first row). Bold: the frozen recipe and the lowest test loss.}
\label{tab:ablation}
\begin{tabular}{@{}llccc@{}}
\toprule
\headrow Component & Variant & Dev loss & Low-$\rho$ test loss & GFLOP/token \\
\midrule
\oursrow \multicolumn{2}{@{}l}{\textbf{\methodname (frozen recipe)}} & \textbf{0.54} [$-0.61$, 1.81] & \textbf{1.38} [0.71, 2.01] & 2.99 \\
\midrule
Early layers 0--8 & own frame only & \fail{8.41} [0.10, 22.26] & -- & 2.97 \\
 & own frame + special tokens & \fail{5.69} [0.24, 14.18] & -- & 2.97 \\
 & + count-biased mean token & 0.95 [$-0.69$, 3.03] & -- & 2.97 \\
\midrule
Hub estimator & uniform proposal (weights kept) & 1.34 [0.00, 3.64] & \fail{3.34} [2.26, 4.74] & 2.97 \\
 & no log-weight correction & \fail{24.22} [9.73, 38.90] & \fail{36.17} [25.88, 46.31] & 2.99 \\
 & systematic / heavy-tail / quantile draws & 0.61--0.86 & -- & 2.99 \\
\midrule
Hub count & 128 / 512 & 2.36 [0.69, 5.40] & -- & 2.97 \\
 & 1024 / 2048 & 0.62 [$-0.12$, 1.43] & \textbf{0.66} [0.00, 1.26] & 3.06 \\
\midrule
Late layers 17--23 & AVGGT(9) & \fail{5.12} [1.41, 9.85] & -- & 3.07 \\
 & 256 hubs & 3.14 [0.99, 6.73] & -- & 2.97 \\
 & frame 0 exact & 0.93 [0.05, 1.87] & -- & 3.03 \\
\midrule
Layers 9--10 & dense & 0.77 [0.00, 1.70] & -- & 3.25 \\
 & AVGGT(9) & 1.26 [0.34, 2.39] & -- & 3.02 \\
\midrule
Matching layers 11--16 & shared hubs (1024) & \fail{25.56} [14.06, 41.46] & \fail{29.17} [22.41, 36.17] & 2.24 \\
 & $\sigma{=}4$ lattice + tubes (\methodname-fast) & 3.91 [2.04, 5.71] & 4.65 & 2.49 \\
 & $\sigma{=}4$ lattice, no tubes & 4.38 [1.81, 6.70] & -- & 2.41 \\
 & $\sigma{=}2$ lattice + tubes & 1.96 [0.83, 3.47] & -- & 2.68 \\
\midrule
Selector cut-off & 1.0 nats (exact 10--16) & 1.61 [0.28, 3.54] & -- & 3.12 \\
 & 2.0 nats (exact 11--15) & 3.90 [2.15, 6.40] & \fail{4.27} [3.12, 5.61] & 2.87 \\
 & 3.0 nats (exact 14, 15) & \fail{12.48} [3.30, 25.53] & \fail{12.34} [9.73, 15.45] & 2.49 \\
\bottomrule
\end{tabular}
\end{table*}

The ablation separates the contributions as follows.
\begin{itemize}
  \item \textbf{Importance-weight correction.} The hubs are drawn non-uniformly, so each must be reweighted by its log-weight. Without this correction the loss rises to 24 AUC@5 on the development scenes and 36 on the test scenes, so the hubs are useless without it.
  \item \textbf{Pilot proposal.} Drawing hubs uniformly, while keeping the correct weights, costs 2.0 more on the test scenes than drawing them from the pilot's column mass. Variance-reducing draws (systematic, heavy-tail, quantile) change little.
  \item \textbf{Early layers.} They need cross-frame context: restricting them to the own frame, with or without special tokens, is costly. A count-biased mean token recovers most of the loss, and the hubs recover the rest.
  \item \textbf{Later diffuse layers.} In layers 17--23, 1024 hubs beat both 256 hubs and AVGGT's lattice. Making layers 9--10 dense brings no gain for 8\% more FLOPs.
  \item \textbf{Matching layers.} Every sparse key set tested in the matching layers costs 1.4--25 more than exact attention, and tubes (routed correspondence windows) recover only part of what a lattice loses.
  \item \textbf{Selector cut-off.} The 1.5-nat cut-off sits at the knee of the accuracy curve. Raising it to 2.0 nats removes layer 16 from the exact set and costs 2.9 on the test scenes; raising it to 3.0 nats keeps only layers 14 and 15 exact and costs 11.0. Lowering it to 1.0 adds layer 10 to the exact set without benefit.
\end{itemize}

\section{Robustness and generality}
\subsection{Post-hoc robustness analysis}
\label{sec:robust}

This post-hoc analysis tests whether the main conclusions are robust to how redundancy is measured, to which datasets are included, and to the worst scenes rather than the mean.

\emph{This analysis is post hoc: it was not part of the pre-registered plan, was run after all results of \cref{tab:main} were in, and has no pre-registered bar.}
It asks how the per-scene difference between \methodname and the two strongest baselines depends on redundancy, and how large each method's tail losses are.
A scene's $\rho$ is the mean over the frame subsets on which it was measured (one to five per scene; \cref{sec:protocol}), and every per-scene value is a 5-seed mean.

\emph{Redundancy from ground truth.}
$\rho$ is computed from the dense model's own depth and cameras.
A poorly reconstructed scene could therefore look less redundant than it really is, which could place it on the wrong side of the split.
To check this, $\rho$ was recomputed with the same code from ground-truth depth and poses, which are available for ETH3D, TUM, 7-Scenes and DTU but not for Tanks and Temples.
The ground-truth $\rho$ has a Spearman correlation of 0.98 with the model-based $\rho$ over 58 frame subsets (0.99 over the 21 scenes with ground truth, 16 test and 5 development), so the two measures rank scenes almost identically.
No unseen low-$\rho$ scene crosses the 10.5 split; the highest is ETH3D relief at 9.75.
One high-$\rho$ test scene, 7-Scenes pumpkin, falls just below the split (from 12.60 to 10.36), partly because Kinect depth has holes; it stays above the split when targets without depth count as visible (12.10).
On the 16 test scenes with ground truth, the correlation of the gain over GPA-VGGT with $\rho$ is $-0.55$ (CI [$-0.78$, $-0.12$]) with the model-based $\rho$ and $-0.48$ (CI [$-0.72$, $-0.01$]) with the ground-truth $\rho$.
With ground-truth $\rho$, the mean gain is $+1.1$ below the split and $-0.6$ above it, so the pattern is similar with either measure of $\rho$.

\emph{Leaving out one dataset.}
This check asks whether any single dataset drives the result.
On the 13 unseen low-$\rho$ scenes, the loss of \methodname against dense attention stays between 1.07 and 1.52 AUC@5 when ETH3D (8 scenes), Tanks and Temples (4) or TUM (1) is left out, so the loss itself does not depend on one dataset.
The gains over the baselines do.
The gain over GPA-VGGT comes mainly from ETH3D: it is 1.44 on ETH3D alone, and 0.39 (CI [$-0.83$, 2.14]) without it.
The gain over SAF3R, in contrast, comes mainly from Tanks and Temples and TUM: it is 5.69 (CI [3.05, 9.60]) without ETH3D, and its CI includes 0 without either of them.

\begin{figure*}[t]
\centering
\includegraphics[width=\textwidth]{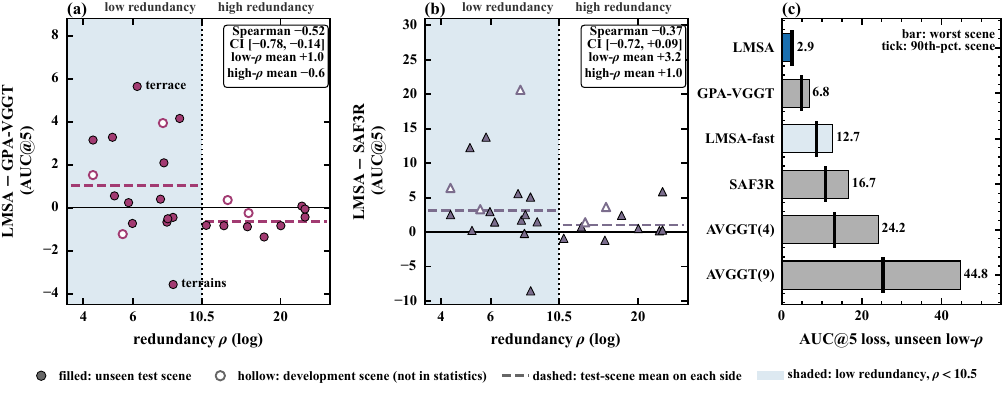}
\caption{\textbf{Post-hoc robustness analysis} (not pre-registered). \textbf{(a)}~AUC@5 of \methodname minus that of GPA-VGGT on the 21 test scenes (filled) and the 5 development scenes (open) of \cref{tab:main} against the scene's redundancy $\rho$ (log axis); shaded: $\rho<10.5$; dashed lines and box: mean on each side of the split and Spearman correlation with a bootstrap 95\% CI, over the test scenes only. \textbf{(b)}~The same against SAF3R. \textbf{(c)}~Loss against dense attention on the 13 unseen low-$\rho$ scenes: worst scene (bar) and 90th-percentile scene (tick).}
\label{fig:robust}
\end{figure*}

\begin{table*}[t]
\centering
\small
\setlength{\tabcolsep}{5pt}
\caption{\textbf{Tail losses} (post hoc): AUC@5 lost against dense attention. Scene losses use 5-seed means; run losses use single scene-seed runs (65 on the unseen low-$\rho$ scenes; ``worst 5\%'' is the mean of the 4 worst). Dense loses 0 by definition. Bold: lowest per column.}
\label{tab:tails}
\begin{tabular}{@{}lcccccc@{}}
\toprule
\headrow & \multicolumn{5}{c}{Unseen low-$\rho$ (13 scenes)} & All 26 scenes \\
\headrow Method & mean & 90th pct.\ scene & worst scene & 95th pct.\ run & worst 5\% of runs & worst scene \\
\midrule
AVGGT ($\sigma{=}4$) & 10.3 & 13.1 & 24.2 (terrace) & 20.9 & 25.2 & 37.5 (courtyard) \\
AVGGT ($\sigma{=}9$) & 18.9 & 25.3 & 44.8 (terrace) & 38.0 & 47.0 & 50.4 (courtyard) \\
GPA-VGGT & 2.4 & 4.9 & 6.8 (terrace) & 6.3 & 10.2 & 6.8 (terrace) \\
SAF3R & 4.5 & 10.9 & 16.7 (electro) & 18.3 & 22.6 & 22.0 (courtyard) \\
\oursrow \methodname & \textbf{1.4} & \textbf{2.5} & \textbf{2.9 (electro)} & \textbf{5.2} & \textbf{5.9} & \textbf{2.9 (electro)} \\
\methodname-fast & 4.6 & 8.7 & 12.7 (terrace) & 10.2 & 14.5 & 12.7 (terrace) \\

\bottomrule
\end{tabular}
\end{table*}

\emph{Dependence on redundancy.}
On the 21 test scenes, the per-scene difference between \methodname and GPA-VGGT falls as $\rho$ grows (\cref{fig:robust}a; Spearman $-0.52$, bootstrap 95\% CI $[-0.78, -0.14]$).
Below the split $\rho=10.5$, \methodname is ahead by 1.0 AUC@5 on average (13 scenes, ahead on 8); above it, \methodname is behind by 0.6 (8 scenes, ahead on 1).
With the 5 development scenes included, the correlation is $-0.43$ (CI $[-0.69, -0.07]$).
This correlation reflects a contrast between the two sides of the split rather than a trend within either side: within each side it is $-0.18$ below and $+0.21$ above, and neither is significant.
Against SAF3R, the test-scene correlation is weaker and its CI includes 0 (\cref{fig:robust}b; $-0.37$, CI $[-0.72, 0.09]$; means $+3.2$ below the split and $+1.0$ above).
\emph{Tail losses.}
A user who runs a method on one scene cares about its worst case as well as its mean, and the methods differ more in their tails than in their means (\cref{tab:tails}, \cref{fig:robust}c).
On the 13 unseen low-$\rho$ scenes, the worst scene costs \methodname 2.9 AUC@5 (ETH3D electro) and costs GPA-VGGT 6.8 (terrace); the worst 5\% of single runs cost them 5.9 and 10.2.
On the six scenes on which GPA-VGGT itself loses at least 3 AUC@5 (ETH3D electro, playground, relief, relief\_2 and terrace, and courtyard), \methodname is ahead of it by 0.2 to 5.6, so \methodname is ahead where GPA-VGGT struggles most.
The whole analysis is descriptive: the split $\rho=10.5$ was fixed by the protocol (\cref{sec:protocol}), but the comparisons themselves were chosen after the results were known, so they are not tests of a prior hypothesis.

\subsection{Sources of run-to-run variance}
\label{sec:variance}

The loss of \methodname varies from run to run for two reasons: the random hub draw, and the choice of frame subset.
This post-hoc analysis measures how much of the variation each source causes.

On the five development scenes, \methodname was rerun with two frame subsets per scene and three hub draws per subset, giving 30 runs; a new draw changes the seed of both the hub pilot and the hub sample.
A nested method-of-moments decomposition then splits the variance of the per-run AUC@5 loss into its sources (\cref{tab:variance}).
It attributes 84\% of the variance to the hub draw (SD 2.28), 16\% to the frame subset (SD 1.00) and none to the scene, so hub sampling noise is the main source.
Most of the draw variance comes from one courtyard subset, whose three draws span 12.7 AUC@5; without courtyard, the draw SD is 1.14 (70\%).
Over the 21 test scenes, with one draw per subset, the scene term explains 3.5\% of the per-run variance.
Since the noise comes from sampling, drawing more hubs should reduce it.
More hubs (1024 in layers 0--8 and 2048 in layers 9--10 and 17--23) indeed lower the draw SD by 44\%, from 2.28 to 1.27, at an unchanged mean loss (0.36 against 0.39) and for 2.2\% more FLOPs at $N{=}32$.
With more hubs, the frame subset becomes the dominant source of variation.

\emph{More hubs on the test scenes (post hoc).} The same variant with more hubs was then run on all 21 test scenes with 5 seeds.
It loses 0.66 AUC@5 on the unseen low-$\rho$ scenes (CI [0.00, 1.26]) and 0.20 on the high-$\rho$ scenes (CI [0.00, 0.43]), against 1.38 and 0.73 for \methodname.
Its paired gains over \methodname (0.72, CI [$-0.16$, 1.66], and 0.53, CI [$-0.03$, 1.17]) are not significant, however.
On the low-$\rho$ scenes, the variant leads GPA-VGGT by 1.76 AUC@5 (CI [0.60, 3.17]) and 1.45 AUC@30 (CI [0.44, 2.78]); on the high-$\rho$ scenes, it does not differ significantly from GPA-VGGT ($-0.12$, CI [$-0.54$, 0.34]).
Against the stronger GPA-VGGT with the selected layers protected (\cref{sec:gpasel}), its lead shrinks to 0.55 AUC@5 (CI [$-0.15$, 1.27]) and 0.75 AUC@30 (CI [0.06, 1.67]).
On its worst unseen low-$\rho$ scene (ETH3D electro), the variant loses 2.05 AUC@5, against 2.94 for \methodname and 6.81 for GPA-VGGT.
Because the variant was chosen after the development results, these numbers are not a pre-registered test.

\emph{Other samplers at equal cost (post hoc).} Instead of drawing more hubs, the noise could in principle be reduced by drawing the same number of hubs more cleverly.
A layer-local audit tested this by comparing the current sampler with five alternatives on the 18 hub layers of the frozen recipe, using the 5 development scenes, 2 seeds and 2048 random queries, with 8 independent draws each.
All alternatives keep the numerator and denominator unbiased, and the plan and its continue rules were registered before the run.
\begin{itemize}
  \item \textbf{Fixed share of top-mass keys.} Keeping 10, 25 or 50\% of the top-mass keys exactly and sampling the rest from the renormalised residual raised the output error by 3, 13 and 33\%, so it made the error worse.
  \item \textbf{Other proposals.} An RMS proposal and a head-balanced proposal changed the error by less than 0.2\%, and a value-aware proposal lowered it by only 1.4\%.
  \item \textbf{More draws.} The output error fell as $m^{-1/2}$ with the number of draws; twice the hubs gave 0.71 of the error.
\end{itemize}
The reason the proposal matters so little is that it is close to flat relative to the draw budget: about 254 of 256 draws are distinct keys, so almost no draw is wasted on a repeated key.
Drawing more hubs is therefore the only effective lever, which is why the post-hoc variant above changes the number of hubs rather than the sampler.

\begin{table}[t]
\centering
\small
\setlength{\tabcolsep}{3pt}
\caption{\textbf{Variance components} of the per-run AUC@5 loss of \methodname: 5 development scenes $\times$ 2 frame subsets $\times$ 3 hub draws, nested method of moments; share with a bootstrap 95\% interval over scenes. $^\dagger$Negative estimate ($-0.70$) truncated at 0. $^\ddagger$With more hubs; share of that arm's total (4.41). Bold: the dominant component, the hub draw, and its SD with more hubs.}
\label{tab:variance}
\begin{tabular}{@{}lcccc@{}}
\toprule
\headrow Component & Levels & Var. & SD & Share \\
\midrule
Scene & 5 & 0.00$^\dagger$ & 0.00 & 0\% [0, 23] \\
Frame subset & 10 & 1.00 & 1.00 & 16\% [0, 66] \\
Hub draw & 30 & 5.21 & \textbf{2.28} & \textbf{84\%} [34, 100] \\
\midrule
Total per run & 30 & 6.21 & 2.49 & 100\% \\
Hub draw, more hubs$^\ddagger$ & 30 & 1.62 & \textbf{1.27} & 37\% [19, 100] \\
\bottomrule
\end{tabular}
\end{table}

\subsection{Transfer to \texorpdfstring{$\pi^3$}{pi3}}
\label{sec:pi3}

This preliminary transfer evaluation asks whether the selector, applied without tuning to another model, \pithree, picks the layers that need exact attention; the sparse runs use the original \pithree checkpoint.

\begin{table}[t]
\centering
\small
\compactpending
\setlength{\tabcolsep}{3pt}
\caption{\textbf{Selector on \pithree.} $\Gbar$ (group mean) and predicted role for the Pi3X release, and for the original \pithree checkpoint (last column); loss share: share of the summed positive AUC@5 loss when only that group uses hubs (original checkpoint, 5 development scenes $\times$ 3 subsets). Groups are mapped from VGGT's by relative depth (VGGT 0--8 / 9--10 / 11--16 / 17--23). Predictions were written and hashed before any sparse run. Bold: groups labelled matching, and the observed loss share of the late group.}
\label{tab:pi3}
\begin{tabular}{@{}lcccc@{}}
\toprule
\headrow Group (\pithree layers) & $\Gbar$ & Predicted & Loss share & Orig.\ \pithree ($\Gbar$) \\
\midrule
early (0--6) & 0.76 & diffuse & 0.6\% & diffuse (0.82) \\
middle (7) & 1.42 & diffuse & 0\% & \textbf{matching} (1.54) \\
middle (8--12) & 1.64 & diffuse$^*$ & 14.6\% & diffuse$^*$ (1.70) \\
late (13--17) & 2.96 & \textbf{matching} & \textbf{84.8\%} & \textbf{matching} (3.02) \\
\bottomrule
\end{tabular}
\end{table}

\Cref{tab:pi3} gives the prediction behind \cref{sec:xarch}.
The selector was applied to two \pithree checkpoints: the newer release, whose values the table lists first, and the original checkpoint, on which the sparse runs were made.
\emph{Newer release.} Per layer, the selector labels layers 11--15 as matching ($\Gbar\ge1.5$), layers 0, 1 and 17 as plain ($\Gbar<0.2$), and the rest as diffuse; the row-free rule $s^2\ge3$ picks the same layers 11--15.
$^*$Group labels follow the majority of their layers.
The middle group (three diffuse and two matching layers) is therefore labelled diffuse, although its mean $\Gbar$ is 1.64.
\emph{Pass rule.} The pre-registered pass rule compares the predicted groups with the observed ones.
The observed matching set is the smallest set of groups that carries at least half of the summed positive AUC@5 loss when each group in turn uses hubs, and the rule passes if this set equals the predicted matching set.
\emph{Original checkpoint.} For the original \pithree checkpoint, the selector labels layers 7 and 11--15 as matching, with layer 7 at $\Gbar=1.54$, just above the cut-off.
The row-free rule picks only layers 11--15, and the predicted matching groups are layer 7 and layers 13--17.
\emph{Outcome on the original checkpoint.} The predicted matching groups were layer 7 and layers 13--17, but the observed set is layers 13--17 alone, which carry 84.8\% of the summed loss.
Under hubs, layer 7 loses $-0.11$ AUC@5, that is, nothing.
The pre-registered rule therefore fails because of this one borderline layer ($\Gbar=1.54$), which shows that a statistic just above the cut-off is not sufficient to make a layer need exact attention.
The untuned rule (exact layers 7 and 11--15, hubs elsewhere) loses 4.87 AUC@5 against dense \pithree on the unseen low-$\rho$ scenes and 2.53 on the development scenes (\cref{sec:xarch}).

\emph{Three frame subsets (post hoc).} To check the result on more data, the same untuned rule was run on frame subsets 0--2: 39 runs on the unseen low-$\rho$ scenes and 24 on the high-$\rho$ test scenes, scored against the registered bars.
On the unseen low-$\rho$ scenes, it loses 4.71 AUC@5 (CI [2.48, 7.21]), again above the registered target of 3, against 13.05 for AVGGT(4) and 28.93 for AVGGT(9).
Its gain of 8.35 over AVGGT(4) (CI [3.73, 12.95]) does not meet the registered cost clause, because the rule uses about 2.1$\times$ AVGGT(4)'s attention cost.
On the high-$\rho$ test scenes, it loses 2.13 (CI [0.40, 4.03]), which passes the registered accuracy bar, but its gain over AVGGT(4) there (1.02, CI [$-4.16$, 5.91]) is not significant.
Because both the low-$\rho$ accuracy target and the cost clause are missed, the transfer result remains preliminary.

\FloatBarrier

\end{document}